\documentclass{article} 
\usepackage{iclr2027_conference,times}

\usepackage{amsmath,amsfonts,bm}

\def\eqref#1{equation~\ref{#1}}

\def\1{\bm{1}}

\DeclareMathAlphabet{\mathsfit}{\encodingdefault}{\sfdefault}{m}{sl}
\SetMathAlphabet{\mathsfit}{bold}{\encodingdefault}{\sfdefault}{bx}{n}

\usepackage{hyperref}
\usepackage{url}

\usepackage{booktabs}       
\usepackage{amsfonts}       
\usepackage{nicefrac}       
\usepackage{microtype}      
\usepackage{xcolor}         
\usepackage[normalem]{ulem}
\useunder{\uline}{\ul}{}
\usepackage{pifont}
\usepackage{multirow}
\usepackage{amsfonts}
\usepackage{float}
\usepackage{graphicx}
\usepackage{algorithm}
\usepackage{algpseudocode} 
\usepackage[table]{xcolor}
\usepackage{enumitem}

\usepackage{amsmath,amssymb}
\usepackage[nameinlink,capitalize]{cleveref}
\usepackage{subcaption}  
\usepackage{adjustbox}  
\usepackage{booktabs}
\definecolor{darkgreen}{RGB}{0,150,0}  
\definecolor{darkred}{RGB}{150,0,0}   
\definecolor{red_2}{RGB}{192,0,0}  
\definecolor{darkblue}{RGB}{0,0,150} 
\definecolor{darkpurple}{RGB}{110, 0, 150}
\definecolor{darkorange}{RGB}{255, 140, 0}
\definecolor{Baselineblue}{RGB}{0, 90, 190}     
\definecolor{hotpink}{RGB}{255,105,180}
\definecolor{hermesorange}{RGB}{255,127,0}
\definecolor{orange}{RGB}{234,112,12}
\definecolor{palegreen}{RGB}{120,148,64}

\definecolor{bestbg}{HTML}{F4B6B6}    
\definecolor{secondbg}{HTML}{BFD8FF}  
\definecolor{fdepbg}{HTML}{F3F4F6}    

\definecolor{bestcolor}{HTML}{E64B50}
\definecolor{secondcolor}{HTML}{3C78E0}

\usepackage{mathrsfs}

\definecolor{restategreen}{RGB}{120, 148, 64}
\definecolor{restategold}{RGB}{234, 112, 13}

\title{From Stages to States: Unified Iterative Reasoning for MLLMs}

\author{\textbf{Chuang Yu}$^{1}$ \,
\textbf{Xuefei Ma}$^{2}$ \,
\textbf{Yuanhao Feng}$^{3}$ \,
\textbf{Wuxiong Zeng}$^{3}$ \, 
\textbf{Bo Wang}$^{3}$ \, 
\textbf{Minghong Cai}$^{1}$ \\
\textbf{Dongming Wu}$^{1}$  \,
\textbf{Fei Zhou}$^{1}$ \, 
\textbf{Chen Tang}$^{1}$ \,
\textbf{Wenwen Tong}$^{4}$ \,
\textbf{Lewei Lu}$^{5}$ \,
\textbf{Xiangyu Yue}$^{1}$\footnotemark[2] \\
\\
$^{1}$MMLab, CUHK \quad 
$^{2}$PolyU \quad
$^{3}$Tencent \quad
$^{4}$Peking University  \quad
$^{5}$ CPII  \quad
}

\iclrfinalcopy 
\begin{document}

\maketitle

\begin{abstract}
Multimodal Chain-of-Thought (MCoT) substantially enhances the complex reasoning capabilities of Multimodal Large Language Models (MLLMs) by explicitly modeling intermediate reasoning processes. However, most learning-based MCoT methods follow a stage-centric paradigm, where different reasoning behaviors are assigned to predefined stages and optimized separately. This results in fragmented reasoning and a training-inference reasoning-state gap. To address these limitations, we propose \textbf{\textit{ReState}}, a unified iterative reasoning framework that reformulates multimodal reasoning from a stage-driven pipeline into a state-driven dynamic process. It comprises two complementary components: Unified Reasoning State Modeling (URSM) and Closed-Loop Reasoning-State Learning (CRSL). URSM unifies reasoning construction, correct reasoning preservation, and erroneous reasoning correction as reasoning-state transitions within a shared state space, enabling these heterogeneous reasoning behaviors to be learned jointly. CRSL further incorporates self-generated historical reasoning states during training and recursively updates the current state during inference, thereby mitigating the reasoning-state gap and enabling iterative self-refinement without requiring an additional reflection module, critic, or independent correction stage. Furthermore, we construct VersaCoT-10K, a high-quality CoT dataset spanning eight visual reasoning domains, which provides diverse supervision for multi-domain reasoning and reasoning-state refinement. Extensive experiments demonstrate that ReState exhibits strong adaptability across different MLLM backbones and achieves strong performance on multiple public reasoning benchmarks, supporting the shift from stage-centric reasoning toward state-centric reasoning. Our data and code are available at \href{https://restate-project.github.io}
{\textcolor{hotpink}{\textbf{https://restate-project.github.io}}}.
\end{abstract}

\begin{figure*}[!t]
    \centering
    \includegraphics[width=\linewidth]{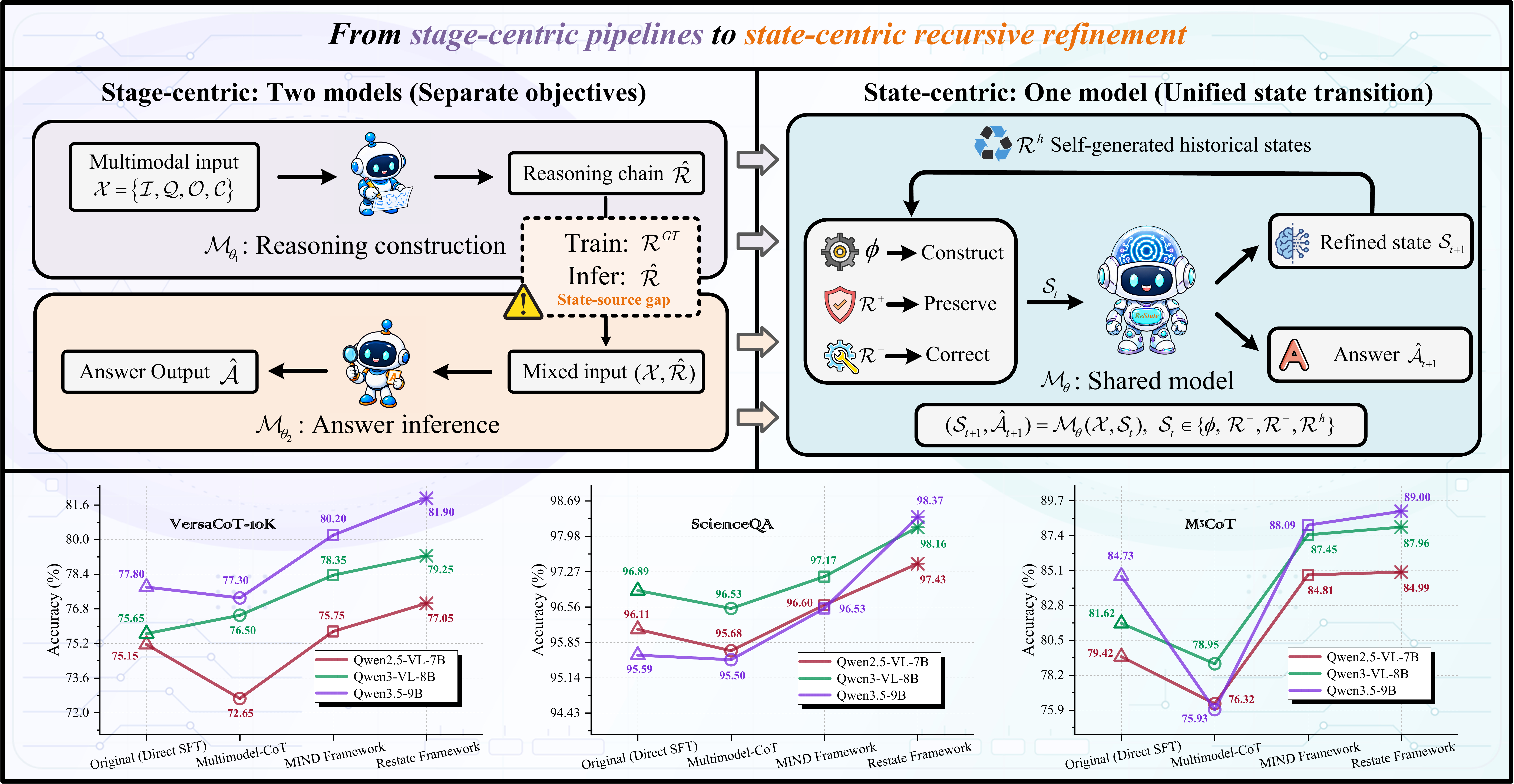}
    \vspace{-0.5cm}
    \caption{Comparison of reasoning paradigms and performance. The stage-centric paradigm assigns different reasoning behaviors to predefined stages, leading to fragmented modeling and a training-inference reasoning-state gap. In contrast, ReState proposes a state-centric paradigm that recursively refines explicit reasoning states with a unified state-update operator, unifying reasoning construction, preservation, and correction while leveraging model-generated historical states for closed-loop learning. Across three backbones, ReState improves performance by \textbf{0.90\%-4.60\%}, \textbf{0.83\%-2.87\%}, and \textbf{0.18\%-13.07\%} on VersaCoT-10K, ScienceQA, and M$^3$CoT, respectively.}
    \label{figs:fig-01}
    \vspace{-0.20cm}
\end{figure*}

\vspace{-0.06cm}
\section{Introduction}
\vspace{-0.06cm}
Multimodal Large Language Models (MLLMs) have achieved remarkable progress across a wide range of tasks, including visual question answering~\citep{yu2026mind,cheng2025comt,yang2025magic,chen2024m3cot}, scientific reasoning~\citep{lu2022learn,wang2024t,li2024m3sciqa,burgess2025microvqa,zhang2025mavis}, and chart understanding~\citep{masry2022chartqa,xu2025chartmoe,masry2025chartqapro,yang2025chartmimic}. As their perceptual and cross-modal capabilities improve, research has increasingly shifted from direct answer prediction toward explicit modeling of intermediate reasoning processes~\citep{zhang2024multimodal,cheng2026visual,dong2026revealing}. Multimodal Chain-of-Thought (MCoT) follows this direction by introducing explicit intermediate reasoning to decompose complex problems into interconnected reasoning steps~\citep{wang2025multimodal}.

Existing MCoT research mainly focuses on constructing and optimizing reasoning chains. Multimodal-CoT~\citep{zhang2024multimodal} introduced a two-stage paradigm of CoT generation and answer inference, while subsequent methods incorporated structured reasoning representations~\citep{mitra2024compositional}, visual grounding~\citep{shao2024visual}, and other mechanisms~\citep{tan2024boosting,wang2024t,he2024multi}. More recently, MIND~\citep{yu2026mind} jointly learns positive and negative CoTs, extending MLLMs from reasoning generation toward reasoning discrimination and correction. Despite these advances, most learning-based MCoT methods remain \emph{stage-centric}: different reasoning behaviors are assigned to predefined stages with stage-specific models, mappings, or objectives. This leads to two limitations. First, reasoning behaviors are modeled separately, resulting in fragmented reasoning without a unified mechanism for state evolution. Second, downstream stages are typically trained on annotated or pre-constructed reasoning states but receive upstream model-generated states during inference, introducing a training-inference reasoning-state gap.

Rather than designing increasingly complex reasoning stages, we ask whether heterogeneous reasoning behaviors can instead be unified as the evolution of explicit reasoning states. Motivated by this view, we propose \textbf{\textit{ReState}} (Reasoning-State Refinement), a unified iterative reasoning framework that reformulates multimodal reasoning from a stage-driven pipeline into a state-driven dynamic process. As illustrated in \cref{figs:fig-01}, ReState focuses not on which predefined stage the model is in, but on what reasoning state it currently has and how that state should be updated. Reasoning states of different sources and qualities can therefore be handled by a shared state-update operator.

Specifically, ReState consists of two complementary designs. First, Unified Reasoning State Modeling (URSM) reformulates reasoning construction, correct reasoning preservation, and erroneous reasoning correction as transitions within a shared state space. Given an empty, correct, or erroneous reasoning state, the same MLLM learns to update it toward a correct reasoning state, unifying previously fragmented reasoning behaviors within a common state-evolution process. Second, Closed-Loop Reasoning-State Learning (CRSL) incorporates self-generated historical reasoning states during training and recursively feeds the current state back into the same model during inference. This closed loop of ``state generation $\rightarrow$ state feedback $\rightarrow$ state update'' reduces the training-inference reasoning-state gap and enables iterative self-refinement without additional reflection modules, critics, or independent correction stages. From \cref{figs:fig-01}, ReState consistently improves performance across multiple datasets and backbones, with gains ranging from 0.18\% to 13.07\%. To further support multi-domain reasoning and reasoning-state refinement, we construct VersaCoT-10K, a high-quality CoT dataset containing 10,000 single-image multiple-choice samples across eight heterogeneous visual reasoning domains for research. The main contributions are summarized as follows:
\begin{itemize}[leftmargin=*]
    \item We propose ReState, a unified iterative reasoning framework for MLLMs. By reformulating the reasoning process from a stage-driven pipeline paradigm to a state-driven dynamic process, ReState enables unified organization and continuous refinement of heterogeneous reasoning behaviors while establishing a paradigm shift from stage-centric to state-centric reasoning.
    \item To address the fragmented modeling of reasoning behaviors in stage-based methods, we propose the URSM mechanism, which unifies heterogeneous reasoning behaviors as reasoning-state transitions, enabling joint optimization of reasoning construction, correct reasoning preservation, and erroneous reasoning correction within a unified state space.
    \item To bridge the reasoning-state gap between training and inference, we propose the CRSL strategy, which learns from self-generated reasoning states during training and recursively refines them during inference, enabling continuous reasoning-state refinement without introducing additional reflection modules or independent correction stages.
    \item We construct VersaCoT-10K, a high-quality CoT dataset covering eight heterogeneous visual reasoning scenarios, providing new training and evaluation resources for studying multi-domain reasoning and reasoning-state refinement in MLLMs.
\end{itemize}

\vspace{-0.06cm}
\section{Related Work}
\vspace{-0.06cm}
\textbf{Prompt-based MCoT Reasoning} elicits explicit reasoning from MLLMs through prompt design, in-context examples, or external information without parameter updates~\citep{wang2025multimodal}. Early approaches mainly relied on few-shot demonstrations and multi-path sampling~\citep{wei2022chain,wang2022self}, while later studies strengthened visual-language interaction through explicit visual descriptions, structured representations, and richer intermediate reasoning forms~\citep{chen2023see,mitra2024compositional,meng2023chain,hu2024visual,cheng2026visual}. Retrieval and tool use further extend reasoning with external knowledge and specialized visual capabilities~\citep{lu2023chameleon,you2023idealgpt,wu2024dettoolchain}. Despite their flexibility, these methods remain sensitive to model capability, prompt design, and sampling strategy~\citep{jin2026look}.

\textbf{Learning-based MCoT Reasoning} explicitly supervises intermediate reasoning to connect visual evidence, reasoning processes, and final answers. Multimodal-CoT~\citep{zhang2024multimodal} first established a two-stage paradigm of CoT generation and answer inference. Building on this, MC-CoT~\citep{tan2024boosting}, DPMM-CoT~\citep{he2024multi}, DDCoT~\citep{zheng2023ddcot}, and T-SciQ~\citep{wang2024t} enhance reasoning learning from the perspectives of multi-path collaboration, cross-modal semantic alignment, responsibility decomposition, and knowledge transfer, respectively. Recent studies have further extended MCoT through directions such as visual grounding, interleaved representations, long-chain reasoning, and reinforcement learning~\citep{li2025vocot,man2025argus,chen2026mint,wu2024mind,lee2024multimodal,dong2025insight,tan2026reason,huang2026vision}. In addition, MIND~\citep{yu2026mind} further learns from positive and negative CoTs, extending MLLMs from reasoning generation to reasoning discrimination and correction. Nevertheless, existing methods generally treat intermediate CoTs as stage-specific outputs rather than reusable states for recursive refinement. This stage-centric design results in fragmented reasoning and a training--inference reasoning-state gap, motivating our shift toward state-centric reasoning.

\begin{figure*}[!t]
    \centering
    \includegraphics[width=\linewidth]{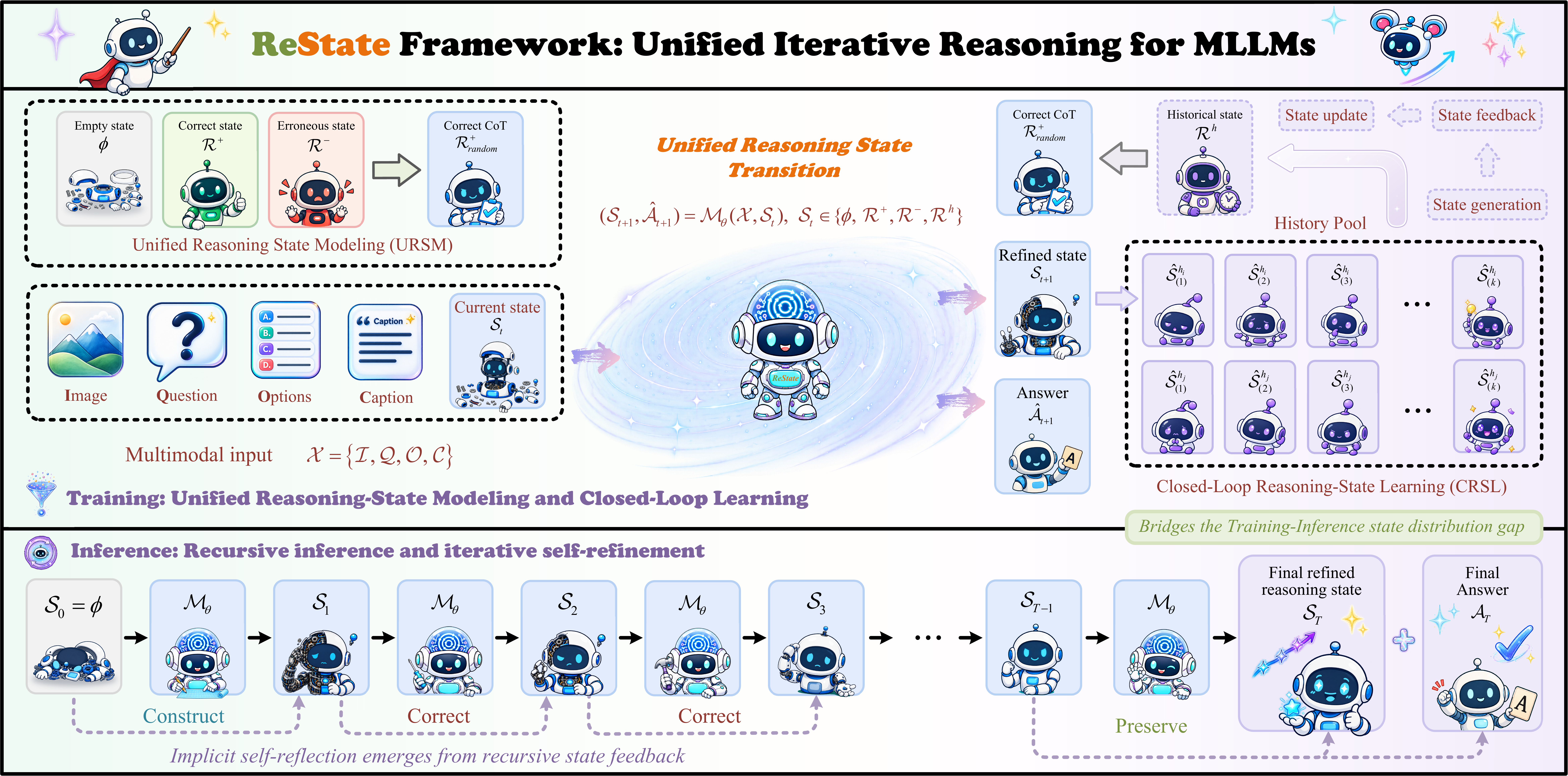}
    \vspace{-0.5cm}
    \caption{Overview of the ReState framework. During training, URSM learns a shared state-update operator from empty, correct, and erroneous reasoning states, while CRSL feeds model-generated historical states back into training to reduce the training-inference reasoning-state mismatch. At inference, the same model recursively updates the current reasoning state and answer, enabling iterative self-refinement without additional critics, reflection modules, or separate correction stages.}
    \label{figs:fig-02}
    \vspace{-0.2cm}
\end{figure*}

\vspace{-0.06cm}
\section{Preliminary: Stage-centric Reasoning Paradigm}
\vspace{-0.06cm}
\label{sec:preliminary}
Most existing learning-based MCoT methods follow a stage-centric reasoning paradigm~\citep{zhang2024multimodal,tan2024boosting,he2024multi,zheng2023ddcot,wang2024t,yu2026mind}, typically decomposing reasoning into CoT generation and answer inference. These stages are modeled independently rather than through unified reasoning-state evolution.

In the CoT generation stage, the model generates an explicit reasoning chain from the input:
\begin{gather}
\hat {\cal R} = {{\cal M}_{{\theta _1}}}({\cal X}),\;\;{\cal X} = \left\{ {{\cal I},\;{\cal Q},\;{\cal O},\;{\cal C}} \right\}
\end{gather}
where ${\cal I}$, ${\cal Q}$, ${\cal O}$ and ${\cal C}$ denote the image, question, candidate options, and image caption, respectively. ${{\cal M}_{{\theta _1}}}$ denotes an MLLM parameterized by ${\theta _1}$. This stage is supervised using annotated reasoning chains $\hat {\cal R}^{GT}$, learning a mapping from multimodal inputs to explicit reasoning processes.

In the answer inference stage, the generated reasoning chain is used as an additional condition:
\begin{gather}
\hat {\cal A} = {{\cal M}_{{\theta _2}}}({\cal X},\hat {\cal R})
\end{gather}
where $\hat {\cal A}$ denotes the predicted answer, ${{\cal M}_{{\theta _2}}}$ denotes an MLLM parameterized by ${\theta _2}$. Notably, this stage is trained on annotated reasoning ${\cal R}^{GT}$ but receives model-generated reasoning $\hat{\cal R}$ by the preceding stage at inference, resulting in a training-inference reasoning-state gap.

Thus, predefined reasoning stages lead to fragmented reasoning and a training-inference reasoning-state gap. These limitations motivate a unified formulation based on reasoning-state evolution.

\section{ReState: A Unified Iterative Reasoning Framework}
\label{sec:restate_overview}
\subsection{Overview}
To address the limitations of stage-centric reasoning in Section \ref{sec:preliminary}, we propose ReState, which reformulates multimodal reasoning from a stage-driven process into a unified dynamic evolution of explicit reasoning states. As shown in \cref{figs:fig-02}, ReState consists of unified state learning during training and recursive state refinement during inference, built upon two complementary designs: URSM and CRSL. During training, unlike existing MCoT methods~\citep{yu2026mind}, URSM unifies reasoning construction, correct reasoning preservation, and erroneous reasoning correction as reasoning-state transitions implemented by the same MLLM. Building on URSM, CRSL further incorporates self-generated historical states to expose the state-update operator to model-induced states and reduce the training-inference reasoning-state gap. During inference, ReState starts from an empty state and recursively feeds the current reasoning state back into the same state-update operator, progressively preserving valid reasoning, correcting potential errors, and updating the answer. The entire process reuses the same model throughout and achieves iterative self-refinement without requiring an additional reflection module, critic, or independent correction stage.

\subsection{Unified Reasoning State Modeling (URSM) Mechanism}
Existing MCoT methods typically treat reasoning chains as stage-specific intermediate outputs. In contrast, URSM represents intermediate reasoning of different qualities as continuously updatable states and models them with a shared state-update operator.
\begin{gather}
({\cal S}_{t + 1},{\hat {\cal  A}_{t + 1}}) = {{\cal M}_\theta }({\cal X}, {\cal S}_t),\;\;{\cal S}_t \in \{ \phi ,\,\,{\cal R }^ +,{\cal R }^ - \} 
\end{gather}
where ${{\cal M}_\theta }$ is the shared reasoning-state transition function implemented by a single MLLM parameterized by $\theta$. ${\cal S}_t$, ${\cal S}_{t+1}$ and $\hat{\cal A}_{t+1}$ denote the current reasoning state, updated reasoning state and predicted answer. $\phi ,\,\,{\cal R }^ +,{\cal R }^ -$ denote the empty, valid, and erroneous reasoning states, respectively, with $\phi$ represented by the special token $[EMPTY\_SOLUTION]$.

Within this unified state space, heterogeneous reasoning behaviors are reformulated as different reasoning-state transitions. When ${{\cal S}_t} = \phi$, the model constructs an initial reasoning state from the multimodal input, corresponding to reasoning construction. When ${{\cal S}_t} = {\cal R }^ +$, it preserves valid reasoning evidence while improving completeness and consistency, corresponding to correct reasoning preservation. When ${{\cal S}_t} = {\cal R }^ -$, it revisits the original multimodal evidence to correct errors in the current state, corresponding to erroneous reasoning correction. By jointly learning these transitions, the model moves beyond generating reasoning chains from scratch to understanding, re-evaluating, and refining existing reasoning states, extending its capability from reasoning generation to reasoning-state understanding and refinement. 


\begin{figure*}[t]
    \centering
    \includegraphics[width=\linewidth]{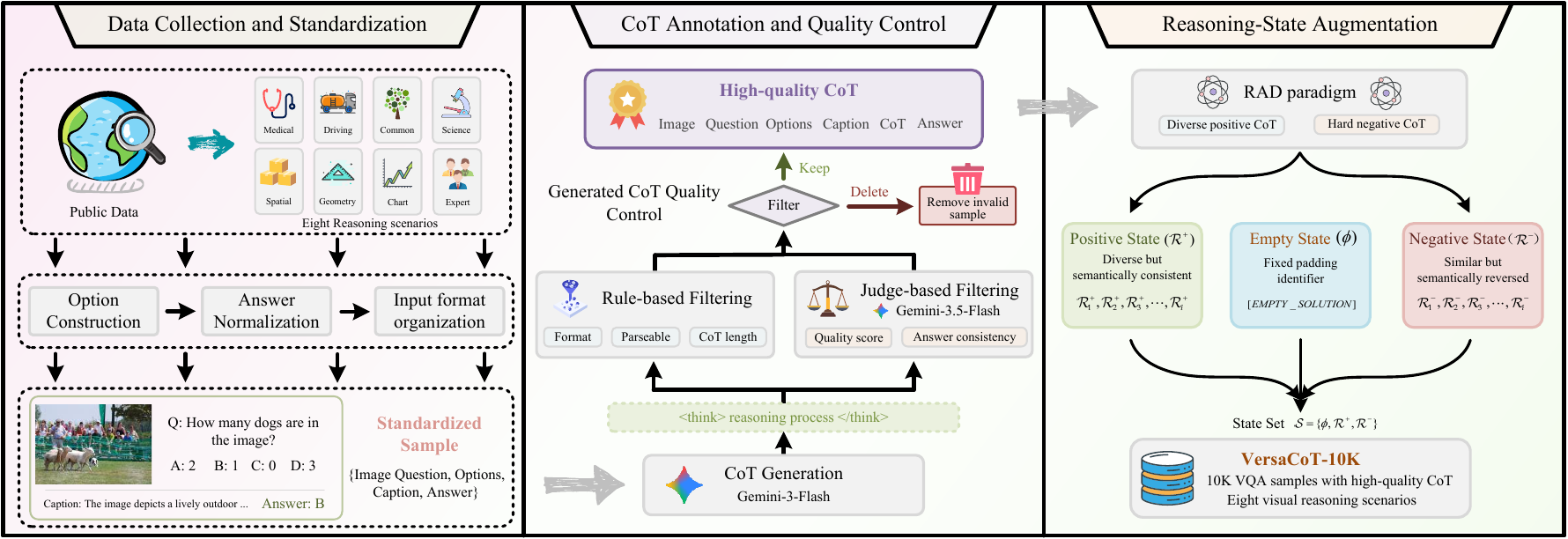}
    \vspace{-0.5cm}
    \caption{Construction pipeline of VersaCoT-10K dataset. It consists of three stages: \textit{Data Collection and Standardization}, \textit{CoT Annotation and Quality Control}, and \textit{Reasoning-State Augmentation}.}
    \label{figs:fig-03}
    \vspace{-0.2cm}
\end{figure*}

\subsection{Closed-loop Reasoning-State Learning (CRSL) Strategy}
Although URSM provides a unified reasoning-state transition mechanism, training only on pre-constructed states cannot fully cover the model-generated states encountered during recursive inference. Differences in their source and error patterns therefore introduce a training--inference reasoning-state gap, limiting the model's ability to refine its own historical states.

To address this issue, CRSL feeds self-generated historical reasoning states back into training, extending the URSM state space to ${{\cal S}_t} \in \{ \phi ,\;{{\cal R}^ + },{{\cal R}^ - },{{\cal R}^h}\}$. For each training sample $i$, CRSL maintains a sample-specific, fixed-capacity history pool:
\begin{gather}
{{\cal H}_i} = \{ \hat {\cal S}_{(1)}^i,\hat {\cal S}_{(2)}^i, \ldots ,\hat {\cal S}_{(k)}^i\}
\end{gather}
where $\hat {\cal S}_{(k)}^i$ denotes the reasoning state generated for sample $i$ at the $k_{th}$ round. The pool is updated in a first-in-first-out manner, allowing ${{\cal H}_i}$ to continuously track the state distribution induced by the model throughout training. In our experiments, the pool capacity is set to 2.

By reusing these historical states as subsequent training inputs, CRSL forms a closed loop of ``state generation $\rightarrow$ state feedback $\rightarrow$ state update''. This exposes the state-update operator to model-induced states closer to those encountered during recursive inference, thereby reducing the reasoning-state gap. Moreover, revisiting historical states encourages the model to re-evaluate and correct its reasoning, providing a basis for iterative self-refinement.

At inference, ReState recursively applies the same state-update operator from an empty state:
\begin{gather}
{{\cal S}_0} = \phi ,\quad ({{\cal S}_{t + 1}},{\hat {\cal A}^{t + 1}}) = {{\cal M}_\theta }({\cal X},{{\cal S}_t}),\quad t = 0, \ldots , {\cal T} - 1
\end{gather}
This produces the state-evolution trajectory ${{\cal S}_0} \to {{\cal S}_1} \to  \cdots  \to {{\cal S}_T}, \,\, \hat {\cal A} = {\hat {\cal A}^T}$. At each iteration, the model feeds back the current state to preserve valid reasoning and correct potential errors. Because the entire recursive process reuses the same state-update operator, ReState achieves iterative self-refinement without additional reflection modules, critics, or independent correction stages.

\begin{table*}[t]
\caption{Main results (\%) on the VersaCoT-10K dataset. MED = medical, DRIVE = driving and safety, COMMON = visual commonsense, DIAG = scientific diagram, SPAT = spatial perception, GEO = geometric and mathematical, CHART = chart computation, and MULTI = multidisciplinary knowledge. \textcolor{bestcolor}{Red} denotes the best result, and \textcolor{secondcolor}{blue} denotes the second best result. }
\vspace{-6pt}
\label{tab:tab01}
\setlength{\tabcolsep}{2mm}
\renewcommand{\arraystretch}{1.00}
\resizebox{\linewidth}{!}{
\begin{tabular}{cccccccccccc}
\hline
\multicolumn{1}{c|}{Model}                                                           & \multicolumn{1}{c|}{Learning}                          & \multicolumn{1}{c|}{Size}                       & \multicolumn{1}{c|}{MED}                                    & \multicolumn{1}{c|}{DRIVE}                                  & \multicolumn{1}{c|}{COMMON}                                 & \multicolumn{1}{c|}{DIAG}                                   & \multicolumn{1}{c|}{SPAT}                                   & \multicolumn{1}{c|}{GEO}                                    & \multicolumn{1}{c|}{CHART}                         & \multicolumn{1}{c|}{MULTI}                                  & \textbf{Avg}   \\ \hline
\multicolumn{1}{c|}{Random}                                                          & \multicolumn{1}{c|}{-}                                 & \multicolumn{1}{c|}{-}                          & \multicolumn{1}{c|}{22.00}                                       & \multicolumn{1}{c|}{46.80}                                       & \multicolumn{1}{c|}{26.80}                                       & \multicolumn{1}{c|}{26.27}                                       & \multicolumn{1}{c|}{39.87}                                       & \multicolumn{1}{c|}{22.93}                                       & \multicolumn{1}{c|}{22.00}                              & \multicolumn{1}{c|}{25.60}                                       &  29.03              \\
\hline
\end{tabular}
}
\vspace{-0.2cm}
\end{table*}

\section{VersaCoT-10K Dataset}
Existing multimodal reasoning datasets often lack explicit CoT annotations, broad domain coverage, or diverse reasoning states for reasoning refinement. To address these limitations, we construct VersaCoT-10K, a high-quality dataset containing 10,000 single-image multiple-choice samples across eight heterogeneous visual reasoning domains. As shown in \cref{figs:fig-03}, VersaCoT-10K is constructed through three stages. First, we collect reasoning-intensive samples from multiple public benchmarks and standardize their options, answers, and multimodal inputs. Second, Gemini-3-Flash~\citep{google2025gemini3flash} generates visually grounded CoTs, followed by rule-based filtering and multimodal quality assessment with Gemini-3.5-Flash~\citep{google2026gemini35flash}. Third, following RAD~\citep{yu2026mind}, each quality-filtered CoT is augmented with positive and hard negative reasoning states. Together with the empty state, they provide supervision for learning reasoning construction, correct reasoning preservation, and erroneous reasoning correction. Notably, historical states used by CRSL are instead generated dynamically during ReState training. Detailed construction procedures, prompts, quality-control criteria, and dataset statistics are provided in \textit{Appendix}~\ref{sec:a-detailed-construction-for-versaCoT-10K} and ~\ref{sec:a-data statistics-for-versaCoT-10K}.

\section{Experiments}

\subsection{Experimental Setup}
\textbf{\textit{1) Datasets.}} We evaluate ReState on VersaCoT-10K and two public multimodal reasoning benchmarks, ScienceQA~\citep{lu2022learn} and M$^3$CoT~\citep{chen2024m3cot}. Following~\citep{yu2026mind}, we adopt the corresponding RAD versions, which provide diverse reasoning states.

\textbf{\textit{2) Implementation Details.}} We train all Qwen models for 5 epochs with a learning rate of $1e^{-5}$. Iterative self-refinement uses 9 rounds by default and reports peak accuracy. For settings without iterative refinement, the model performs one additional answer generation or correction round after initial CoT generation. All experiments are conducted on 8 NVIDIA H20 GPUs.

\begin{table*}[t]
\caption{Main results (\%) on the ScienceQA dataset. NAT = natural science, SOC= social science, LAN = language science, TXT = text context, IMG = image context, NO = no context, G1-6 = grades 1-6, G7-12 = grades 7-12. \textcolor{bestcolor}{Red} denotes the best result, and \textcolor{secondcolor}{blue} denotes the second best result. }
\vspace{-6pt}
\label{tab:tab02}
\setlength{\tabcolsep}{2mm}
\renewcommand{\arraystretch}{1.00}
\resizebox{\linewidth}{!}{
\begin{tabular}{c|c|c|c|c|c|c|c|c|c|c|c}
\hline
Model                                   & Learning           & Size     & NAT            & SOC            & LAN            & TXT            & IMG            & NO             & G1-6           & G7-12          & \textbf{Avg}   \\ \hline
Random                                  & -                  & -        & 40.28          & 46.13          & 29.25          & 47.45          & 40.08          & 33.66          & 39.35          & 40.67          & 39.83          \\ 
Human                                   & -                  & -        & 90.23          & 84.97          & 87.48          & 89.60          & 87.50          & 88.10          & 91.59          & 82.42          & 88.40          \\ \hline
ViLT                                    & \textit{Fine-tune}          & 113M     & 60.48          & 63.89          & 60.27          & 63.20          & 61.38          & 57.00          & 60.72          & 61.90          & 61.14          \\ 
Patch-TRM                               & \textit{Fine-tune}          & 90M      & 65.19          & 46.79          & 65.55          & 66.96          & 55.28          & 64.95          & 58.04          & 67.50          & 61.42          \\ 
VisualBERT                              & \textit{Fine-tune}          & 111M     & 59.33          & 69.18          & 61.18          & 62.71          & 62.17          & 58.54          & 62.96          & 59.92          & 61.87          \\ \hline
LLaMA-Adapter                           & \textit{Fine-tune}          & 6B       & 84.37          & 88.30          & 84.36          & 83.72          & 80.32          & 86.90          & 85.83          & 84.05          & 85.19          \\ 
LLaVA                                   & \textit{Fine-tune}          & 13B      & 90.36          & 95.95          & 88.00          & 89.49          & 88.00          & 90.66          & 90.93          & 90.90          & 90.92          \\
\hline
\end{tabular}
}
\vspace{-0.2cm}
\end{table*}

\begin{table*}[t]
\caption{Main results (\%) on the M$^3$CoT dataset. The ``Random'' and ``Human'' results are averaged over three trials. \textcolor{bestcolor}{Red} denotes the best result, and \textcolor{secondcolor}{blue} denotes the second best result. }
\vspace{-6pt}
\label{tab:tab03}
\setlength{\tabcolsep}{1.5mm}
\renewcommand{\arraystretch}{1.00}
\resizebox{\linewidth}{!}{

\begin{tabular}{cccccccccccc}
\hline
\multicolumn{1}{c|}{Model}                                                           & \multicolumn{1}{c|}{Size}                       & \multicolumn{3}{c|}{Science}                                                                                                                                                            & \multicolumn{3}{c|}{Commonsense}                                                                                                                                                        & \multicolumn{3}{c|}{Mathematics}                                                                                                                                                        &                                  \\ \cline{1-11}
\multicolumn{1}{c|}{}                                                                & \multicolumn{1}{c|}{}                           & \multicolumn{1}{c|}{Lang}                                   & \multicolumn{1}{c|}{Natural}                                & \multicolumn{1}{c|}{Social}                                 & \multicolumn{1}{c|}{Physical}                               & \multicolumn{1}{c|}{Social}                                 & \multicolumn{1}{c|}{Temporal}                               & \multicolumn{1}{c|}{Algebra}                                & \multicolumn{1}{c|}{Geometry}                               & \multicolumn{1}{c|}{Theory}                                 & \multirow{-2}{*}{\textbf{Total}} \\
\hline
\end{tabular}

}
\vspace{-0.2cm}
\end{table*}

\subsection{Comparison with SOTA Methods}
\textit{\textbf{1) Evaluation on the VersaCoT-10K dataset.}} As shown in \cref{tab:tab01}, VersaCoT-10K exhibits clear performance gaps across MLLMs, with overall zero-shot accuracy ranging from 67.45\% for Qwen2.5-VL-7B~\citep{bai2025qwen2} to 86.85\% for Grok-4.6~\citep{xai2026grok46}. Performance also varies substantially across reasoning categories, and even frontier models remain unsaturated. These results establish VersaCoT-10K as a challenging and discriminative benchmark for fine-grained multimodal reasoning.
 
More importantly, under identical backbone and training settings, ReState achieves the best average performance on all three backbones, reaching 77.05\%, 79.25\%, and 81.90\% with Qwen2.5-VL-7B~\citep{bai2025qwen2}, Qwen3-VL-8B~\citep{bai2025qwen3vltechnicalreport}, and Qwen3.5-9B~\citep{qwen3.5}, respectively. Compared with direct fine-tuning, Multimodal-CoT~\citep{zhang2024multimodal}, and MIND~\citep{yu2026mind}, ReState improves accuracy by 1.90\%-4.10\%, 2.75\%-4.60\%, and 0.90\%-1.70\%, respectively. In addition, compared with the latest MIND, ReState also reduces training time by 31.2\% (from 9.86 h to 6.78 h) and requires only one model to be deployed instead of two. Moreover, combining the results in \cref{tab:tab01} and \cref{tab:tab04}, even without iterative self-refinement, i.e., under an inference cost comparable to MIND, ReState still improves accuracy by 0.15\%-0.90\% across the three backbones.

Notably, ReState with Qwen3.5-9B reaches 81.90\%, outperforming substantially larger models such as Qwen3.6-27B~\citep{qwen3.6-27b}, GLM-4.6V (106B)~\citep{vteam2026glm45vglm41vthinkingversatilemultimodal}, MiniMax-M3 (428B)~\citep{lai2026minimaxsparseattention}, and Kimi-K2.6 (1T)~\citep{moonshot2026kimik26}. Models that further surpass ReState are generally much larger. For example, the 2.8T-parameter Kimi-K3~\citep{kimiteam2026kimik3openfrontier} reaches 85.65\%. Although fine-tuning and zero-shot evaluation are not directly comparable, these results show that ReState effectively enhances the reasoning capability of medium-scale MLLMs. Its consistent gains across backbones and reasoning categories further support the effectiveness of state-centric reasoning. More comparisons and analyses are provided in \textit{Appendix} ~\ref{sec:s-evaluation-on-the-versaCoT-10K-dataset} and ~\cref{tab:s-tab08}.

\textit{\textbf{2) Evaluation on the ScienceQA dataset.}} As shown in \cref{tab:tab02}, ReState achieves strong overall performance on ScienceQA. In particular, ReState with Qwen3.5-9B attains the best result of 98.37\%, outperforming strong closed-source zero-shot models such as Qwen3.8-Max (98.11\%)~\citep{qwen3.8}, Claude Opus 5 (98.16\%)~\citep{anthropic2026opus5}, and Kimi-K3 (98.33\%)~\citep{kimiteam2026kimik3openfrontier}, as well as existing learning-based MCoT methods~\citep{zhang2024multimodal,tan2024boosting,wang2024t,yu2026mind}. Under identical backbone and training settings, ReState consistently outperforms direct SFT, Multimodal-CoT~\citep{zhang2024multimodal}, and MIND~\citep{yu2026mind} by 1.27\%-2.78\%, 1.63\%-2.87\%, and 0.83\%-1.84\%, respectively. Moreover, ReState achieves the best results across all eight sub-categories. Overall, these results show the consistent effectiveness of ReState across MLLMs. More comparisons and analyses are provided in \textit{Appendix} ~\ref{sec:s-evaluation-on-the-scienceQA-dataset} and ~\cref{tab:s-tab09}.

\textit{\textbf{3) Evaluation on the M$^3$CoT dataset.}} As shown in \cref{tab:tab03}, ReState achieves 84.99\%, 87.96\%, and 89.00\% with Qwen2.5-VL-7B, Qwen3-VL-8B, and Qwen3.5-9B, respectively. In particular, ReState with Qwen3.5-9B attains the best result of 89.00\%, outperforming strong zero-shot models such as Gemini-3.5-Flash (82.96\%)~\citep{google2026gemini35flash}, Kimi-K3 (81.32\%)~\citep{kimiteam2026kimik3openfrontier}, and Grok-4.6 (81.02\%)~\citep{xai2026grok46}, while trailing human performance (91.61\%) by only 2.61\%. Under identical backbone settings, ReState improves over direct SFT, Multimodal-CoT~\citep{zhang2024multimodal}, and MIND~\citep{yu2026mind} by 4.27\%–6.34\%, 8.67\%–13.07\%, and 0.18\%–0.91\%, respectively, demonstrating consistent gains across different base models. It also achieves the best results on 7 of 9 sub-tasks, with notable gains on Theory (from 61.90\% to 90.48\%) and Algebra (from 70.00\% to 80.00\%). These results show that ReState provides consistent benefits across diverse knowledge structures and reasoning patterns, with particularly clear advantages on more challenging multimodal reasoning tasks. More comparisons and analyses are provided in \textit{Appendix} ~\ref{sec:s-evaluation-on-the-m3cot-dataset} and ~\cref{tab:s-tab10}.

\subsection{Break-Down Ablation}
We conduct progressive ablations on VersaCoT-10K across three backbones to evaluate the contributions of URSM and CRSL. From \cref{tab:tab04}, URSM alone improves the baseline by 0.50\% (from 75.15\% to 75.65\%)-2.55\% (from 75.65\% to 78.20\%), demonstrating the benefit of unifying reasoning construction, preservation, and correction as state transitions within the same MLLM. Since both training-time self-history state (SHS) and inference-time iterative self-refinement (ISR) in CRSL build upon URSM, we further ablate SHS and ISR on top of URSM. Adding training-time SHS yields an additional 0.30\% (from 78.20\% to 78.50\%) - 2.15\% (from 78.95\% to 81.10\%) gain, indicating that self-generated historical states help the model better adapt to model-induced states. Adding inference-time ISR further improves performance by 0.45\% (78.20\% to 78.65\%) - 0.55\% (from 75.65\% to 76.20\%), validating the effectiveness of recursive state feedback for refining existing reasoning. Combining URSM with both SHS and ISR achieves the best performance across all three backbones, improving the respective baselines by 1.90\%, 3.60\%, and 4.10\%. Overall, URSM provides unified state modeling, while SHS reduces the training-inference reasoning-state gap and ISR enables recursive refinement. Their complementary gains support the effectiveness of the proposed state-centric reasoning framework. More experiments and analyses are provided in \textit{Appendix} ~\ref{sec:a-more-ablation experiments}.

\begin{table*}[t]
\centering
\caption{The Break-Down ablation of ReState on the VersaCoT-10K dataset. SHS and ISR denote the training-time self-history state mechanism and inference-time iterative self-refinement in CRSL.}
\vspace{-6pt}
\label{tab:tab04}
\setlength{\tabcolsep}{3.4mm}
\renewcommand{\arraystretch}{1.00}
\resizebox{1\linewidth}{!}{

\begin{tabular}{c|c|cc|c|c|c}
\hline
\multirow{2}{*}{Scheme} 
& \multirow{2}{*}{URSM} 
& \multicolumn{2}{c|}{CRSL} 
& \multirow{2}{*}{Qwen2.5-VL-7B} 
& \multirow{2}{*}{Qwen3-VL-8B} 
& \multirow{2}{*}{Qwen3.5-9B} \\ 
\cline{3-4}

& 
& \multicolumn{1}{c|}{SHS} 
& ISR 
& 
& 
& \\ 
\hline

Baseline
& \textbf{\ding{55}}
& \multicolumn{1}{c|}{\textbf{\ding{55}}}
& \textbf{\ding{55}}
& 75.15 ({\color[HTML]{7030A0}+1.90})
& 75.65 ({\color[HTML]{7030A0}+3.60})
& 77.80 ({\color[HTML]{7030A0}+4.10})
\\ 
\hline

Baseline w/ URSM
& \textbf{\ding{51}}
& \multicolumn{1}{c|}{\textbf{\ding{55}}}
& \textbf{\ding{55}}
& 75.65 ({\color[HTML]{7030A0}+1.40})
& 78.20 ({\color[HTML]{7030A0}+1.05})
& 78.95 ({\color[HTML]{7030A0}+2.95})
\\

Baseline w/ URSM, CRSL(SHS)
& \textbf{\ding{51}}
& \multicolumn{1}{c|}{\textbf{\ding{51}}}
& \textbf{\ding{55}}
& 76.20 ({\color[HTML]{7030A0}+0.85})
& 78.50 ({\color[HTML]{7030A0}+0.75})
& 81.10 ({\color[HTML]{7030A0}+0.80})
\\

Baseline w/ URSM, CRSL(ISR)
& \textbf{\ding{51}}
& \multicolumn{1}{c|}{\textbf{\ding{55}}}
& \textbf{\ding{51}}
& 76.20 ({\color[HTML]{7030A0}+0.85})
& 78.65 ({\color[HTML]{7030A0}+0.60})
& 79.45 ({\color[HTML]{7030A0}+2.45})
\\ 
\hline

\rowcolor[HTML]{EFEFEF}
\textbf{ReState (Final)}
& \textbf{\ding{51}}
& \multicolumn{1}{c|}{\textbf{\ding{51}}}
& \textbf{\ding{51}}
& \textbf{77.05}
& \textbf{79.25}
& \textbf{81.90}
\\ 
\hline

\end{tabular}

}
\vspace{-0.2cm}
\end{table*}

\begin{figure*}[!t]
    \centering
    \includegraphics[width=\linewidth]{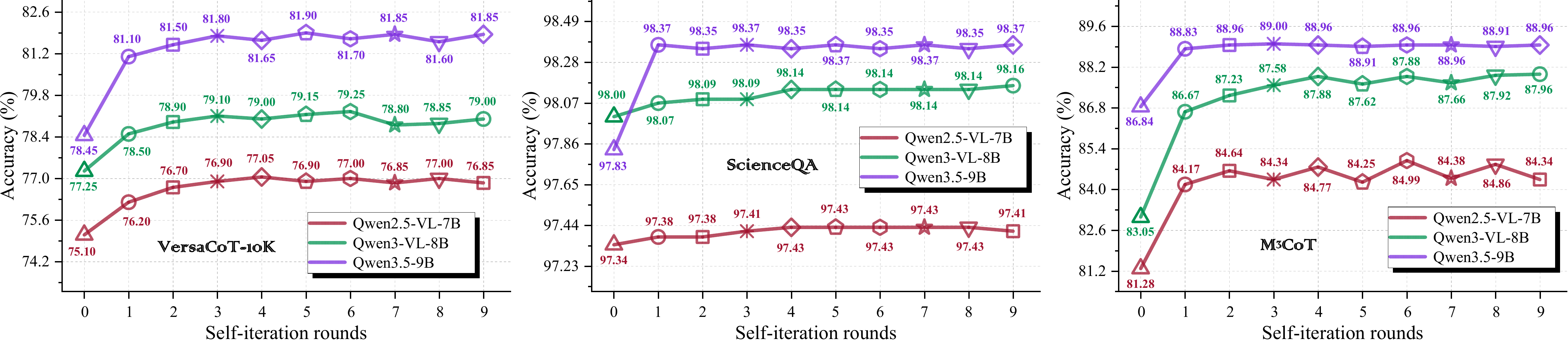}
    \vspace{-0.5cm}
    \caption{Performance evolution over iterative self-refinement rounds across different MLLMs and datasets. Round 0 denotes the initial reasoning, with most gains achieved in the first few rounds.}
    \label{figs:fig-05}
    \vspace{-0.2cm}
\end{figure*}

\subsection{Discussion of ``Iterative Self-Refinement''}
To further analyze the iterative self-refinement capability of ReState at inference, we evaluate its performance under different numbers of refinement rounds in \cref{figs:fig-05}. Across all three backbones and datasets, ReState consistently benefits from iterative self-refinement. After only one refinement round, performance improves by 1.10\%-2.65\% on VersaCoT-10K, 0.04\%-0.54\% on ScienceQA, and 1.99\%-3.62\% on M$^3$CoT. Compared with round 0, the best gains reach 1.95\%-3.45\%, 0.09\%-0.54\%, and 2.16\%-4.91\%, respectively. Performance generally improves rapidly in early rounds and then stabilizes with minor fluctuations, with larger gains on the more challenging VersaCoT-10K and M$^3$CoT. More importantly, ReState reuses the same MLLM throughout the refinement process and requires no additional reflection module, critic, or independent correction model. Instead, refinement is achieved through a recursive loop of ``state feedback → re-evaluation → state update''. These results validate its ability to iteratively refine explicit reasoning states during inference, further supporting the effectiveness of state-centric reasoning. More discussions are provided in \textit{Appendix} ~\ref{sec:more-discussion}.

\section{Conclusion}
This work revisits the stage-centric paradigm of learning-based MCoT and identifies two key limitations: fragmented modeling of reasoning behaviors and the training-inference reasoning-state gap. Therefore, we propose ReState, a unified iterative reasoning framework that reformulates multimodal reasoning as dynamic state evolution. URSM unifies reasoning construction, preservation, and correction within a shared state-update mechanism, while CRSL incorporates historical-state learning and recursive feedback to reduce the distribution gap and enable iterative self-refinement without additional reflection or correction modules. We also construct VersaCoT-10K, a high-quality CoT dataset spanning eight visual reasoning domains. Extensive experiments demonstrate the consistent effectiveness and cross-model generalizability of ReState. We hope this work can draw attention to the research on state-centric multimodal reasoning and advance MLLMs toward more general, reliable, and adaptive reasoning systems.

\newpage

\subsection*{AI use statement}
We used generative AI tools for manuscript polishing and several components of the data construction and evaluation pipelines. Specifically, generative AI was used to assist with language polishing. For VersaCoT-10K, Gemini-3-Flash was used to generate candidate CoT annotations, Gemini-3.5-Flash was used as a multimodal quality judge to filter generated rationales, and Qwen2.5-VL-72B was used to generate image captions. Generative models were used as assisting tools rather than as substitutes for author verification. We reviewed the AI-assisted writing, data-processing procedures, generated annotations, and reported results, and take full responsibility for the final content of this work, including all text, data, and artifacts produced with the aid of generative AI.

\bibliography{iclr2027_conference}

@inproceedings{yu2026mind,
  title={MIND: Multi-rationale INtegrated Discriminative Reasoning Framework for Multi-modal Large Models},
  author={Yu, Chuang and Zhao, Jinmiao and Zhao, Mingxuan and Liu, Yunpeng and Shu, Xiujun and Feng, Yuanhao and Wang, Bo and Yue, Xiangyu},
  booktitle={Forty-third International Conference on Machine Learning},
  year={2026}
}

@inproceedings{cheng2025comt,
  title={Comt: A novel benchmark for chain of multi-modal thought on large vision-language models},
  author={Cheng, Zihui and Chen, Qiguang and Zhang, Jin and Fei, Hao and Feng, Xiaocheng and Che, Wanxiang and Li, Min and Qin, Libo},
  booktitle={Proceedings of the AAAI Conference on Artificial Intelligence},
  volume={39},
  number={22},
  pages={23678--23686},
  year={2025}
}

@inproceedings{yang2025magic,
  title={Magic-vqa: Multimodal and grounded inference with commonsense knowledge for visual question answering},
  author={Yang, Shuo and Han, Caren and Luo, Siwen and Hovy, Eduard},
  booktitle={Findings of the Association for Computational Linguistics: ACL 2025},
  pages={16967--16986},
  year={2025}
}

@inproceedings{chen2024m3cot,
  title={M3cot: A novel benchmark for multi-domain multi-step multi-modal chain-of-thought},
  author={Chen, Qiguang and Qin, Libo and Zhang, Jin and Chen, Zhi and Xu, Xiao and Che, Wanxiang},
  booktitle={Proceedings of the 62nd Annual Meeting of the Association for Computational Linguistics (Volume 1: Long Papers)},
  pages={8199--8221},
  year={2024}
}

@article{lu2022learn,
  title={Learn to explain: Multimodal reasoning via thought chains for science question answering},
  author={Lu, Pan and Mishra, Swaroop and Xia, Tanglin and Qiu, Liang and Chang, Kai-Wei and Zhu, Song-Chun and Tafjord, Oyvind and Clark, Peter and Kalyan, Ashwin},
  journal={Advances in neural information processing systems},
  volume={35},
  pages={2507--2521},
  year={2022}
}

@inproceedings{wang2024t,
  title={T-sciq: Teaching multimodal chain-of-thought reasoning via large language model signals for science question answering},
  author={Wang, Lei and Hu, Yi and He, Jiabang and Xu, Xing and Liu, Ning and Liu, Hui and Shen, Heng Tao},
  booktitle={Proceedings of the AAAI Conference on Artificial Intelligence},
  volume={38},
  number={17},
  pages={19162--19170},
  year={2024}
}

@inproceedings{li2024m3sciqa,
  title={M3SciQA: A multi-modal multi-document scientific QA benchmark for evaluating foundation models},
  author={Li, Chuhan and Shangguan, Ziyao and Zhao, Yilun and Li, Deyuan and Liu, Yixin and Cohan, Arman},
  booktitle={Findings of the Association for Computational Linguistics: EMNLP 2024},
  pages={15419--15446},
  year={2024}
}

@inproceedings{burgess2025microvqa,
  title={Microvqa: A multimodal reasoning benchmark for microscopy-based scientific research},
  author={Burgess, James and Nirschl, Jeffrey J and Bravo-S{\'a}nchez, Laura and Lozano, Alejandro and Gupte, Sanket Rajan and Galaz-Montoya, Jesus G and Zhang, Yuhui and Su, Yuchang and Bhowmik, Disha and Coman, Zachary and others},
  booktitle={2025 IEEE/CVF Conference on Computer Vision and Pattern Recognition (CVPR)},
  pages={19553--19564},
  year={2025},
  organization={IEEE}
}

@inproceedings{zhang2025mavis,
  title={Mavis: Mathematical visual instruction tuning with an automatic data engine},
  author={Zhang, Renrui and Wei, Xinyu and Jiang, Dongzhi and Guo, Ziyu and Zhang, Yichi and Tong, Chengzhuo and Liu, Jiaming and Zhou, Aojun and Zhang, Shanghang and Peng, Gao and others},
  booktitle={International Conference on Learning Representations},
  volume={2025},
  pages={87955--87989},
  year={2025}
}

@inproceedings{masry2022chartqa,
  title={Chartqa: A benchmark for question answering about charts with visual and logical reasoning},
  author={Masry, Ahmed and Tan, Jia Qing and Joty, Shafiq and Hoque, Enamul and others},
  booktitle={Findings of the association for computational linguistics: ACL 2022},
  pages={2263--2279},
  year={2022}
}

@inproceedings{xu2025chartmoe,
  title={Chartmoe: Mixture of diversely aligned expert connector for chart understanding},
  author={Xu, Zhengzhuo and Qu, Bowen and Qi, Yiyan and Du, Sinan and Xu, Chengjin and Yuan, Chun and Guo, Jian},
  booktitle={International Conference on Learning Representations},
  volume={2025},
  pages={78550--78572},
  year={2025}
}

@inproceedings{masry2025chartqapro,
  title={Chartqapro: A more diverse and challenging benchmark for chart question answering},
  author={Masry, Ahmed and Islam, Mohammed Saidul and Ahmed, Mahir and Bajaj, Aayush and Kabir, Firoz and Kartha, Aaryaman and Laskar, Md Tahmid Rahman and Rahman, Mizanur and Rahman, Shadikur and Shahmohammadi, Mehrad and others},
  booktitle={Findings of the Association for Computational Linguistics: ACL 2025},
  pages={19123--19151},
  year={2025}
}

@inproceedings{yang2025chartmimic,
  title={Chartmimic: Evaluating lmm's cross-modal reasoning capability via chart-to-code generation},
  author={Yang, Cheng and Shi, Chufan and Liu, Yaxin and Shui, Bo and Wang, Junjie and Jing, Mohan and Xu, Linran and Zhu, Xinyu and Li, Siheng and Zhang, Yuxiang and others},
  booktitle={International Conference on Learning Representations},
  volume={2025},
  pages={26590--26646},
  year={2025}
}

@article{zhang2024multimodal,
  title={Multimodal Chain-of-Thought Reasoning in Language Models},
  author={Zhang, Zhuosheng and Zhang, Aston and Li, Mu and Zhao, Hai and Karypis, George and Smola, Alex},
  journal={Transactions on Machine Learning Research},
  volume={2024},
  year={2024},
  publisher={Transactions on Machine Learning Research}
}

@article{cheng2026visual,
  title={Visual thoughts: A unified perspective of understanding multimodal chain-of-thought},
  author={Cheng, Zihui and Chen, Qiguang and Xu, Xiao and Wang, Jiaqi and Wang, Weiyun and Fei, Hao and Wang, Yidong and Wang, Alex Jinpeng and Chen, Zhi and Che, Wanxiang and others},
  journal={Advances in Neural Information Processing Systems},
  volume={38},
  pages={96084--96112},
  year={2026}
}

@inproceedings{dong2026revealing,
  title={Revealing the Seen, Imagining the Beyond: A Survey of Image-Grounded Chain-of-Thought Reasoning in Multimodal LLMs},
  author={Dong, Qihua and Zhang, Yitian and Zeng, Huimin and Wang, Yizhou and Lu, Jianglin and Yang, Kuo and Fu, Yun},
  booktitle={Proceedings of the 64th Annual Meeting of the Association for Computational Linguistics (Volume 1: Long Papers)},
  pages={45055--45070},
  year={2026}
}

@article{wang2025multimodal,
  title={Multimodal chain-of-thought reasoning: A comprehensive survey},
  author={Wang, Yaoting and Wu, Shengqiong and Zhang, Yuecheng and Yan, Shuicheng and Liu, Ziwei and Luo, Jiebo and Fei, Hao},
  journal={arXiv preprint arXiv:2503.12605},
  year={2025}
}

@article{zheng2023ddcot,
  title={Ddcot: Duty-distinct chain-of-thought prompting for multimodal reasoning in language models},
  author={Zheng, Ge and Yang, Bin and Tang, Jiajin and Zhou, Hong-Yu and Yang, Sibei},
  journal={Advances in Neural Information Processing Systems},
  volume={36},
  pages={5168--5191},
  year={2023}
}

@inproceedings{mitra2024compositional,
  title={Compositional chain-of-thought prompting for large multimodal models},
  author={Mitra, Chancharik and Huang, Brandon and Darrell, Trevor and Herzig, Roei},
  booktitle={2024 IEEE/CVF Conference on Computer Vision and Pattern Recognition (CVPR)},
  pages={14420--14431},
  year={2024},
  organization={IEEE}
}

@article{shao2024visual,
  title={Visual cot: Advancing multi-modal language models with a comprehensive dataset and benchmark for chain-of-thought reasoning},
  author={Shao, Hao and Qian, Shengju and Xiao, Han and Song, Guanglu and Zong, Zhuofan and Wang, Letian and Liu, Yu and Li, Hongsheng},
  journal={Advances in Neural Information Processing Systems},
  volume={37},
  pages={8612--8642},
  year={2024}
}

@inproceedings{tan2024boosting,
  title={Boosting the power of small multimodal reasoning models to match larger models with self-consistency training},
  author={Tan, Cheng and Wei, Jingxuan and Gao, Zhangyang and Sun, Linzhuang and Li, Siyuan and Guo, Ruifeng and Yu, Bihui and Li, Stan Z},
  booktitle={European Conference on Computer Vision},
  pages={305--322},
  year={2024},
  organization={Springer}
}

@inproceedings{he2024multi,
  title={Multi-modal latent space learning for chain-of-thought reasoning in language models},
  author={He, Liqi and Li, Zuchao and Cai, Xiantao and Wang, Ping},
  booktitle={Proceedings of the AAAI conference on artificial intelligence},
  volume={38},
  number={16},
  pages={18180--18187},
  year={2024}
}

@article{wei2022chain,
  title={Chain-of-thought prompting elicits reasoning in large language models},
  author={Wei, Jason and Wang, Xuezhi and Schuurmans, Dale and Bosma, Maarten and Xia, Fei and Chi, Ed and Le, Quoc V and Zhou, Denny and others},
  journal={Advances in neural information processing systems},
  volume={35},
  pages={24824--24837},
  year={2022}
}

@article{wang2022self,
  title={Self-consistency improves chain of thought reasoning in language models},
  author={Wang, Xuezhi and Wei, Jason and Schuurmans, Dale and Le, Quoc and Chi, Ed and Narang, Sharan and Chowdhery, Aakanksha and Zhou, Denny},
  journal={arXiv preprint arXiv:2203.11171},
  year={2022}
}

@article{chen2023see,
  title={See, think, confirm: Interactive prompting between vision and language models for knowledge-based visual reasoning},
  author={Chen, Zhenfang and Zhou, Qinhong and Shen, Yikang and Hong, Yining and Zhang, Hao and Gan, Chuang},
  journal={arXiv preprint arXiv:2301.05226},
  year={2023}
}

@article{meng2023chain,
  title={Chain of images for intuitively reasoning},
  author={Meng, Fanxu and Yang, Haotong and Wang, Yiding and Zhang, Muhan},
  journal={arXiv preprint arXiv:2311.09241},
  year={2023}
}

@article{hu2024visual,
  title={Visual sketchpad: Sketching as a visual chain of thought for multimodal language models},
  author={Hu, Yushi and Shi, Weijia and Fu, Xingyu and Roth, Dan and Ostendorf, Mari and Zettlemoyer, Luke and Smith, Noah A and Krishna, Ranjay},
  journal={Advances in Neural Information Processing Systems},
  volume={37},
  pages={139348--139379},
  year={2024}
}

@article{lu2023chameleon,
  title={Chameleon: Plug-and-play compositional reasoning with large language models},
  author={Lu, Pan and Peng, Baolin and Cheng, Hao and Galley, Michel and Chang, Kai-Wei and Wu, Ying Nian and Zhu, Song-Chun and Gao, Jianfeng},
  journal={Advances in Neural Information Processing Systems},
  volume={36},
  pages={43447--43478},
  year={2023}
}

@inproceedings{you2023idealgpt,
  title={Idealgpt: Iteratively decomposing vision and language reasoning via large language models},
  author={You, Haoxuan and Sun, Rui and Wang, Zhecan and Chen, Long and Wang, Gengyu and Ayyubi, Hammad and Chang, Kai-Wei and Chang, Shih-Fu},
  booktitle={Findings of the Association for Computational Linguistics: EMNLP 2023},
  pages={11289--11303},
  year={2023}
}

@inproceedings{wu2024dettoolchain,
  title={Dettoolchain: A new prompting paradigm to unleash detection ability of mllm},
  author={Wu, Yixuan and Wang, Yizhou and Tang, Shixiang and Wu, Wenhao and He, Tong and Ouyang, Wanli and Torr, Philip and Wu, Jian},
  booktitle={European Conference on Computer Vision},
  pages={164--182},
  year={2024},
  organization={Springer}
}

@inproceedings{jin2026look,
  title={Look Light, Think Heavy: What Multimodal Chain-of-Thought Reasoning Can and Cannot Do},
  author={Jin, Zhuoran and Zhu, Kejian and Yuan, Hongbang and Hao, Yupu and Cao, Pengfei and Chen, Yubo and Liu, Kang and Zhao, Jun},
  booktitle={Proceedings of the 64th Annual Meeting of the Association for Computational Linguistics (Volume 1: Long Papers)},
  pages={8567--8608},
  year={2026}
}

@inproceedings{li2025vocot,
  title={Vocot: Unleashing visually grounded multi-step reasoning in large multi-modal models},
  author={Li, Zejun and Luo, Ruipu and Zhang, Jiwen and Qiu, Minghui and Huang, Xuan-Jing and Wei, Zhongyu},
  booktitle={Proceedings of the 2025 Conference of the Nations of the Americas Chapter of the Association for Computational Linguistics: Human Language Technologies (Volume 1: Long Papers)},
  pages={3769--3798},
  year={2025}
}

@inproceedings{man2025argus,
  title={Argus: Vision-centric reasoning with grounded chain-of-thought},
  author={Man, Yunze and Huang, De-An and Liu, Guilin and Sheng, Shiwei and Liu, Shilong and Gui, Liang-Yan and Kautz, Jan and Wang, Yu-Xiong and Yu, Zhiding},
  booktitle={2025 IEEE/CVF Conference on Computer Vision and Pattern Recognition (CVPR)},
  pages={14268--14280},
  year={2025},
  organization={IEEE}
}

@article{chen2026mint,
  title={Mint-cot: Enabling interleaved visual tokens in mathematical chain-of-thought reasoning},
  author={Chen, Xinyan and Zhang, Renrui and Jiang, Dongzhi and Zhou, Aojun and Yan, Shilin and Lin, Weifeng and Li, Hongsheng},
  journal={Advances in neural information processing systems},
  volume={38},
  pages={69110--69139},
  year={2026}
}

@article{wu2024mind,
  title={Mind's eye of LLMs: visualization-of-thought elicits spatial reasoning in large language models},
  author={Wu, Wenshan and Mao, Shaoguang and Zhang, Yadong and Xia, Yan and Dong, Li and Cui, Lei and Wei, Furu},
  journal={Advances in Neural Information Processing Systems},
  volume={37},
  pages={90277--90317},
  year={2024}
}

@inproceedings{lee2024multimodal,
  title={Multimodal reasoning with multimodal knowledge graph},
  author={Lee, Junlin and Wang, Yequan and Li, Jing and Zhang, Min},
  booktitle={Proceedings of the 62nd Annual Meeting of the Association for Computational Linguistics (Volume 1: Long Papers)},
  pages={10767--10782},
  year={2024}
}

@inproceedings{dong2025insight,
  title={Insight-v: Exploring long-chain visual reasoning with multimodal large language models},
  author={Dong, Yuhao and Liu, Zuyan and Sun, Hai-Long and Yang, Jingkang and Hu, Winston and Rao, Yongming and Liu, Ziwei},
  booktitle={2025 IEEE/CVF Conference on Computer Vision and Pattern Recognition (CVPR)},
  pages={9062--9072},
  year={2025},
  organization={IEEE}
}

@article{tan2026reason,
  title={Reason-rft: Reinforcement fine-tuning for visual reasoning of vision language models},
  author={Tan, Huajie and Ji, Yuheng and Hao, Xiaoshuai and Chen, Xiansheng and Wang, Pengwei and Wang, Zhongyuan and Zhang, Shanghang},
  journal={Advances in neural information processing systems},
  volume={38},
  pages={5772--5822},
  year={2026}
}

@inproceedings{huang2026vision,
  title={Vision-r1: Incentivizing reasoning capability in multimodal large language models},
  author={Huang, Wenxuan and Jia, Bohan and Cao, Shaosheng and Ye, Zheyu and Xu, Zhe and Hu, Yao and Lin, Shaohui and others},
  booktitle={International Conference on Learning Representations},
  volume={2026},
  pages={63794--63812},
  year={2026}
}

@article{zhang2023pmc,
  title={Pmc-vqa: Visual instruction tuning for medical visual question answering},
  author={Zhang, Xiaoman and Wu, Chaoyi and Zhao, Ziheng and Lin, Weixiong and Zhang, Ya and Wang, Yanfeng and Xie, Weidi},
  journal={arXiv preprint arXiv:2305.10415},
  year={2023}
}

@inproceedings{corbiere2026drivingvqa,
  title={Drivingvqa: A dataset for interleaved visual chain-of-thought in real-world driving scenarios},
  author={Corbi{\`e}re, Charles and Roburin, Simon and Montariol, Syrielle and Bosselut, Antoine and Alahi, Alexandre},
  booktitle={Findings of the Association for Computational Linguistics: EACL 2026},
  pages={3309--3333},
  year={2026}
}

@inproceedings{zhu2016visual7w,
  title={Visual7w: Grounded question answering in images},
  author={Zhu, Yuke and Groth, Oliver and Bernstein, Michael and Fei-Fei, Li},
  booktitle={Proceedings of the IEEE conference on computer vision and pattern recognition},
  pages={4995--5004},
  year={2016}
}

@inproceedings{kembhavi2016diagram,
  title={A diagram is worth a dozen images},
  author={Kembhavi, Aniruddha and Salvato, Mike and Kolve, Eric and Seo, Minjoon and Hajishirzi, Hannaneh and Farhadi, Ali},
  booktitle={European conference on computer vision},
  pages={235--251},
  year={2016},
  organization={Springer}
}

@article{tong2024cambrian,
  title={Cambrian-1: A fully open, vision-centric exploration of multimodal llms},
  author={Tong, Shengbang and Brown, Ellis and Wu, Penghao and Woo, Sanghyun and Middepogu, Manoj and Akula, Sai C and Yang, Jihan and Yang, Shusheng and Iyer, Adithya and Pan, Xichen and others},
  journal={Advances in Neural Information Processing Systems},
  volume={37},
  pages={87310--87356},
  year={2024}
}

@inproceedings{chen2021geoqa,
  title={Geoqa: A geometric question answering benchmark towards multimodal numerical reasoning},
  author={Chen, Jiaqi and Tang, Jianheng and Qin, Jinghui and Liang, Xiaodan and Liu, Lingbo and Xing, Eric and Lin, Liang},
  booktitle={Findings of the Association for Computational Linguistics: ACL-IJCNLP 2021},
  pages={513--523},
  year={2021}
}

@inproceedings{yue2024mmmu,
  title={Mmmu: A massive multi-discipline multimodal understanding and reasoning benchmark for expert agi},
  author={Yue, Xiang and Ni, Yuansheng and Zhang, Kai and Zheng, Tianyu and Liu, Ruoqi and Zhang, Ge and Stevens, Samuel and Jiang, Dongfu and Ren, Weiming and Sun, Yuxuan and others},
  booktitle={Proceedings of the IEEE/CVF conference on computer vision and pattern recognition},
  pages={9556--9567},
  year={2024}
}

@article{bai2025qwen2,
  title={Qwen2.5-vl technical report},
  author={Bai, Shuai and Chen, Keqin and Liu, Xuejing and Wang, Jialin and Ge, Wenbin and Song, Sibo and Dang, Kai and Wang, Peng and Wang, Shijie and Tang, Jun and others},
  journal={arXiv preprint arXiv:2502.13923},
  year={2025}
}

@misc{bai2025qwen3vltechnicalreport,
      title={Qwen3-VL Technical Report}, 
      author={Shuai Bai and Yuxuan Cai and Ruizhe Chen and Keqin Chen and Xionghui Chen and Zesen Cheng and Lianghao Deng and Wei Ding and Chang Gao and Chunjiang Ge and Wenbin Ge and others},
      year={2025},
      eprint={2511.21631},
      archivePrefix={arXiv},
      primaryClass={cs.CV},
      url={https://arxiv.org/abs/2511.21631}, 
}

@misc{qwen3.5,
    title  = {{Qwen3.5}: Towards Native Multimodal Agents},
    author = {{Qwen Team}},
    month  = {February},
    year   = {2026},
    url    = {https://qwen.ai/blog?id=qwen3.5}
}

@misc{qwen3.6-27b,
  title  = {{Qwen3.6-27B}: Flagship-Level Coding in a {27B} Dense Model},
  author = {{Qwen Team}},
  year   = {2026},
  month  = {April},
  url    = {https://qwen.ai/blog?id=qwen3.6-27b}
}

@misc{qwen3.6-35b-a3b,
  title  = {{Qwen3.6-35B-A3B}: Agentic Coding Power, Now Open to All},
  author = {{Qwen Team}},
  year   = {2026},
  month  = {April},
  url    = {https://qwen.ai/blog?id=qwen3.6-35b-a3b}
}

@misc{qwen36plus,
    title = {{Qwen3.6-Plus}: Towards Real World Agents},
    url = {https://qwen.ai/blog?id=qwen3.6},
    author = {{Qwen Team}},
    month = {April},
    year = {2026}
}

@misc{qwen37,
    title = {{Qwen3.7}: The Agent Frontier},
    url = {https://qwen.ai/blog?id=qwen3.7},
    author = {{Qwen Team}},
    month = {May},
    year = {2026}
}

@misc{qwen3.8,
    title  = {{Qwen3.8-Max}: A New Bar for Coding and Cowork},
    author = {{Qwen Team}},
    year   = {2026},
    month  = {August},
    url    = {https://qwen.ai/blog?id=qwen3.8}
}

@misc{bai2023qwenvlversatilevisionlanguagemodel,
      title={Qwen-VL: A Versatile Vision-Language Model for Understanding, Localization, Text Reading, and Beyond}, 
      author={Jinze Bai and Shuai Bai and Shusheng Yang and Shijie Wang and Sinan Tan and Peng Wang and Junyang Lin and Chang Zhou and Jingren Zhou},
      year={2023},
      eprint={2308.12966},
      archivePrefix={arXiv},
      primaryClass={cs.CV},
      url={https://arxiv.org/abs/2308.12966}, 
}

@misc{lai2026minimaxsparseattention,
      title={MiniMax Sparse Attention}, 
      author={Xunhao Lai and Weiqi Xu and Yufeng Yang and Qiaorui Chen and Yang Xu and Lunbin Zeng and Xiaolong Li and Haohai Sun and Haichao Zhu and Vito Zhang and Jinkai Hu and Jiayao Li and Rui Gao and Zekun Li and Songquan Zhu and Jingkai Zhou and Pengyu Zhao},
      year={2026},
      eprint={2606.13392},
      archivePrefix={arXiv},
      primaryClass={cs.AI},
      url={https://arxiv.org/abs/2606.13392}, 
}

@misc{vteam2026glm45vglm41vthinkingversatilemultimodal,
      title={GLM-4.5V and GLM-4.1V-Thinking: Towards Versatile Multimodal Reasoning with Scalable Reinforcement Learning}, 
      author={V Team and Wenyi Hong and Wenmeng Yu and Xiaotao Gu and Guo Wang and Guobing Gan and Haomiao Tang and Jiale Cheng and Ji Qi and Junhui Ji and Lihang Pan and Shuaiqi Duan and others},
      year={2026},
      eprint={2507.01006},
      archivePrefix={arXiv},
      primaryClass={cs.CV},
      url={https://arxiv.org/abs/2507.01006}, 
}

@misc{vteam2026glm5vturbonativefoundationmodel,
      title={GLM-5V-Turbo: Toward a Native Foundation Model for Multimodal Agents}, 
      author={V Team and Wenyi Hong and Xiaotao Gu and Ziyang Pan and Zhen Yang and Yuting Wang and Yue Wang and Yuanchang Yue and Yu Wang and Yanling Wang and Yan Wang and Xijun Liu and others},
      year={2026},
      eprint={2604.26752},
      archivePrefix={arXiv},
      primaryClass={cs.CV},
      url={https://arxiv.org/abs/2604.26752}, 
}

@misc{xai2026grok43,
  author       = {{xAI}},
  title        = {Grok 4.3},
  year         = {2026},
  howpublished = {\url{https://docs.x.ai/developers/models/grok-4.3}}
}

@misc{xai2026grok45,
  author       = {{xAI}},
  title        = {Introducing Grok 4.5},
  year         = {2026},
  month        = jul,
  howpublished = {\url{https://x.ai/news/grok-4-5}}
}

@misc{xai2026grok46,
  author       = {{xAI}},
  title        = {Introducing Grok 4.6},
  year         = {2026},
  month        = aug,
  howpublished = {\url{https://x.ai/news/grok-4-6}},
}

@misc{openai2023gpt4v,
  author       = {{OpenAI}},
  title        = {GPT-4V(ision) System Card},
  year         = {2023},
  month        = sep,
  howpublished = {\url{https://openai.com/index/gpt-4v-system-card/}},
  note         = {Published September 25, 2023}
}

@misc{openai2024gpt4ocard,
      title={GPT-4o System Card}, 
      author={{OpenAI}},
      year={2024},
      eprint={2410.21276},
      archivePrefix={arXiv},
      primaryClass={cs.CL},
      url={https://arxiv.org/abs/2410.21276}, 
}

@misc{singh2026openaigpt5card,
      title={OpenAI GPT-5 System Card}, 
      author={{OpenAI}},
      year={2026},
      eprint={2601.03267},
      archivePrefix={arXiv},
      primaryClass={cs.CL},
      url={https://arxiv.org/abs/2601.03267}, 
}

@misc{openai2026gpt54,
  author       = {{OpenAI}},
  title        = {GPT-5.4 Thinking System Card},
  year         = {2026},
  month        = mar,
  howpublished = {\url{https://openai.com/index/gpt-5-4-thinking-system-card/}},
  note         = {Published March 5, 2026}
}

@misc{openai2026gpt55,
  author       = {{OpenAI}},
  title        = {GPT-5.5 System Card},
  year         = {2026},
  month        = apr,
  howpublished = {\url{https://openai.com/index/gpt-5-5-system-card/}},
  note         = {Published April 23, 2026}
}

@misc{openai2026gpt56,
  author       = {{OpenAI}},
  title        = {GPT-5.6 System Card},
  year         = {2026},
  month        = jul,
  howpublished = {\url{https://deploymentsafety.openai.com/gpt-5-6}},
  note         = {Published July 9, 2026}
}

@misc{anthropic2025haiku45,
  author       = {{Anthropic}},
  title        = {Claude Haiku 4.5},
  year         = {2025},
  month        = oct,
  howpublished = {\url{https://www.anthropic.com/news/claude-haiku-4-5}}
}

@misc{anthropic2026sonnet46,
  author       = {{Anthropic}},
  title        = {Claude Sonnet 4.6},
  year         = {2026},
  month        = feb,
  howpublished = {\url{https://www.anthropic.com/news/claude-sonnet-4-6}}
}

@misc{anthropic2026sonnet5,
  author       = {{Anthropic}},
  title        = {Claude Sonnet 5},
  year         = {2026},
  month        = jun,
  howpublished = {\url{https://www.anthropic.com/news/claude-sonnet-5}}
}

@misc{anthropic2026opus48,
  author       = {{Anthropic}},
  title        = {Claude Opus 4.8},
  year         = {2026},
  month        = may,
  howpublished = {\url{https://www.anthropic.com/news/claude-opus-4-8}}
}

@misc{anthropic2026opus5,
  author       = {{Anthropic}},
  title        = {Claude Opus 5},
  year         = {2026},
  month        = jul,
  howpublished = {\url{https://www.anthropic.com/news/claude-opus-5}}
}

@misc{bytedance2025seed16,
  author       = {{ByteDance Seed}},
  title        = {Introduction to Techniques Used in Seed1.6},
  year         = {2025},
  month        = jun,
  howpublished = {\url{https://seed.bytedance.com/en/blog/introduction-to-techniques-used-in-seed1-6}},
  note         = {Published June 25, 2025}
}

@misc{seed2026seed18modelcardgeneralized,
      title={Seed1.8 Model Card: Towards Generalized Real-World Agency}, 
      author={Bytedance Seed},
      year={2026},
      eprint={2603.20633},
      archivePrefix={arXiv},
      primaryClass={cs.AI},
      url={https://arxiv.org/abs/2603.20633}, 
}

@misc{seed2026seed20modelcardintelligence,
      title={Seed2.0 Model Card: Towards Intelligence Frontier for Real-World Complexity}, 
      author={Bytedance Seed},
      year={2026},
      eprint={2607.00248},
      archivePrefix={arXiv},
      primaryClass={cs.AI},
      url={https://arxiv.org/abs/2607.00248}, 
}

@misc{bytedance2026seed21,
  author       = {{ByteDance Seed}},
  title        = {Seed2.1: A Next-Generation Agent for Real-World Productivity},
  year         = {2026},
  month        = jun,
  howpublished = {\url{https://seed.bytedance.com/en/seed2_1}}
}

@misc{moonshot2026kimik26,
  author       = {{Moonshot AI}},
  title        = {Kimi K2.6},
  year         = {2026},
  howpublished = {\url{https://www.kimi.com/ai-models/kimi-k2-6}}
}

@misc{kimiteam2026kimik3openfrontier,
      title={Kimi K3: Open Frontier Intelligence}, 
      author={{Kimi Team}},
      year={2026},
      eprint={2607.24653},
      archivePrefix={arXiv},
      primaryClass={cs.CL},
      url={https://arxiv.org/abs/2607.24653}, 
}

@misc{google2025gemini3flash,
  author       = {{Google DeepMind}},
  title        = {Gemini 3 Flash Model Card},
  year         = {2025},
  month        = dec,
  howpublished = {\url{https://deepmind.google/models/model-cards/gemini-3-flash/}},
  note         = {Published December 17, 2025}
}

@misc{google2026gemini35flash,
  author       = {{Google DeepMind}},
  title        = {Gemini 3.5 Flash Model Card},
  year         = {2026},
  month        = may,
  howpublished = {\url{https://deepmind.google/models/model-cards/gemini-3-5-flash/}},
  note         = {Published May 19, 2026}
}

@inproceedings{yu2019deep,
  title={Deep modular co-attention networks for visual question answering},
  author={Yu, Zhou and Yu, Jun and Cui, Yuhao and Tao, Dacheng and Tian, Qi},
  booktitle={Proceedings of the IEEE/CVF Conference on Computer Vision and Pattern Recognition (CVPR)},
  pages={6281--6290},
  year={2019}
}

@inproceedings{anderson2018bottom,
  title={Bottom-up and top-down attention for image captioning and visual question answering},
  author={Anderson, Peter and He, Xiaodong and Buehler, Chris and Teney, Damien and Johnson, Mark and Gould, Stephen and Zhang, Lei},
  booktitle={Proceedings of the IEEE/CVF Conference on Computer Vision and Pattern Recognition (CVPR)},
  pages={6077--6086},
  year={2018}
}

@article{kim2018bilinear,
  title={Bilinear attention networks},
  author={Kim, Jin-Hwa and Jun, Jaehyun and Zhang, Byoung-Tak},
  journal={Advances in Neural Information Processing Systems (NeurIPS)},
  volume={31},
  year={2018}
}

@inproceedings{gao2019dynamic,
  title={Dynamic fusion with intra-and inter-modality attention flow for visual question answering},
  author={Gao, Peng and Jiang, Zhengkai and You, Haoxuan and Lu, Pan and Hoi, Steven CH and Wang, Xiaogang and Li, Hongsheng},
  booktitle={Proceedings of the IEEE/CVF Conference on Computer Vision and Pattern Recognition (CVPR)},
  pages={6639--6648},
  year={2019}
}

@inproceedings{kim2021vilt,
  title={Vilt: Vision-and-language transformer without convolution or region supervision},
  author={Kim, Wonjae and Son, Bokyung and Kim, Ildoo},
  booktitle={Proceedings of the International Conference on Machine Learning (ICML)},
  pages={5583--5594},
  year={2021},
  organization={PMLR}
}

@article{lu2021iconqa,
  title={Iconqa: A new benchmark for abstract diagram understanding and visual language reasoning},
  author={Lu, Pan and Qiu, Liang and Chen, Jiaqi and Xia, Tony and Zhao, Yizhou and Zhang, Wei and Yu, Zhou and Liang, Xiaodan and Zhu, Song-Chun},
  journal={arXiv preprint arXiv:2110.13214},
  year={2021}
}

@inproceedings{li2020does,
  title={What does BERT with vision look at?},
  author={Li, Liunian Harold and Yatskar, Mark and Yin, Da and Hsieh, Cho-Jui and Chang, Kai-Wei},
  booktitle={Proceedings of the Association for Computational Linguistics (ACL)},
  pages={5265--5275},
  year={2020}
}

@article{achiam2023gpt,
  title={Gpt-4 technical report},
  author={Achiam, Josh and Adler, Steven and Agarwal, Sandhini and Ahmad, Lama and Akkaya, Ilge and Aleman, Florencia Leoni and Almeida, Diogo and Altenschmidt, Janko and Altman, Sam and Anadkat, Shyamal and others},
  journal={arXiv preprint arXiv:2303.08774},
  year={2023}
}

@article{team2023gemini,
  title={Gemini: a family of highly capable multimodal models},
  author={Team, Gemini and Anil, Rohan and Borgeaud, Sebastian and Alayrac, Jean-Baptiste and Yu, Jiahui and Soricut, Radu and Schalkwyk, Johan and Dai, Andrew M and Hauth, Anja and Millican, Katie and others},
  journal={arXiv preprint arXiv:2312.11805},
  year={2023}
}

@article{tan2024retrieval,
  title={Retrieval meets reasoning: Even high-school textbook knowledge benefits multimodal reasoning},
  author={Tan, Cheng and Wei, Jingxuan and Sun, Linzhuang and Gao, Zhangyang and Li, Siyuan and Yu, Bihui and Guo, Ruifeng and Li, Stan Z},
  journal={arXiv preprint arXiv:2405.20834},
  year={2024}
}

@article{zhang2023llama,
  title={Llama-adapter: Efficient fine-tuning of language models with zero-init attention},
  author={Zhang, Renrui and Han, Jiaming and Liu, Chris and Gao, Peng and Zhou, Aojun and Hu, Xiangfei and Yan, Shilin and Lu, Pan and Li, Hongsheng and Qiao, Yu},
  journal={arXiv preprint arXiv:2303.16199},
  year={2023}
}

@inproceedings{liu2024improved,
  title={Improved baselines with visual instruction tuning},
  author={Liu, Haotian and Li, Chunyuan and Li, Yuheng and Lee, Yong Jae},
  booktitle={Proceedings of the IEEE/CVF Conference on Computer Vision and Pattern Recognition (CVPR)},
  pages={26296--26306},
  year={2024}
}

@article{luo2023cheap,
  title={Cheap and quick: Efficient vision-language instruction tuning for large language models},
  author={Luo, Gen and Zhou, Yiyi and Ren, Tianhe and Chen, Shengxin and Sun, Xiaoshuai and Ji, Rongrong},
  journal={Advances in Neural Information Processing Systems (NeurIPS)},
  volume={36},
  pages={29615--29627},
  year={2023}
}

@article{khashabi2020unifiedqa,
  title={Unifiedqa: Crossing format boundaries with a single qa system},
  author={Khashabi, Daniel and Min, Sewon and Khot, Tushar and Sabharwal, Ashish and Tafjord, Oyvind and Clark, Peter and Hajishirzi, Hannaneh},
  journal={arXiv preprint arXiv:2005.00700},
  year={2020}
}

@article{shen2023hugginggpt,
  title={Hugginggpt: Solving ai tasks with chatgpt and its friends in hugging face},
  author={Shen, Yongliang and Song, Kaitao and Tan, Xu and Li, Dongsheng and Lu, Weiming and Zhuang, Yueting},
  journal={Advances in Neural Information Processing Systems (NeurIPS)},
  volume={36},
  pages={38154--38180},
  year={2023}
}

@article{wu2023visual,
  title={Visual chatgpt: Talking, drawing and editing with visual foundation models},
  author={Wu, Chenfei and Yin, Shengming and Qi, Weizhen and Wang, Xiaodong and Tang, Zecheng and Duan, Nan},
  journal={arXiv preprint arXiv:2303.04671},
  year={2023}
}

@article{peng2023kosmos,
  title={Kosmos-2: Grounding multimodal large language models to the world},
  author={Peng, Zhiliang and Wang, Wenhui and Dong, Li and Hao, Yaru and Huang, Shaohan and Ma, Shuming and Wei, Furu},
  journal={arXiv preprint arXiv:2306.14824},
  year={2023}
}

@article{dai2023instructblip,
  title={Instructblip: Towards general-purpose vision-language models with instruction tuning},
  author={Dai, Wenliang and Li, Junnan and Li, Dongxu and Tiong, Anthony and Zhao, Junqi and Wang, Weisheng and Li, Boyang and Fung, Pascale N and Hoi, Steven},
  journal={Advances in Neural Information Processing Systems (NeurIPS)},
  volume={36},
  pages={49250--49267},
  year={2023}
}

@article{wang2024cogvlm,
  title={Cogvlm: Visual expert for pretrained language models},
  author={Wang, Weihan and Lv, Qingsong and Yu, Wenmeng and Hong, Wenyi and Qi, Ji and Wang, Yan and Ji, Junhui and Yang, Zhuoyi and Zhao, Lei and XiXuan, Song and others},
  journal={Advances in Neural Information Processing Systems (NeurIPS)},
  volume={37},
  pages={121475--121499},
  year={2024}
}

@inproceedings{giannou2023looped,
  title={Looped transformers as programmable computers},
  author={Giannou, Angeliki and Rajput, Shashank and Sohn, Jy-yong and Lee, Kangwook and Lee, Jason D and Papailiopoulos, Dimitris},
  booktitle={International Conference on Machine Learning},
  pages={11398--11442},
  year={2023},
  organization={PMLR}
}
\bibliographystyle{iclr2027_conference}

\newpage

\appendix

\section{Detailed Construction for VersaCoT-10K}
\label{sec:a-detailed-construction-for-versaCoT-10K}
As shown in \cref{figs:fig-03}, VersaCoT-10K is constructed through three consecutive stages: \textit{Data Collection and Standardization}, \textit{CoT Annotation and Quality Control}, and \textit{Reasoning-State Augmentation}. We aim to integrate reasoning-intensive samples spanning diverse visual content, domain knowledge, and reasoning patterns into a unified single-image multiple-choice format, providing high-quality supervision for multi-domain reasoning and reasoning-state refinement. For each of the eight heterogeneous visual reasoning domains, we retain 1,250 samples and split them into training, validation, and test sets at a 7:1:2 ratio. VersaCoT-10K therefore contains 10,000 samples in total, including 7,000 for training, 1,000 for validation, and 2,000 for testing. The following subsections detail each stage of the construction pipeline.

\subsection{Data Collection and Standardization}
To ensure broad coverage of multimodal reasoning, we collect reasoning-intensive samples with explicit visual inputs and valid answer annotations from multiple public benchmarks. According to their primary visual content and reasoning characteristics, these samples are organized into eight heterogeneous visual reasoning domains: medical reasoning, driving and safety, visual commonsense, scientific diagram, spatial perception, geometric and mathematical, chart computation, and multidisciplinary knowledge. The corresponding source datasets are PMC-VQA~\citep{zhang2023pmc}, DrivingVQA~\citep{corbiere2026drivingvqa}, Visual7W~\citep{zhu2016visual7w}, AI2D~\citep{kembhavi2016diagram}, CV-Bench~\citep{tong2024cambrian}, GeoQA~\citep{chen2021geoqa}, ChartQA~\citep{masry2022chartqa}, and MMMU~\citep{yue2024mmmu}, respectively. Together, these datasets provide diverse visual evidence, domain knowledge, and reasoning patterns. Because the source datasets differ substantially in task formulation, answer representation, and annotation structure, we further standardize the collected samples into a unified single-image multiple-choice format. As illustrated in \cref{figs:fig-03}, this process consists of three steps: \textit{Option Construction}, \textit{Answer Normalization}, and \textit{Input Format Organization}.

\textbf{Option Construction.} For source datasets that already adopt a multiple-choice format, we preserve the original candidate options whenever possible while standardizing their textual representation. For datasets originally formulated as open-ended question answering, we construct discriminative distractors from the available annotations and associated structured information, thereby converting them into a unified multiple-choice format. For ChartQA in particular, originally free-form Yes/No questions are directly converted into binary-choice questions. For questions whose answers correspond to textual entries, years, or numerical values, distractors are preferentially sampled from the structured table associated with the chart. For numerical questions, we first select table values with the same numerical format as the correct answer. When an insufficient number of valid candidates is available, limited deterministic numerical perturbations are applied to construct additional distractors. Samples are discarded when reliable and discriminative incorrect options cannot be constructed. The resulting candidate set for sample $i$ is represented as
\begin{gather}
{\cal O}_i = \{o_{i,1}, \ldots, o_{i,n_i}\},
\end{gather}
where $n_i$ denotes the number of candidate options. We preserve the original variation in candidate-set size across source datasets rather than forcing all samples into a fixed answer space.

\textbf{Answer Normalization.} The source datasets encode correct answers in heterogeneous forms, including letter labels, numerical indices, and text strings. To establish a unified training and evaluation interface, we normalize the correct answer of each sample to its zero-based index in the standardized candidate set:
\begin{gather}
{\cal A}_i \in \{0, \ldots, n_i - 1\}.
\end{gather}
For answers represented by letter or numerical labels, we first recover their positions according to the original option ordering and then map them to the standardized candidate set. Textual answers are matched against the standardized options before being converted into the corresponding indices. We additionally verify that each answer uniquely corresponds to exactly one candidate option, ensuring that every retained sample has a unique and valid ground-truth answer.

\begin{figure*}[t]
    \centering
    \includegraphics[width=\linewidth]{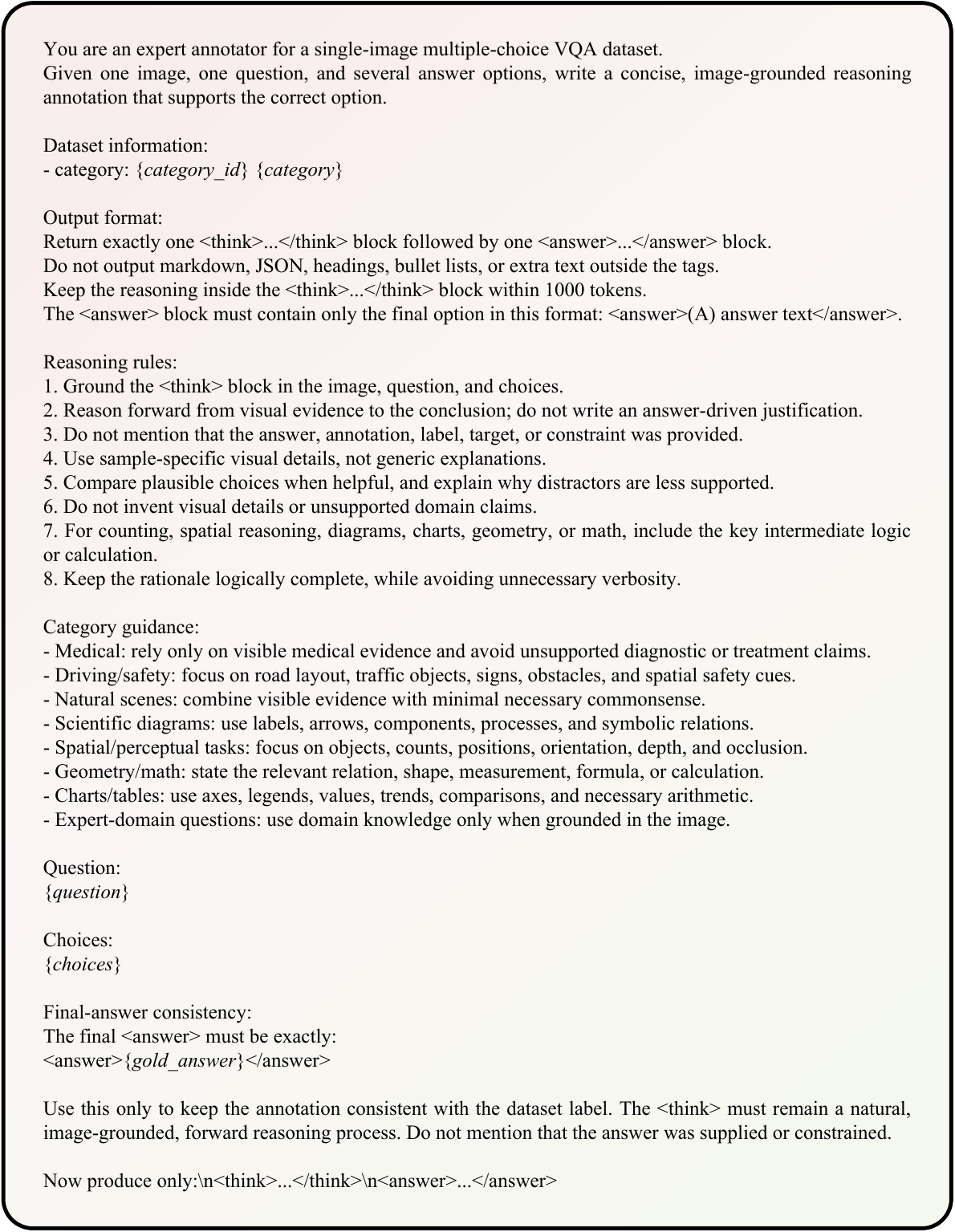}
    \vspace{-0.5cm}
    \caption{Prompt Design for CoT Annotation during the construction of the VersaCoT-10K dataset. Italicized text represents assignment variables.}
    \label{figs:fig-s-prompt-cg}
    \vspace{-0.4cm}
\end{figure*}

\begin{figure*}[!t]
    \centering
    \includegraphics[width=\linewidth]{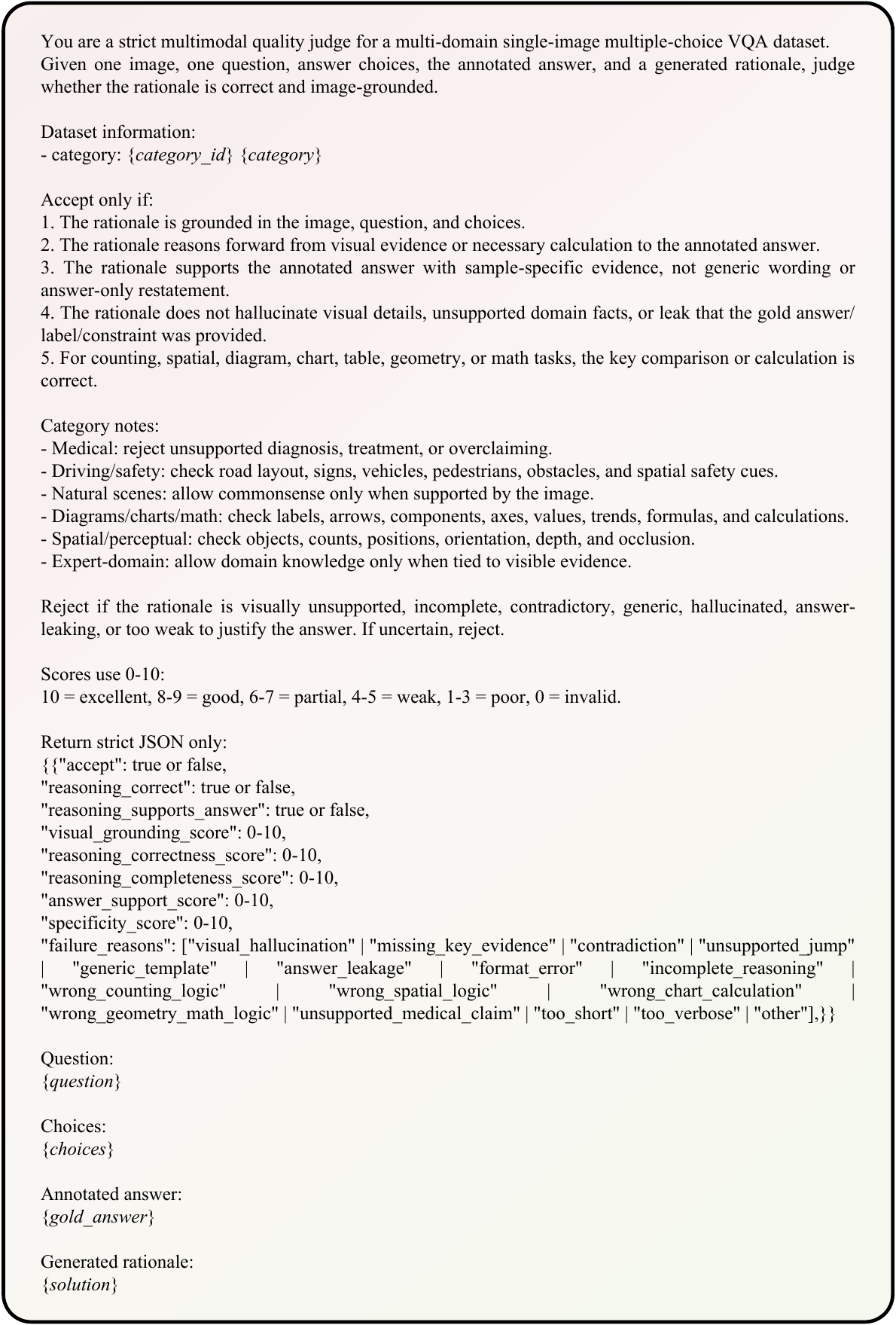}
    \vspace{-0.5cm}
    \caption{Prompt Design for CoT filtering during the construction of the VersaCoT-10K dataset. Italicized text represents assignment variables. A quality score of 8 or higher is deemed acceptable.}
    \label{figs:fig-s-prompt-cf}
    \vspace{-0.3cm}
\end{figure*}

\textbf{Input Format Organization.} After option construction and answer normalization, we further standardize the multimodal input structure across datasets. Each sample $i$ is represented as
\begin{gather}
{\cal D}_i =
\{
{\cal I}_i,\;
{\cal Q}_i,\;
{\cal O}_i,\;
{\cal C}_i,\;
{\cal A}_i
\},
\end{gather}
where ${\cal I}_i$, ${\cal Q}_i$, ${\cal O}_i$, ${\cal C}_i$, and ${\cal A}_i$ denote the input image, question, candidate-option set, image caption, and correct answer index, respectively. We standardize image storage, question fields, and sample identifiers, and retain only samples compatible with single-image reasoning. Additional task-specific filtering is applied when necessary. For Visual7W, we retain only samples from the \textit{telling} subset that are compatible with text-based multiple-choice reasoning. For MMMU we retain only samples satisfying the single-image multiple-choice requirement. We also preserve metadata such as the source dataset and original split to support subsequent partitioning and statistical analysis. In addition, for source datasets with predefined training and test partitions, we maintain source-level split consistency during reorganization to avoid mixing samples from their original training and test sets. Through these steps, heterogeneous source datasets are transformed into a unified single-image multiple-choice format for subsequent CoT annotation and reasoning-state construction.

\subsection{CoT Annotation and Quality Control}
After data standardization, we construct explicit reasoning annotations through \textit{CoT Annotation and Quality Control}. Since most source visual question answering datasets do not provide high-quality explicit CoTs, they cannot directly support reasoning-process learning and subsequent reasoning-state construction. We therefore employ an automated MLLM-assisted annotation pipeline to generate visually grounded reasoning trajectories and apply strict quality control to retain reliable annotations. Before annotation, samples within each domain-specific data pool are randomly shuffled to reduce the dependence of final sample selection on the original dataset ordering.

\textbf{CoT Annotation.}
Manually annotating reasoning trajectories for 10K samples spanning multiple visual reasoning domains would require substantial human effort. We therefore employ Gemini-3-Flash~\citep{google2025gemini3flash} as the CoT generator. Given the task information and a predefined prompt, the model produces a reasoning trajectory ${\cal R}_i$ together with a predicted answer $\hat{\cal A}_i$:
\begin{gather}
({\cal R}_i,\hat{\cal A}_i)
=
{\cal M}_{gen}({\cal D}_i,{\cal P}_{gen}),
\end{gather}
where ${\cal P}_{gen}$ denotes the CoT generation prompt, provided in full in \cref{figs:fig-s-prompt-cg}.

To construct label-consistent and visually grounded reasoning annotations, ${\cal P}_{gen}$ provides the source benchmark answer as the target conclusion and explicitly constrains how that conclusion should be justified. In particular, the generated rationale must establish a step-by-step reasoning path from concrete visual evidence and task-relevant information, without referring to or revealing the provided answer, gold label, dataset annotation, or any equivalent supervision signal. It must also avoid introducing visual evidence unsupported by the input image. For reasoning-intensive tasks such as counting, spatial reasoning, chart understanding, scientific diagrams, geometry, and mathematics, the rationale is further required to include the key comparisons, relational judgments, or intermediate calculations necessary to support the final conclusion.

To standardize the annotation format, the generator is required to produce outputs following
\begin{gather}
< think > \;reasoning\;process\; < /think > \;\; < answer > final\;option\; < /answer >
\end{gather}
The generated answer $\hat{\cal A}_i$ is constrained to match the original annotation ${\cal A}^{GT}_i$ for label consistency. We then extract the \texttt{<think>} content as the candidate CoT annotation ${\cal R}_i$.

\textbf{Quality Control.}
All generated CoTs undergo a two-stage quality-control procedure consisting of \textit{rule-based filtering} and \textit{judge-based filtering}. Rule-based filtering first verifies output structure, field validity, answer consistency, and text parsability. We restrict the reasoning length to 200-4,000 characters to remove overly short rationales with insufficient reasoning content as well as abnormally long outputs. We further detect answer-leakage expressions, such as ``provided answer'', ``given answer'', ``gold answer'', ``target answer'', and ``dataset label''. Samples whose rationales explicitly reveal dependence on externally supplied answer annotations are discarded.

Satisfying these structural constraints alone does not guarantee that a generated CoT is visually grounded or logically reliable. We therefore employ Gemini-3.5-Flash~\citep{google2026gemini35flash}, which is separate from the CoT generator, as a multimodal quality judge. The judge temperature is set to 0 to reduce sampling variability. Given the original task information ${\cal D}_i$, the generated reasoning trajectory ${\cal R}_i$, and a predefined filtering prompt ${\cal P}_{filter}$, the judge evaluates the visual grounding, logical validity, and answer-supporting quality of each candidate CoT. Importantly, the judge is used for reasoning-annotation quality control rather than independent answer verification. Since the source benchmarks already provide validated target answers, the judge assesses whether the generated rationale provides a visually grounded and logically coherent justification for the annotated target. In this way, the source benchmark specifies the target answer, while the multimodal judge evaluates the quality of the reasoning supporting that target.

For each candidate, the judge produces three binary quality judgments and five fine-grained scores. The binary judgments indicate whether the rationale is \textit{overall acceptable}, whether the \textit{reasoning is correct}, and whether the \textit{reasoning supports the answer}. The five scores evaluate \textit{visual grounding}, \textit{reasoning correctness}, \textit{reasoning completeness}, \textit{answer support}, and \textit{specificity}, respectively, each on a scale from 0 to 10. For sample $i$, we denote the resulting score vector as
\begin{gather}
{\mathbf s}_i =
[s_i^{vg},s_i^{rc},s_i^{comp},s_i^{as},s_i^{spec}].
\end{gather}
Let $c_i^{acc}$, $c_i^{cor}$, and $c_i^{sup}$ denote the three binary judgments. We retain a sample only when all three judgments are positive and every fine-grained score is at least 8:
\begin{gather}
{\cal D}_{keep}
=
\left\{
{\cal D}_i
\;\middle|\;
c_i^{acc}=c_i^{cor}=c_i^{sup}=1,\;
\min({\mathbf s}_i)\geq 8
\right\}.
\end{gather}
The complete multimodal judging prompt is provided in \cref{figs:fig-s-prompt-cf}. Through this combination of rule-based filtering and multimodal quality assessment, we retain quality-controlled CoT annotations that satisfy stringent criteria for visual grounding, logical coherence, answer support, and specificity, providing reliable supervision for subsequent reasoning-state augmentation.

\begin{figure*}[!t]
    \centering
    \includegraphics[width=\linewidth]{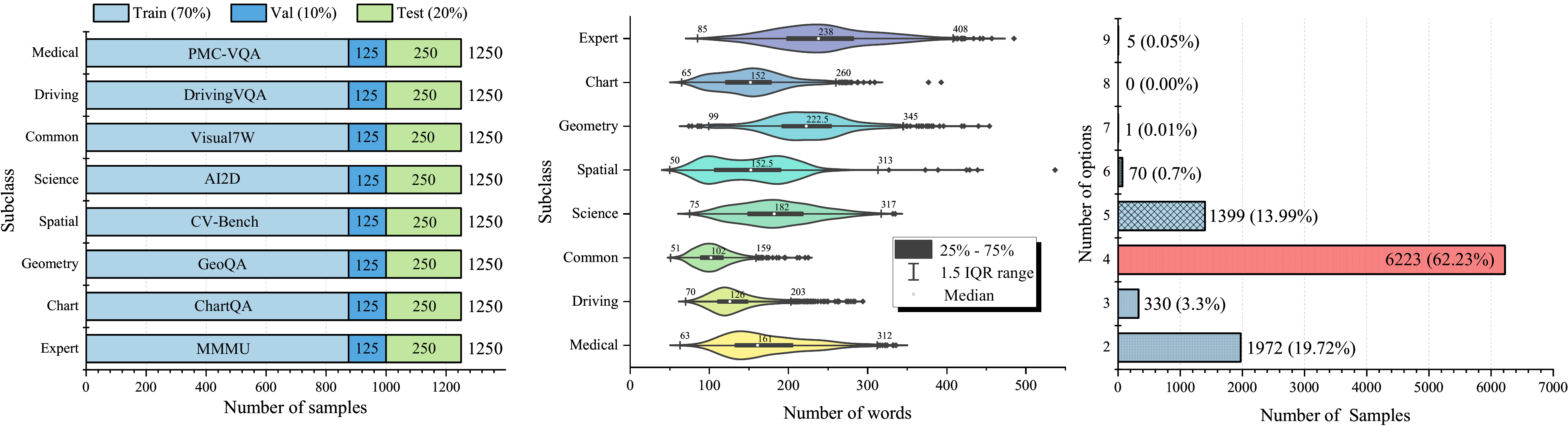}
    \vspace{-0.5cm}
    \caption{Statistics of VersaCoT-10K dataset. \textbf{Left:} data sources and splits across eight reasoning categories.\textbf{ Middle:} CoT length distribution. \textbf{Right:} distribution of candidate-option counts.}
    \label{figs:fig-04}
    \vspace{-0.2cm}
\end{figure*}

\subsection{Reasoning-State Augmentation}
After CoT Annotation and Quality Control, each retained sample is associated with a high-quality correct reasoning trajectory. However, supervision from a single correct CoT alone cannot expose the model to the diverse reasoning-state conditions required for reasoning-state understanding and refinement. We therefore adopt the Rationale Augmentation and Discrimination (RAD) paradigm~\citep{yu2026mind} to construct diverse positive states and hard negative states from each filtered CoT.

Specifically, given a correct CoT, we construct positive reasoning states ${\cal R}_i^{+}$ by introducing variations in wording, organization, and selected reasoning details while preserving the original reasoning semantics and conclusion. These diverse but semantically consistent states expose the model to different valid realizations of the same reasoning process, providing supervision for preserving and refining correct reasoning. In parallel, we construct hard negative reasoning states ${\cal R}_i^{-}$ by modifying a small number of critical semantic elements, visual relations, or inference steps. Such negative states remain superficially similar to the correct reasoning trajectory while containing key reasoning errors, thereby providing challenging supervision for identifying and correcting erroneous reasoning.

Together with the empty state $\phi$, the pre-constructed reasoning-state set for sample $i$ is defined as
\begin{gather}
{\cal S}_i^{RAD}
=
\left\{
\phi,\;
{\cal R}_i^{+},\;
{\cal R}_i^{-}
\right\},
\end{gather}
where $\phi$ is represented by the special token \texttt{[EMPTY\_SOLUTION]}. These three state types correspond to the three reasoning behaviors modeled by URSM: the empty state $\phi$ provides supervision for reasoning construction, positive states ${\cal R}_i^{+}$ for correct reasoning preservation and refinement, and negative states ${\cal R}_i^{-}$ for erroneous reasoning correction. During training, positive and negative states are sampled with equal probability when a non-empty reasoning state is selected.

It is important to distinguish these RAD-generated states from the historical reasoning states used by CRSL. The positive and negative states ${\cal R}_i^{+}$ and ${\cal R}_i^{-}$ are pre-constructed dataset-level annotations and remain fixed throughout training. In contrast, the historical states ${\cal R}_i^{h}$ are not part of the static VersaCoT-10K annotations. They are dynamically generated from ReState's own outputs and maintained during training, exposing the state-update operator to model-induced reasoning states that more closely resemble those encountered during recursive inference. Therefore, RAD provides diverse pre-constructed reasoning-state supervision, while CRSL further introduces dynamically evolving historical states to bridge the training-inference reasoning-state gap.

\begin{table*}[t]
\centering
\caption{The Break-Down ablation of ReState on the M$^3$CoT dataset. SHS and ISR denote the training-time self-history state mechanism and inference-time iterative self-refinement in CRSL.}
\vspace{-6pt}
\label{tab:s-tab05}
\setlength{\tabcolsep}{3mm}
\renewcommand{\arraystretch}{1.1}
\resizebox{0.95\linewidth}{!}{
\begin{tabular}{c|c|cc|c|c|c}
\hline
\multirow{2}{*}{Scheme}
& \multirow{2}{*}{URSM}
& \multicolumn{2}{c|}{CRSL}
& \multirow{2}{*}{Qwen2.5-VL-7B}
& \multirow{2}{*}{Qwen3-VL-8B}
& \multirow{2}{*}{Qwen3.5-9B}
\\
\cline{3-4}

&
& \multicolumn{1}{c|}{SHS}
& ISR
&
&
&
\\
\hline

Baseline
& \textbf{\ding{55}}
& \multicolumn{1}{c|}{\textbf{\ding{55}}}
& \textbf{\ding{55}}
& 79.42 ({\color[HTML]{7030A0}+5.57})
& 81.62 ({\color[HTML]{7030A0}+6.34})
& 84.73 ({\color[HTML]{7030A0}+4.27})
\\
\hline

Baseline w/ URSM
& \textbf{\ding{51}}
& \multicolumn{1}{c|}{\textbf{\ding{55}}}
& \textbf{\ding{55}}
& 84.12 ({\color[HTML]{7030A0}+0.87})
& 86.58 ({\color[HTML]{7030A0}+1.38})
& 88.65 ({\color[HTML]{7030A0}+0.35})
\\

Baseline w/ URSM, CRSL(SHS)
& \textbf{\ding{51}}
& \multicolumn{1}{c|}{\textbf{\ding{51}}}
& \textbf{\ding{55}}
& 84.17 ({\color[HTML]{7030A0}+0.82})
& 86.67 ({\color[HTML]{7030A0}+1.29})
& 88.83 ({\color[HTML]{7030A0}+0.17})
\\

Baseline w/ URSM, CRSL(ISR)
& \textbf{\ding{51}}
& \multicolumn{1}{c|}{\textbf{\ding{55}}}
& \textbf{\ding{51}}
& 84.43 ({\color[HTML]{7030A0}+0.56})
& 87.14 ({\color[HTML]{7030A0}+0.82})
& 88.65 ({\color[HTML]{7030A0}+0.35})
\\
\hline

\rowcolor[HTML]{F1F1F1}
\textbf{ReState (Final)}
& \textbf{\ding{51}}
& \multicolumn{1}{c|}{\cellcolor[HTML]{F1F1F1}\textbf{\ding{51}}}
& \textbf{\ding{51}}
& \textbf{84.99}
& \textbf{87.96}
& \textbf{89.00}
\\
\hline

\end{tabular}
}

\end{table*}

\begin{table*}[t]
\centering
\caption{The Break-Down ablation of ReState on the ScienceQA dataset. SHS and ISR denote the training-time self-history state mechanism and inference-time iterative self-refinement in CRSL.}
\vspace{-6pt}
\label{tab:s-tab06}
\setlength{\tabcolsep}{3mm}
\renewcommand{\arraystretch}{1.1}
\resizebox{0.95\linewidth}{!}{
\begin{tabular}{c|c|cc|c|c|c}
\hline
\multirow{2}{*}{Scheme}
& \multirow{2}{*}{URSM}
& \multicolumn{2}{c|}{CRSL}
& \multirow{2}{*}{Qwen2.5-VL-7B}
& \multirow{2}{*}{Qwen3-VL-8B}
& \multirow{2}{*}{Qwen3.5-9B}
\\
\cline{3-4}

&
& \multicolumn{1}{c|}{SHS}
& ISR
&
&
&
\\
\hline

Baseline
& \textbf{\ding{55}}
& \multicolumn{1}{c|}{\textbf{\ding{55}}}
& \textbf{\ding{55}}
& 96.11 ({\color[HTML]{7030A0}+1.32})
& 96.89 ({\color[HTML]{7030A0}+1.27})
& 95.59 ({\color[HTML]{7030A0}+2.78})
\\
\hline

Baseline w/ URSM
& \textbf{\ding{51}}
& \multicolumn{1}{c|}{\textbf{\ding{55}}}
& \textbf{\ding{55}}
& 97.31 ({\color[HTML]{7030A0}+0.12})
& 97.95 ({\color[HTML]{7030A0}+0.21})
& 98.04 ({\color[HTML]{7030A0}+0.33})
\\

Baseline w/ URSM, CRSL(SHS)
& \textbf{\ding{51}}
& \multicolumn{1}{c|}{\textbf{\ding{51}}}
& \textbf{\ding{55}}
& 97.38 ({\color[HTML]{7030A0}+0.05})
& 98.07 ({\color[HTML]{7030A0}+0.09})
& 98.37 ({\color[HTML]{7030A0}+0.00})
\\

Baseline w/ URSM, CRSL(ISR)
& \textbf{\ding{51}}
& \multicolumn{1}{c|}{\textbf{\ding{55}}}
& \textbf{\ding{51}}
& 97.31 ({\color[HTML]{7030A0}+0.12})
& 97.95 ({\color[HTML]{7030A0}+0.21})
& 98.07 ({\color[HTML]{7030A0}+0.30})
\\
\hline

\rowcolor[HTML]{F1F1F1}
\textbf{ReState (Final)}
& \textbf{\ding{51}}
& \multicolumn{1}{c|}{\cellcolor[HTML]{F1F1F1}\textbf{\ding{51}}}
& \textbf{\ding{51}}
& \textbf{97.43}
& \textbf{98.16}
& \textbf{98.37}
\\
\hline

\end{tabular}
}
\vspace{-0.2cm}
\end{table*}

\begin{table*}[t]
\centering
\caption{Ablation of different reasoning states in ReState on VersaCoT-10K. $\phi\;$, ${{\cal R}^ + }$, ${{\cal R}^ - }$, and ${{\cal R}^h}$ denote empty, positive, negative, and self-generated historical reasoning states, respectively.}
\vspace{-6pt}
\label{tab:s-tab07}
\setlength{\tabcolsep}{3mm}
\renewcommand{\arraystretch}{1.1}
\resizebox{0.95\linewidth}{!}{

\begin{tabular}{c|c|c|c|c|c|c|c}
\hline
Scheme
& $\phi$
& ${{\cal R}^ + }$
& ${{\cal R}^ - }$
& ${{\cal R}^h}$
& Qwen2.5-VL-7B
& Qwen3-VL-8B
& Qwen3.5-9B
\\
\hline

\textbf{ReState (Final) w/o $\phi$}
& \textbf{\ding{55}}
& \textbf{\ding{51}}
& \textbf{\ding{51}}
& \textbf{\ding{51}}
& 75.70 ({\color[HTML]{7030A0}+1.35})
& 77.95 ({\color[HTML]{7030A0}+1.30})
& 80.15 ({\color[HTML]{7030A0}+1.75})
\\

\textbf{ReState (Final) w/o ${{\cal R}^ + }$}
& \textbf{\ding{51}}
& \textbf{\ding{55}}
& \textbf{\ding{51}}
& \textbf{\ding{51}}
& 75.65 ({\color[HTML]{7030A0}+1.40})
& 78.50 ({\color[HTML]{7030A0}+0.75})
& 79.50 ({\color[HTML]{7030A0}+2.40})
\\

\textbf{ReState (Final) w/o ${{\cal R}^ - }$}
& \textbf{\ding{51}}
& \textbf{\ding{51}}
& \textbf{\ding{55}}
& \textbf{\ding{51}}
& 76.45 ({\color[HTML]{7030A0}+0.60})
& 78.90 ({\color[HTML]{7030A0}+0.35})
& 81.75 ({\color[HTML]{7030A0}+0.15})
\\

\textbf{ReState (Final) w/o ${{\cal R}^h}$}
& \textbf{\ding{51}}
& \textbf{\ding{51}}
& \textbf{\ding{51}}
& \textbf{\ding{55}}
& 76.20 ({\color[HTML]{7030A0}+0.85})
& 78.65 ({\color[HTML]{7030A0}+0.60})
& 79.45 ({\color[HTML]{7030A0}+2.45})
\\
\hline

\rowcolor[HTML]{F1F1F1}
\textbf{ReState (Final)}
& \textbf{\ding{51}}
& \textbf{\ding{51}}
& \textbf{\ding{51}}
& \textbf{\ding{51}}
& \textbf{77.05}
& \textbf{79.25}
& \textbf{81.90}
\\
\hline

\end{tabular}

}

\end{table*}

\section{Data Statistics for VersaCoT-10K}
\label{sec:a-data statistics-for-versaCoT-10K}

VersaCoT-10K contains 10,000 high-quality single-image multiple-choice visual reasoning samples spanning eight heterogeneous visual reasoning domains. From \cref{figs:fig-04}, we characterize the dataset from three perspectives: domain diversity, reasoning-length distribution, and candidate-space variability.

\textbf{\textit{1) Diverse Reasoning Domains.}}
VersaCoT-10K covers eight visual reasoning domains derived from PMC-VQA~\citep{zhang2023pmc}, DrivingVQA~\citep{corbiere2026drivingvqa}, Visual7W~\citep{zhu2016visual7w}, AI2D~\citep{kembhavi2016diagram}, CV-Bench~\citep{tong2024cambrian}, GeoQA~\citep{chen2021geoqa}, ChartQA~\citep{masry2022chartqa}, and MMMU~\citep{yue2024mmmu}, respectively. These domains encompass medical reasoning, driving and safety, visual commonsense, scientific diagram, spatial perception, geometric and mathematical, chart computation, and multidisciplinary knowledge. Each domain contains 1,250 samples, of which 20\% are reserved for testing. This balanced composition prevents the overall benchmark performance from being disproportionately dominated by any single domain and enables more comprehensive evaluation across heterogeneous reasoning patterns.

\textbf{\textit{2) Reasoning-Length Distribution.}}
The CoT annotations have an average length of 170.9 words and a median length of 162 words, with lengths ranging from 50 to 537 words. The detailed distributions across the eight domains are shown in \cref{figs:fig-04}. Although reasoning length alone does not directly reflect reasoning difficulty, the retained CoTs have undergone strict quality control for answer consistency, visual grounding, reasoning correctness, completeness, and relevance. Consequently, they provide explicit and informative supervision involving visual evidence interpretation, knowledge integration, relational reasoning, and intermediate computation across diverse tasks, supporting both multi-domain reasoning and reasoning-state refinement.

\textbf{\textit{3) Variable Candidate Spaces.}}
All samples are standardized into a closed-choice multiple-choice format while preserving variations in the number of candidate options inherited from different task settings. The candidate-set size ranges from 2 to 9, with four-option questions accounting for the largest proportion. Rather than enforcing a fixed answer space, this design preserves task-specific decision structures within a unified evaluation interface, reducing dependence on a single option configuration and providing more diverse multiple-choice reasoning conditions.

\section{More Detailed Experimental Setup}
\label{sec:a-more-detailed-experimental-setup}

\textbf{\textit{1) Datasets.}}
To evaluate the effectiveness and generalizability of ReState, we conduct experiments on the proposed VersaCoT-10K dataset and two widely used multimodal reasoning benchmarks, ScienceQA~\citep{lu2022learn} and M$^3$CoT~\citep{chen2024m3cot}. ScienceQA focuses on scientific reasoning and contains 21,208 samples, while M$^3$CoT contains 11,289 samples spanning science, commonsense, and mathematics. For ScienceQA and M$^3$CoT, we follow their original data splits to ensure fair comparison with prior work. For VersaCoT-10K, we use the training, validation, and test splits described in \cref{sec:a-detailed-construction-for-versaCoT-10K}. Following~\citep{yu2026mind}, we adopt the corresponding RAD versions of ScienceQA and M$^3$CoT, which provide positive and negative reasoning states for learning unified reasoning-state transitions.

\textbf{\textit{2) Implementation Details.}}
We conduct experiments with three MLLM backbones: Qwen2.5-VL-7B~\citep{bai2025qwen2}, Qwen3-VL-8B~\citep{bai2025qwen3vltechnicalreport}, and Qwen3.5-9B~\citep{qwen3.5}. All models are trained for 5 epochs with a learning rate of $1\times10^{-5}$. The batch size is set to 4 for Qwen2.5-VL-7B and Qwen3-VL-8B, and to 2 for Qwen3.5-9B. The maximum input length is uniformly set to 4096. Following prior experimental settings~\citep{yu2026mind} and accounting for differences in reasoning complexity and output length across datasets, the maximum output length is set to 512 for ScienceQA and M$^3$CoT and to 1024 for VersaCoT-10K. Image captions are generated using Qwen2.5-VL-72B~\citep{bai2025qwen2}, following~\citep{yu2026mind}. During ReState training, URSM constructs two views for each sample in every epoch: one with the empty reasoning state $\phi$ and one with a non-empty reasoning state. For the non-empty view, positive and negative reasoning states are sampled with equal probability. Therefore, each sample contributes two training instances per epoch. Building on URSM, CRSL further maintains a sample-specific historical reasoning-state pool with a maximum capacity of 2, from which model-generated historical states can be reused as subsequent training inputs. At inference, Restate evaluates a fixed refinement trajectory consisting of 9 rounds for all backbones and datasets. Following the same protocol, we report the peak dataset-level accuracy along this trajectory to characterize the refinement behavior, while the complete round-wise results are provided in \cref{figs:fig-05}. This evaluation summarizes the refinement trajectory and does not involve per-sample iteration selection. For settings without iterative self-refinement, the model performs only one additional answer-generation or reasoning-correction round after the initial CoT generation, providing a comparable non-iterative inference setting. For Direct SFT, we use ``Answer + CoT'' as the supervision target. All experiments are conducted on 8 NVIDIA H20 GPUs.

\begin{table*}[t]
\centering
\caption{Main results (\%) on the VersaCoT-10K dataset. Learning = Learning and training methods. Size = backbone model size. MED = medical, DRIVE = driving and safety, COMMON = visual commonsense, DIAG = scientific diagram, SPAT = spatial and perceptual, GEO = geometric and mathematical, CHART = chart and quantitative, and MULTI = multidisciplinary. \textcolor{bestcolor}{Red} denotes the best result, and \textcolor{secondcolor}{blue} denotes the second best result. }
\vspace{-6pt}
\label{tab:s-tab08}
\setlength{\tabcolsep}{2mm}
\renewcommand{\arraystretch}{1.11}
\resizebox{1\linewidth}{!}{
\begin{tabular}{cccccccccccc}
\hline
\multicolumn{1}{c|}{Model}                                                           & \multicolumn{1}{c|}{Learning}                          & \multicolumn{1}{c|}{Size}                       & \multicolumn{1}{c|}{MED}                                    & \multicolumn{1}{c|}{DRIVE}                                  & \multicolumn{1}{c|}{COMMON}                                 & \multicolumn{1}{c|}{DIAG}                                   & \multicolumn{1}{c|}{SPAT}                                   & \multicolumn{1}{c|}{GEO}                                    & \multicolumn{1}{c|}{CHART}                         & \multicolumn{1}{c|}{MULTI}                                  & \textbf{Avg}   \\ \hline
\multicolumn{1}{c|}{Random}                                                          & \multicolumn{1}{c|}{-}                                 & \multicolumn{1}{c|}{-}                          & \multicolumn{1}{c|}{22.00}                                       & \multicolumn{1}{c|}{46.80}                                       & \multicolumn{1}{c|}{26.80}                                       & \multicolumn{1}{c|}{26.27}                                       & \multicolumn{1}{c|}{39.87}                                       & \multicolumn{1}{c|}{22.93}                                       & \multicolumn{1}{c|}{22.00}                              & \multicolumn{1}{c|}{25.60}                                       &  29.03              \\
\hline
\end{tabular}

}
\vspace{-0.5cm}
\end{table*}

\section{More Ablation Experiments}
\label{sec:a-more-ablation experiments}
\subsection{Break-Down Ablation on M$^3$CoT and ScienceQA}
To further examine the contributions of individual components across different reasoning benchmarks, we conduct the same break-down ablations on M$^3$CoT and ScienceQA as those reported for VersaCoT-10K in the main text. Specifically, we evaluate URSM together with the two components of CRSL: training-time Self-History State (SHS) learning and inference-time Iterative Self-Refinement (ISR). The results are reported in \cref{tab:s-tab05,tab:s-tab06}.

As shown in \cref{tab:s-tab05}, ReState yields substantial and consistent gains on the more challenging M$^3$CoT benchmark. Introducing URSM alone improves Qwen2.5-VL-7B by 4.70\% (from 79.42\% to 84.12\%), Qwen3-VL-8B by 4.96\% (from 81.62\% to 86.58\%), and Qwen3.5-9B by 3.92\% (from 84.73\% to 88.65\%). These improvements demonstrate the effectiveness of unifying reasoning construction, correct reasoning preservation, and erroneous reasoning correction as transitions within a shared reasoning-state space. Building on URSM, either SHS or ISR provides further improvements, indicating that both training-time exposure to model-induced historical states and inference-time recursive state feedback contribute additional benefits. When both are enabled, the full ReState reaches 84.99\%, 87.96\%, and 89.00\% on the three backbones, corresponding to gains of 5.57\%, 6.34\%, and 4.27\% over their respective baselines. The full model achieves the best performance across all three backbones, showing that unified state modeling, historical-state learning, and iterative self-refinement provide complementary benefits.

A similar trend is observed on ScienceQA in \cref{tab:s-tab06}. URSM alone improves the three backbones by 1.20\% (from 96.11\% to 97.31\%), 1.06\% (from 96.89\% to 97.95\%), and 2.45\% (from 95.59\% to 98.04\%), respectively. Incorporating SHS further increases performance to 97.38\%, 98.07\%, and 98.37\%. The full ReState achieves 97.43\%, 98.16\%, and 98.37\%, improving over the corresponding baselines by 1.32\%, 1.27\%, and 2.78\%. Compared with M$^3$CoT, the additional gains from SHS and ISR are smaller on ScienceQA, which is consistent with the relatively limited headroom under already strong baseline performance. Nevertheless, the full ReState achieves the best result across all three backbones, indicating that reasoning-state modeling and refinement remain effective even when overall benchmark accuracy is already high.

Overall, the results on M$^3$CoT and ScienceQA are consistent with the observations on VersaCoT-10K in the main text. URSM provides the shared state-transition foundation for jointly modeling heterogeneous reasoning behaviors. SHS further exposes the state-update operator to model-induced historical states during training, reducing the training-inference reasoning-state gap, while ISR recursively feeds the current state back into the same operator to enable iterative refinement during inference. Their complementary gains across datasets and backbones demonstrate that the effectiveness of ReState is not specific to VersaCoT-10K, further supporting the general applicability of the proposed state-centric reasoning framework.

\subsection{Effect of Learning with Different Reasoning States}
To further investigate the roles of different reasoning states in ReState, we conduct state-level ablations on VersaCoT-10K using all three backbones. Starting from the full ReState, we separately remove the empty state $\phi$, positive states ${\cal R}^{+}$, negative states ${\cal R}^{-}$, and historical states ${\cal R}^{h}$. As shown in \cref{tab:s-tab07}, removing any state type degrades performance, whereas jointly using all four types consistently yields the strongest overall results. This demonstrates that the different reasoning states provide complementary supervision for unified reasoning-state learning.

Among the three pre-constructed states used by URSM, removing the empty state $\phi$ reduces accuracy by 1.35\%, 1.30\%, and 1.75\% on Qwen2.5-VL-7B, Qwen3-VL-8B, and Qwen3.5-9B, respectively. This confirms the importance of explicitly learning to construct an initial reasoning state from an empty input state. Removing positive states ${\cal R}^{+}$ causes larger drops of 0.75\%-2.40\%, with Qwen3.5-9B decreasing from 81.90\% to 79.50\%. These results show that exposure to diverse valid reasoning states is important for learning to preserve correct evidence while further improving reasoning completeness and consistency. Removing negative states ${\cal R}^{-}$ results in smaller but consistent decreases of 0.15\%-0.60\% across the three backbones, indicating that erroneous states provide useful supervision for identifying and correcting reasoning errors in existing states.

Historical states ${\cal R}^{h}$ play a different role from the three pre-constructed URSM states. Removing them reduces performance by 0.60\%--2.45\%, with Qwen3.5-9B decreasing from 81.90\% to 79.45\%. Unlike $\phi$, ${\cal R}^{+}$, and ${\cal R}^{-}$, which are predefined before training, historical states are dynamically generated by ReState itself and therefore reflect model-induced reasoning patterns and errors. Exposing the state-update operator to these states during training better aligns its input distribution with the states encountered during recursive inference. The observed degradation without ${\cal R}^{h}$ therefore further validates the role of SHS in mitigating the training-inference reasoning-state gap.

Overall, the four state types provide complementary supervision for ReState. The empty state $\phi$ supports reasoning construction, positive states ${\cal R}^{+}$ support correct reasoning preservation and refinement, and negative states ${\cal R}^{-}$ support erroneous reasoning correction. Historical states ${\cal R}^{h}$ further extend this predefined state space with dynamically generated model-induced states, enabling closed-loop reasoning-state learning. Although the magnitude of each contribution varies across backbones, removing any state type consistently reduces performance. These results jointly validate URSM's unified modeling of heterogeneous reasoning behaviors and CRSL's complementary modeling of historical state distributions, further supporting the proposed state-centric reasoning paradigm.

\begin{table*}[t]
\centering
\caption{Main results (\%) on the ScienceQA dataset. Learning = Learning and training methods. Size=backbone model size. NAT = natural science, SOC= social science, LAN = language science, TXT = text context, IMG = image context, NO = no context, G1-6 = grades 1-6, G7-12 = grades 7-12. \textcolor{bestcolor}{Red} denotes the best result, and \textcolor{secondcolor}{blue} denotes the second best result. }
\vspace{-6pt}
\label{tab:s-tab09}
\setlength{\tabcolsep}{2mm}
\renewcommand{\arraystretch}{1.12}
\resizebox{1\linewidth}{!}{
\begin{tabular}{c|c|c|c|c|c|c|c|c|c|c|c}
\hline
Model                                   & Learning           & Size     & NAT            & SOC            & LAN            & TXT            & IMG            & NO             & G1-6           & G7-12          & \textbf{Avg}   \\ \hline
Random                                  & -                  & -        & 40.28          & 46.13          & 29.25          & 47.45          & 40.08          & 33.66          & 39.35          & 40.67          & 39.83          \\ 
Human                                   & -                  & -        & 90.23          & 84.97          & 87.48          & 89.60          & 87.50          & 88.10          & 91.59          & 82.42          & 88.40          \\ \hline
MCAN                                    & \textit{Fine-tune} & 95M      & 56.08          & 46.23          & 58.09          & 59.43          & 51.17          & 55.40          & 51.65          & 59.72          & 54.54          \\ 
Top-Down                                & \textit{Fine-tune} & 70M      & 59.50          & 54.33          & 61.82          & 62.90          & 54.88          & 59.79          & 57.27          & 62.16          & 59.02          \\ 
BAN                                     & \textit{Fine-tune} & 112M     & 60.88          & 46.57          & 66.64          & 62.61          & 52.60          & 65.51          & 56.83          & 63.94          & 59.37          \\ 
DFAF                                    & \textit{Fine-tune} & 74M      & 64.03          & 48.82          & 63.55          & 65.88          & 54.49          & 64.11          & 57.12          & 67.17          & 60.72          \\ 
ViLT                                    & \textit{Fine-tune} & 113M     & 60.48          & 63.89          & 60.27          & 63.20          & 61.38          & 57.00          & 60.72          & 61.90          & 61.14          \\ 
Patch-TRM                               & \textit{Fine-tune} & 90M      & 65.19          & 46.79          & 65.55          & 66.96          & 55.28          & 64.95          & 58.04          & 67.50          & 61.42          \\ 
VisualBERT                              & \textit{Fine-tune} & 111M     & 59.33          & 69.18          & 61.18          & 62.71          & 62.17          & 58.54          & 62.96          & 59.92          & 61.87          \\ \hline
GPT-3.5                                 & \textit{Few-shot}  & 175B     & 74.64          & 69.74          & 76.00          & 74.44          & 67.28          & 77.42          & 76.80          & 68.89          & 73.97          \\ 
GPT-3.5 w/CoT                           & \textit{Few-shot}  & 175B     & 75.44          & 70.87          & 78.09          & 74.68          & 67.43          & 79.93          & 78.23          & 69.68          & 75.17          \\ 
ChatGPT w/CoT                           & \textit{Few-shot}  & 175B     & 78.82          & 70.98          & 83.18          & 77.37          & 67.92          & 86.13          & 80.72          & 74.03          & 78.31          \\ 
GPT-4 w/CoT                             & \textit{Few-shot}  & -        & 85.48          & 72.44          & 90.27          & 82.65          & 71.49          & 92.89          & 86.66          & 79.04          & 83.99          \\ 
Chameleon (ChatGPT)                     & \textit{Few-shot}  & -        & 81.62          & 70.64          & 84.00          & 79.77          & 70.80          & 86.62          & 81.86          & 76.53          & 79.93          \\ 
Chameleon (GPT-4)                       & \textit{Few-shot}  & -        & 89.83          & 74.13          & 89.82          & 88.27          & 77.64          & 92.13          & 88.03          & 83.72          & 86.54          \\ 
Gemini+RMR                              & \textit{Few-shot}  & -        & 91.79          & 94.26          & 89.64          & 91.40          & 89.69          & 91.01          & 92.84          & 89.78          & 91.75          \\ \hline
Qwen2.5-VL-7B                           & \textit{Zero-shot} & 7B       & 83.79          & 88.30          & 82.45          & 83.63          & 84.38          & 81.81          & 86.67          & 80.29          & 84.39          \\ 
Qwen3-VL-8B                             & \textit{Zero-shot} & 8B       & 90.54          & 94.49          & 87.64          & 92.18          & 91.67          & 86.20          & 90.64          & 90.57          & 90.62          \\ 
Qwen3.5-9B                              & \textit{Zero-shot} & 9B       & 91.34          & 94.38          & 88.55          & 92.72          & 92.81          & 86.69          & 92.47          & 89.06          & 91.25          \\ 
Qwen3.5-Flash                           & \textit{Zero-shot} & -        & 94.98          & 98.43          & 89.18          & 95.60          & 95.79          & 89.83          & 94.82          & 93.08          & 94.20          \\ 
Qwen3.5-Plus                            & \textit{Zero-shot} & -        & 97.20          & 97.86          & 95.73          & 97.12          & 96.88          & 95.68          & 97.17          & 96.57          & 96.96          \\ 
Qwen3.6-Flash                           & \textit{Zero-shot} & -        & 94.85          & 97.98          & 90.36          & 95.41          & 95.09          & 90.80          & 94.68          & 93.74          & 94.34          \\ 
Qwen3.6-Plus                            & \textit{Zero-shot} & -        & 96.98          & 97.19          & 94.91          & 96.97          & 96.58          & 94.77          & 96.77          & 95.98          & 96.49          \\
Qwen3.7-Flash                           & \textit{Zero-shot} & -        & 95.20          & 98.31          & 89.09          & 95.80          & 95.64          & 89.97          & 95.30          & 92.42          & 94.27          \\ 
Qwen3.7-Plus                            & \textit{Zero-shot} & -        & 96.89          & 97.98          & 94.91          & 97.12          & 96.93          & 94.77          & 96.95          & 95.98          & 96.60          \\ 
Qwen3.8-Max                             & \textit{Zero-shot} & -        & 98.31          & 99.55          & 96.55          & 98.68          & 98.81          & 96.38          & 98.31          & 97.76          & 98.11          \\
MiniMax-M3                              & \textit{Zero-shot} & -        & 93.16          & 95.05          & 88.91          & 93.45          & 92.96          & 89.34          & 93.61          & 90.38          & 92.45          \\ 
Grok-4.6                                & \textit{Zero-shot} & -        & 95.60          & 95.39          & 94.64          & 96.63          & 94.89          & 93.80          & 96.15          & 93.80          & 95.31          \\ 
GLM-4.6V                                & \textit{Zero-shot} & -        & 92.41          & 93.25          & 93.73          & 91.84          & 90.43          & 93.45          & 93.50          & 91.89          & 92.93          \\ 
GPT-4o                                  & \textit{Zero-shot} & -        &  91.47              & 91.34               &  92.00              & 89.44               &  87.95              &   92.68             &  93.17              &  88.73              &  91.58              \\ 
GPT-5.6-Luna                            & \textit{Zero-shot} & -        & 92.14          & 88.41          & 91.09          & 90.57          & 88.35          & 91.08          & 92.91          & 87.80          & 91.09          \\ 
GPT-5.6-Terra                           & \textit{Zero-shot} & -        & 94.45          & 92.13          & 93.91          & 94.18          & 92.02          & 93.59          & 94.82          & 92.02          & 93.82          \\ 
GPT-5.6-Sol                             & \textit{Zero-shot} & -        & 98.18          & 95.95          & 98.82          & 97.65          & 96.58          & 98.75          & 98.38          & 96.97          & 97.88          \\ 
Gemini-3-Flash                          & \textit{Zero-shot} & -        & 98.31          & 98.20          & 97.91          & 98.04          & 97.97          & 97.84          & 98.46          & 97.69          & 98.18          \\ 
Gemini-3.5-Flash                        & \textit{Zero-shot} & -        & 98.52          & 98.43          & 97.64          & 98.19          & 98.02          & 97.98          & 98.57          & 97.82          & 98.30          \\ 
Claude Haiku 4.5                        & \textit{Zero-shot} & -        & 89.17          & 89.31          & 92.45          & 87.83          & 85.37          & 92.54          & 92.14          & 86.29          & 90.05          \\ 
Claude Sonnet 5                         & \textit{Zero-shot} & -        & 95.60          & 95.39          & 95.45          & 94.13          & 93.90          & 95.89          & 96.26          & 94.20          & 95.52          \\ 
Claude Opus 5                           & \textit{Zero-shot} & -        & 98.45          & 96.85          & 98.64          & 97.41          & 97.22          & 98.68          & 98.60          & 97.36          & 98.16          \\ 
Seed-2.1-Turbo                          & \textit{Zero-shot} & -        & 96.67          & 97.08          & 96.36          & 96.68          & 95.79          & 96.45          & 96.92          & 96.24          & 96.68          \\ 
Seed-2.1-Pro                            & \textit{Zero-shot} & -        & 97.11          & 97.98          & 96.27          & 97.41          & 96.68          & 96.24          & 97.10          & 97.03          & 97.08          \\
Kimi-K3                                 & \textit{Zero-shot} & 2.8T     & 97.96          & 98.31          & 99.09          & 98.00          & 97.57          & 98.82          & 98.46          & 98.09          & 98.33          \\ \hline
LLaMA-Adapter                           & \textit{Fine-tune} & 6B       & 84.37          & 88.30          & 84.36          & 83.72          & 80.32          & 86.90          & 85.83          & 84.05          & 85.19          \\ 
LLaVA                                   & \textit{Fine-tune} & 13B      & 90.36          & 95.95          & 88.00          & 89.49          & 88.00          & 90.66          & 90.93          & 90.90          & 90.92          \\ 
LaVIN                                   & \textit{Fine-tune} & 7B       & 89.25          & 94.94          & 85.24          & 88.51          & 87.46          & 88.08          & 90.16          & 88.07          & 89.41          \\ 
UnifiedQA$_{base}$                           & \textit{Fine-tune} & 223M     & 68.16          & 69.18          & 74.91          & 63.78          & 61.38          & 77.84          & 72.98          & 65.00          & 70.12          \\
UnifiedQA$_{base}$ w/CoT                     & \textit{Fine-tune} & 223M     & 71.00          & 76.04          & 78.91          & 66.42          & 66.53          & 81.81          & 77.06          & 68.82          & 74.11          \\ 
Multimodal-CoT$_{base}$                     & \textit{Fine-tune} & 223M     & 84.06          & 92.35          & 82.18          & 82.75          & 82.75          & 84.74          & 85.79          & 84.44          & 85.31          \\
DDCoT$_{base}$                               & \textit{Fine-tune} & 223M     & 88.72          & 86.84          & 84.91          & 87.59          & 83.34          & 88.08          & 88.58          & 85.10          & 87.34          \\ 
MC-CoT$_{base}$                              & \textit{Fine-tune} & 223M     & 91.87          & 84.59          & 93.00          & 92.28          & 88.30          & 92.75          & 90.64          & 90.64          & 90.64          \\ 
DPMM-CoT$_{base}$                            & \textit{Fine-tune} & 223M+83M & 92.72          & 87.85          & 89.91          & 92.72          & 90.48          & 91.29          & 91.45          & 90.11          & 90.97          \\
Multimodal-T-SciQ$_{base}$                   & \textit{Fine-tune} & 223M     & 91.52          & 91.45          & 92.45          & 91.94          & 90.33          & 92.26          & 92.11          & 91.10          & 91.75          \\ 
MIND$_{base}$                                & \textit{Fine-tune} & 223M     & 93.07          & 96.74          & 87.09          & 92.42          & 92.76          & 89.27          & 92.91          & 91.17          & 92.29          \\ \hline
Qwen2.5-VL-7B                           & \textit{Fine-tune}          & 7B       & 97.02          & 97.86          & 92.82          & 96.77          & 96.33          & 94.15          & 96.92          & 94.66          & 96.11          \\
Qwen2.5-VL-7B + Multimodal-CoT          & \textit{Fine-tune}          & 7B       & 97.47          & 97.19          & 90.82          & 97.46          & 96.58          & 92.40          & 96.04          & 95.06          & 95.68          \\ 
Qwen2.5-VL-7B + MIND                    & \textit{Fine-tune}          & 7B       & \textcolor{secondcolor}{97.56}          & \textcolor{secondcolor}{98.20}          & \textcolor{secondcolor}{93.36}          & \textcolor{secondcolor}{97.56}          & \textcolor{secondcolor}{97.03}          & \textcolor{secondcolor}{94.49}          & \textcolor{secondcolor}{97.06}          & \textcolor{secondcolor}{95.78}          & \textcolor{secondcolor}{96.60}          \\ 
\rowcolor[HTML]{EFEFEF} 
\textbf{Qwen2.5-VL-7B + ReState (Ours)} & \textit{Fine-tune}          & 7B       & \textbf{\textcolor{bestcolor}{97.91}} & \textbf{\textcolor{bestcolor}{99.44}} & \textbf{\textcolor{bestcolor}{94.82}} & \textbf{\textcolor{bestcolor}{98.04}} & \textbf{\textcolor{bestcolor}{98.17}} & \textbf{\textcolor{bestcolor}{95.54}} & \textbf{\textcolor{bestcolor}{97.91}} & \textbf{\textcolor{bestcolor}{96.57}} & \textbf{\textcolor{bestcolor}{97.43}} \\ \hline
Qwen3-VL-8B                             & \textit{Fine-tune}          & 8B       & \textcolor{secondcolor}{98.13}          & 97.19          & 94.09          & \textcolor{secondcolor}{98.00}          & 97.12          & 95.12          & 96.95          & 96.77          & 96.89          \\ 
Qwen3-VL-8B + Multimodal-CoT            & \textit{Fine-tune}          & 8B       & 97.42          & 97.08          & 94.27          & 97.26          & 96.28          & 95.33          & 96.92          & 95.85          & 96.53          \\ 
Qwen3-VL-8B + MIND                      & \textit{Fine-tune}          & 8B       & {97.96}          & \textcolor{secondcolor}{98.31}          & \textcolor{secondcolor}{94.64}          & {97.90}          & \textcolor{secondcolor}{97.37}          & \textcolor{secondcolor}{95.61}          & \textcolor{secondcolor}{97.32}          & \textcolor{secondcolor}{96.90}          & \textcolor{secondcolor}{97.17}          \\
\rowcolor[HTML]{EFEFEF} 
\textbf{Qwen3-VL-8B + ReState (Ours)}   & \textit{Fine-tune}          & 8B       & \textbf{\textcolor{bestcolor}{99.07}} & \textbf{\textcolor{bestcolor}{99.10}} & \textbf{\textcolor{bestcolor}{95.55}} & \textbf{\textcolor{bestcolor}{99.17}} & \textbf{\textcolor{bestcolor}{98.81}} & \textbf{\textcolor{bestcolor}{96.38}} & \textbf{\textcolor{bestcolor}{98.27}} & \textbf{\textcolor{bestcolor}{97.96}} & \textbf{\textcolor{bestcolor}{98.16}} \\ \hline
Qwen3.5-9B                              & \textit{Fine-tune}          & 9B       & 95.69          & 96.40          & 94.73          & 95.31          & 94.15          & \textcolor{secondcolor}{95.68}          & 95.89          & 95.06          & 95.59          \\ 
Qwen3.5-9B + Multimodal-CoT             & \textit{Fine-tune}          & 9B       & \textcolor{secondcolor}{96.40}          & 96.63          & 92.73          & \textcolor{secondcolor}{96.29}          & \textcolor{secondcolor}{95.49}          & 94.01          & 95.78          & 94.99          & 95.50          \\
Qwen3.5-9B + MIND                       & \textit{Fine-tune}          & 9B       & 95.78          & \textcolor{secondcolor}{97.64}          & \textcolor{secondcolor}{97.18}          & 95.45          & 94.60          & \textbf{\textcolor{bestcolor}{97.63}} & \textcolor{secondcolor}{97.10}          & \textcolor{secondcolor}{95.52}          & \textcolor{secondcolor}{96.53}          \\ 
\rowcolor[HTML]{EFEFEF} 
\textbf{Qwen3.5-9B + ReState (Ours)}    & \textit{Fine-tune}          & 9B       & \textbf{\textcolor{bestcolor}{98.71}} & \textbf{\textcolor{bestcolor}{98.76}} & \textbf{\textcolor{bestcolor}{97.36}} & \textbf{\textcolor{bestcolor}{98.78}} & \textbf{\textcolor{bestcolor}{98.46}} & \textbf{\textcolor{bestcolor}{97.63}} & \textbf{\textcolor{bestcolor}{98.60}} & \textbf{\textcolor{bestcolor}{97.96}} & \textbf{\textcolor{bestcolor}{98.37}} \\ \hline
\end{tabular}

}
\vspace{-0.1cm}
\end{table*}

\section{Comprehensive Comparison with Existing Methods}
\label{sec:comprehensive-comparison-with-existing-methods}

\subsection{Evaluation on the VersaCoT-10K Dataset}
\label{sec:s-evaluation-on-the-versaCoT-10K-dataset}
To provide a more comprehensive evaluation of multi-domain visual reasoning, we extend the open-source and closed-source model comparisons reported in \cref{tab:tab01} of the main text. The complete results are presented in \cref{tab:s-tab08}. Overall, substantial performance differences are observed across model families, model scales, and capability configurations, demonstrating that VersaCoT-10K can effectively differentiate the cross-domain visual reasoning capabilities of current MLLMs.

Among the evaluated open-source zero-shot models~\citep{bai2025qwen2,bai2025qwen3vltechnicalreport,qwen3.5,qwen3.6-27b,qwen3.6-35b-a3b,lai2026minimaxsparseattention,vteam2026glm45vglm41vthinkingversatilemultimodal,moonshot2026kimik26,kimiteam2026kimik3openfrontier}, performance varies substantially on VersaCoT-10K. Kimi-K3~\citep{kimiteam2026kimik3openfrontier} achieves the highest average accuracy of 85.65\%, followed by Qwen3.5-397B-A17B~\citep{qwen3.5} at 81.60\%. Qwen3-VL-235B-A22B~\citep{bai2025qwen3vltechnicalreport}, Qwen3.5-122B-A10B~\citep{qwen3.5}, and Qwen3.6-27B~\citep{qwen3.6-27b} achieve 80.65\%, 80.00\%, and 79.55\%, respectively. Although stronger and larger models often obtain higher accuracy, performance is not monotonically determined by total parameter count, indicating that model architecture, active parameterization, and training also affect cross-domain reasoning capability. These results highlight the ability of VersaCoT-10K to distinguish models with different architectural and capability configurations.

Several closed-source models~\citep{bai2023qwenvlversatilevisionlanguagemodel,bai2025qwen3vltechnicalreport,qwen3.5,qwen36plus,qwen37,qwen3.8,xai2026grok43,xai2026grok45,xai2026grok46,vteam2026glm5vturbonativefoundationmodel,openai2024gpt4ocard,singh2026openaigpt5card,openai2026gpt54,openai2026gpt55,openai2026gpt56,anthropic2025haiku45,anthropic2026sonnet46,anthropic2026sonnet5,anthropic2026opus48,anthropic2026opus5,bytedance2025seed16,seed2026seed18modelcardgeneralized,seed2026seed20modelcardintelligence,bytedance2026seed21} achieve strong overall performance. Grok-4.6~\citep{xai2026grok46}, Qwen3.8-Max~\citep{qwen3.8}, and Seed-2.1-Pro~\citep{bytedance2026seed21} reach 86.85\%, 85.35\%, and 85.20\%, respectively. Clear differences are also observed among capability configurations within the same model family. For example, the Qwen Plus variants generally outperform their corresponding Flash variants. GPT-5.6-Luna, Terra, and Sol~\citep{openai2026gpt56} achieve average accuracies of 71.00\%, 78.15\%, and 83.00\%, respectively. A similar progression is observed within the Claude family, from 70.10\% with Haiku 4.5~\citep{anthropic2025haiku45} to 84.70\% with Opus 5~\citep{anthropic2026opus5}. These results show that VersaCoT-10K captures not only differences across model families but also capability differences among variants within the same family.

The fine-grained results across the eight reasoning domains further reveal distinct capability profiles. Stronger models generally achieve high accuracy on \textit{COMMON} and \textit{CHART}, whereas \textit{MED}, \textit{DRIVE}, \textit{GEO}, and \textit{MULTI} exhibit substantially larger performance variation across models. For example, accuracy on \textit{GEO} ranges from 44.80\% to above 96.80\% among the evaluated models, while \textit{MULTI} similarly exhibits a wide performance range. These domains require different combinations of visual perception, domain knowledge, spatial understanding, and quantitative reasoning. Therefore, VersaCoT-10K provides both an overall measure of multi-domain reasoning through average accuracy and a fine-grained view of domain-specific capability differences.

More importantly, under matched backbone and training settings, ReState consistently achieves the highest average performance across all three evaluated backbones. With Qwen2.5-VL-7B~\citep{bai2025qwen2}, ReState reaches 77.05\%, outperforming Direct SFT, Multimodal-CoT~\citep{zhang2024multimodal}, and MIND~\citep{yu2026mind} by 1.90\% (from 75.15\% to 77.05\%), 4.40\% (from 72.65\% to 77.05\%), and 1.30\% (from 75.75\% to 77.05\%), respectively. With Qwen3-VL-8B~\citep{bai2025qwen3vltechnicalreport}, ReState achieves 79.25\%, corresponding to improvements of 3.60\% (from 75.65\% to 79.25\%), 2.75\% (from 76.50\% to 79.25\%), and 0.90\% (from 78.35\% to 79.25\%). With Qwen3.5-9B~\citep{qwen3.5}, ReState reaches 81.90\%, improving by 4.10\% (from 77.80\% to 81.90\%), 4.60\% (from 77.30\% to 81.90\%), and 1.70\% (from 80.20\% to 81.90\%), respectively.

The improvements of ReState are also not confined to a single reasoning domain. With Qwen3.5-9B, ReState achieves the best results on \textit{MED}, \textit{DRIVE}, \textit{COMMON}, \textit{SPAT}, and \textit{MULTI}. It also improves multiple reasoning domains with Qwen2.5-VL-7B and Qwen3-VL-8B. Although other methods retain advantages in several individual domains, ReState consistently achieves the highest average accuracy across all three backbones. This suggests that its gains arise from broadly improved reasoning-state modeling rather than optimization toward a particular domain.

Notably, ReState with Qwen3.5-9B achieves 81.90\%, surpassing all evaluated open-source zero-shot models in \cref{tab:s-tab08} except Kimi-K3~\citep{kimiteam2026kimik3openfrontier}. For example, Qwen3.5-397B-A17B~\citep{qwen3.5} and Qwen3-VL-235B-A22B~\citep{bai2025qwen3vltechnicalreport} achieve 81.60\% and 80.65\%, respectively. We emphasize that fine-tuned and zero-shot models are evaluated under different settings and therefore do not constitute strictly matched comparisons. Nevertheless, these results show that ReState can substantially enhance the reasoning capability of a medium-scale MLLM, enabling competitive multi-domain performance relative to much larger models.

Overall, the extended results demonstrate two complementary properties. First, VersaCoT-10K provides a discriminative benchmark that reveals substantial performance differences across model families, capability configurations, and reasoning domains. Second, under controlled matched-backbone comparisons, ReState consistently improves multi-domain reasoning performance across different MLLMs. Together, these results further support the effectiveness and cross-model adaptability of the proposed state-centric reasoning framework.

\begin{table*}[t]
\centering
\caption{Main results (\%) on the M$^3$CoT dataset. The ``Random'' and ``Human'' results are averaged over three trials. \textcolor{bestcolor}{Red} denotes the best result, and \textcolor{secondcolor}{blue} denotes the second best result. }
\vspace{-6pt}
\label{tab:s-tab10}
\setlength{\tabcolsep}{1.5mm}
\renewcommand{\arraystretch}{1.09}
\resizebox{1\linewidth}{!}{
\begin{tabular}{cccccccccccc}
\hline
\multicolumn{1}{c|}{}                                                                & \multicolumn{1}{c|}{}                           & \multicolumn{3}{c|}{Science}                                                                                                                                                            & \multicolumn{3}{c|}{Commonsense}                                                                                                                                                        & \multicolumn{3}{c|}{Mathematics}                                                                                                                                                        &                                  \\ \cline{3-11}
\multicolumn{1}{c|}{\multirow{-2}{*}{Model}}                                         & \multicolumn{1}{c|}{\multirow{-2}{*}{Size}}     & \multicolumn{1}{c|}{Lang}                                   & \multicolumn{1}{c|}{Natural}                                & \multicolumn{1}{c|}{Social}                                 & \multicolumn{1}{c|}{Physical}                               & \multicolumn{1}{c|}{Social}                                 & \multicolumn{1}{c|}{Temporal}                               & \multicolumn{1}{c|}{Algebra}                                & \multicolumn{1}{c|}{Geometry}                               & \multicolumn{1}{c|}{Theory}                                 & \multirow{-2}{*}{\textbf{Total}} \\
\hline
\end{tabular}

}

\end{table*}

\subsection{Evaluation on the ScienceQA Dataset}
\label{sec:s-evaluation-on-the-scienceQA-dataset}

To provide a more comprehensive evaluation on a standard scientific reasoning benchmark, we extend the ScienceQA comparisons reported in \cref{tab:tab02} of the main text. The complete results in \cref{tab:s-tab09} cover traditional visual question answering methods~\citep{yu2019deep,anderson2018bottom,kim2018bilinear,gao2019dynamic,kim2021vilt,lu2021iconqa,li2020does}, few-shot reasoning methods~\citep{wei2022chain,achiam2023gpt,team2023gemini,lu2023chameleon,tan2024retrieval}, several families of zero-shot MLLMs~\citep{bai2025qwen2,bai2025qwen3vltechnicalreport,qwen3.5,qwen36plus,qwen37,lai2026minimaxsparseattention,vteam2026glm45vglm41vthinkingversatilemultimodal,xai2026grok46,qwen3.8,openai2024gpt4ocard,openai2026gpt56,google2025gemini3flash,google2026gemini35flash,anthropic2025haiku45,anthropic2026sonnet5,anthropic2026opus5,bytedance2026seed21,kimiteam2026kimik3openfrontier}, and additional learning-based MCoT methods~\citep{zhang2023llama,liu2024improved,luo2023cheap,khashabi2020unifiedqa,zhang2024multimodal,zheng2023ddcot,tan2024boosting,he2024multi,wang2024t,yu2026mind,bai2025qwen2,bai2025qwen3vltechnicalreport,qwen3.5}. Overall, recent MLLMs already achieve high accuracy on ScienceQA, providing a complementary setting for evaluating whether ReState can still yield consistent gains when the remaining performance headroom is limited.

Across different method categories, ScienceQA performance improves substantially with advances in model capability and reasoning mechanisms. Traditional vision-language models, including MCAN~\citep{yu2019deep}, BAN~\citep{kim2018bilinear}, ViLT~\citep{kim2021vilt}, and VisualBERT~\citep{li2020does}, mainly achieve average accuracies in the range of 54.54\%-61.87\%. With stronger language models and explicit reasoning mechanisms, performance increases markedly. GPT-4 w/CoT~\citep{achiam2023gpt}, Gemini+RMR~\citep{team2023gemini,tan2024retrieval}, and MIND$_{base}$~\citep{yu2026mind} reach 83.99\%, 91.75\%, and 92.29\%, respectively. More recent closed-source MLLMs further raise the performance level: Qwen3.8-Max~\citep{qwen3.8}, GPT-5.6-Sol~\citep{openai2026gpt56}, Gemini-3.5-Flash~\citep{google2026gemini35flash}, Claude Opus 5~\citep{anthropic2026opus5}, and Kimi-K3~\citep{kimiteam2026kimik3openfrontier} all achieve accuracies around 98\%. These results indicate that ScienceQA has entered a high-accuracy regime for strong MLLMs, making further improvements increasingly challenging. It is worth mentioning that ReState with Qwen3.5-9B also achieves competitive performance against strong zero-shot MLLMs. Its 98.37\% accuracy exceeds Qwen3.8-Max, GPT-5.6-Sol, Gemini-3-Flash, Gemini-3.5-Flash, and Claude Opus 5, and is comparable to Kimi-K3 at 98.33\%.

More importantly, under matched backbone and training settings, ReState consistently achieves the highest average performance across all three evaluated backbones. With Qwen2.5-VL-7B~\citep{bai2025qwen2}, ReState reaches 97.43\%, outperforming Direct SFT, Multimodal-CoT~\citep{zhang2024multimodal}, and MIND~\citep{yu2026mind} by 1.32\% (from 96.11\% to 97.43\%), 1.75\% (from 95.68\% to 97.43\%), and 0.83\% (from 96.60\% to 97.43\%), respectively. With Qwen3-VL-8B~\citep{bai2025qwen3vltechnicalreport}, ReState achieves 98.16\%, corresponding to gains of 1.27\% (from 96.89\% to 98.16\%), 1.63\% (from 96.53\% to 98.16\%), and 0.99\% (from 97.17\% to 98.16\%). With Qwen3.5-9B~\citep{qwen3.5}, ReState further reaches 98.37\%, improving by 2.78\% (from 95.59\% to 98.37\%), 2.87\% (from 95.50\% to 98.37\%), and 1.84\% (from 96.53\% to 98.37\%), respectively. These consistent improvements demonstrate that ReState remains effective across different MLLM backbones even when baseline performance is already high.

The fine-grained results further show that the gains of ReState are not concentrated in a particular ScienceQA subset. With Qwen3-VL-8B, ReState achieves the strongest matched-backbone results across all reported dimensions, while improving the average accuracy from 97.17\% with MIND to 98.16\%. Similar overall trends are observed with Qwen2.5-VL-7B and Qwen3.5-9B. These results indicate that the benefits of ReState extend across different subject categories, context conditions, and grade levels rather than being driven by a particular subset.

Overall, the extended ScienceQA results demonstrate that ReState consistently improves performance across MLLM backbones in a high-accuracy evaluation regime. Together with its broad gains across ScienceQA subsets, these results further support the effectiveness of unified reasoning-state modeling and closed-loop reasoning-state refinement, as well as the cross-model adaptability of ReState.

\begin{figure*}[t]
    \centering
    \includegraphics[width=\linewidth]{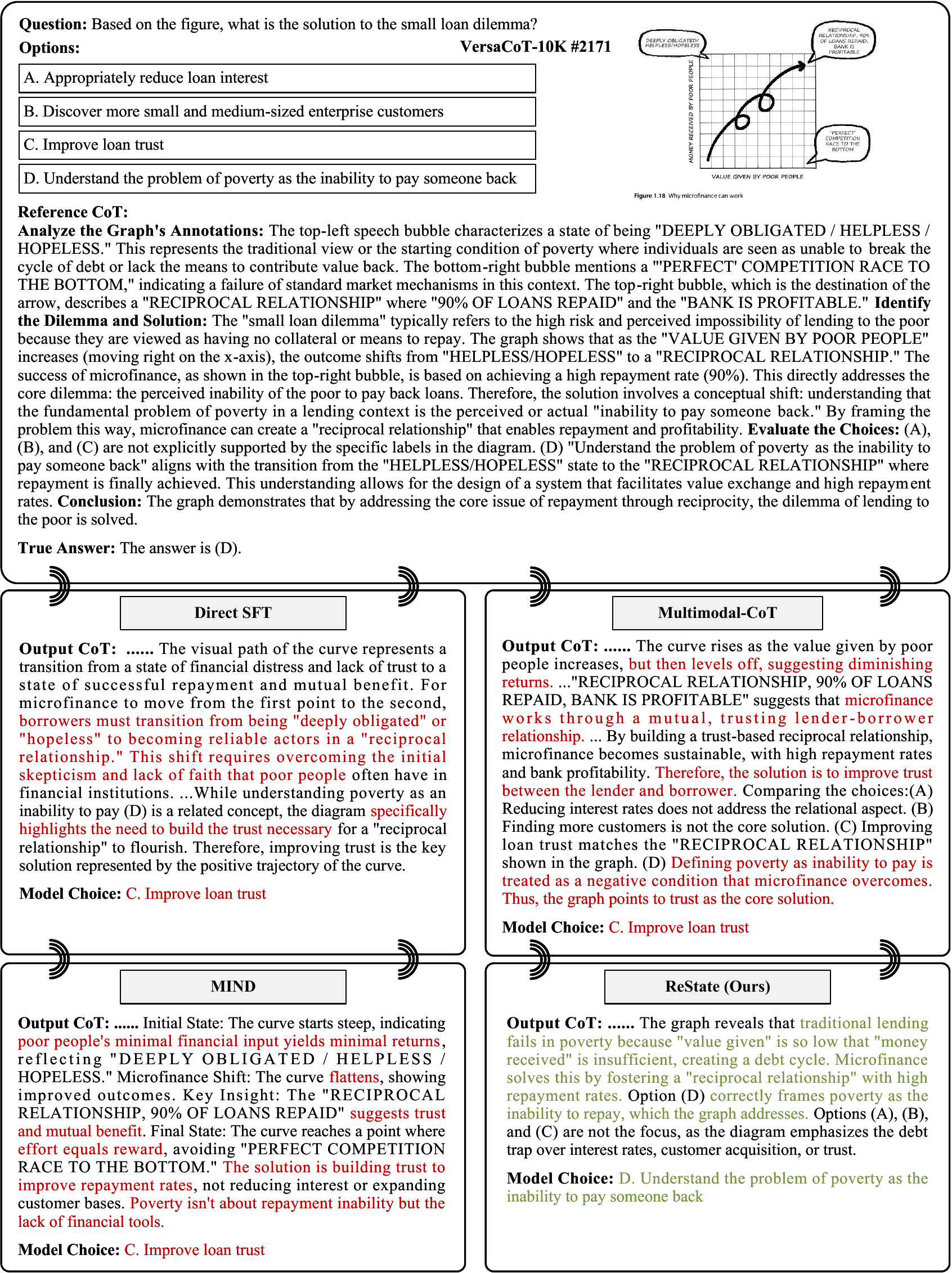}
    \vspace{-0.5cm}
    \caption{Qualitative comparison on an example of small loan dilemma reasoning in VersaCoT-10K.}
    \label{figs:versacot10k-case1}
    \vspace{-0.2cm}
\end{figure*}

\begin{figure*}[t]
    \centering
    \includegraphics[width=\linewidth]{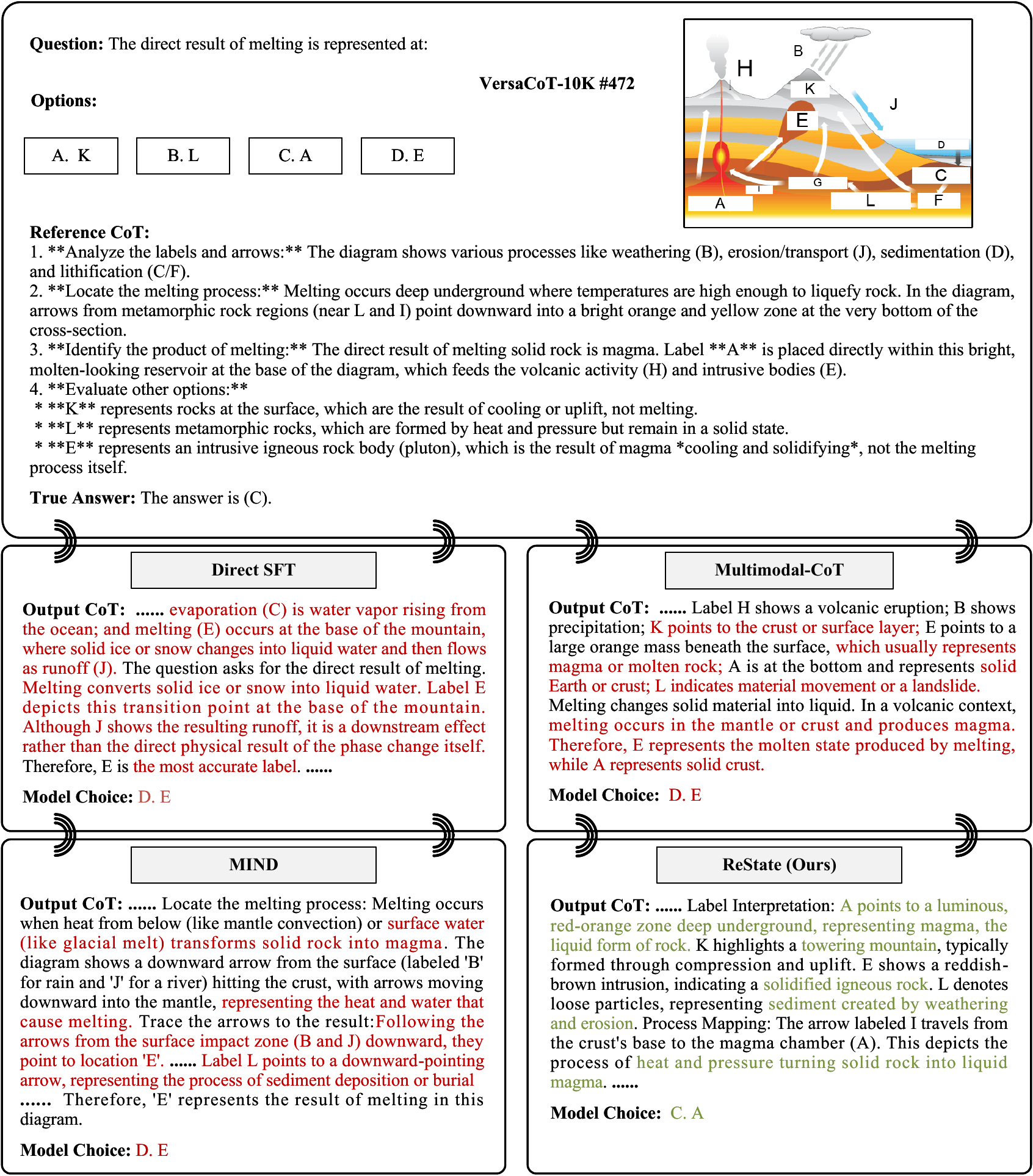}
    \vspace{-0.5cm}
    \caption{Qualitative comparison on an example of rock cycle reasoning in VersaCoT-10K.}
    \label{figs:versacot10k-case2}
    \vspace{-0.2cm}
\end{figure*}

\begin{figure*}[t]
    \centering
    \includegraphics[width=\linewidth]{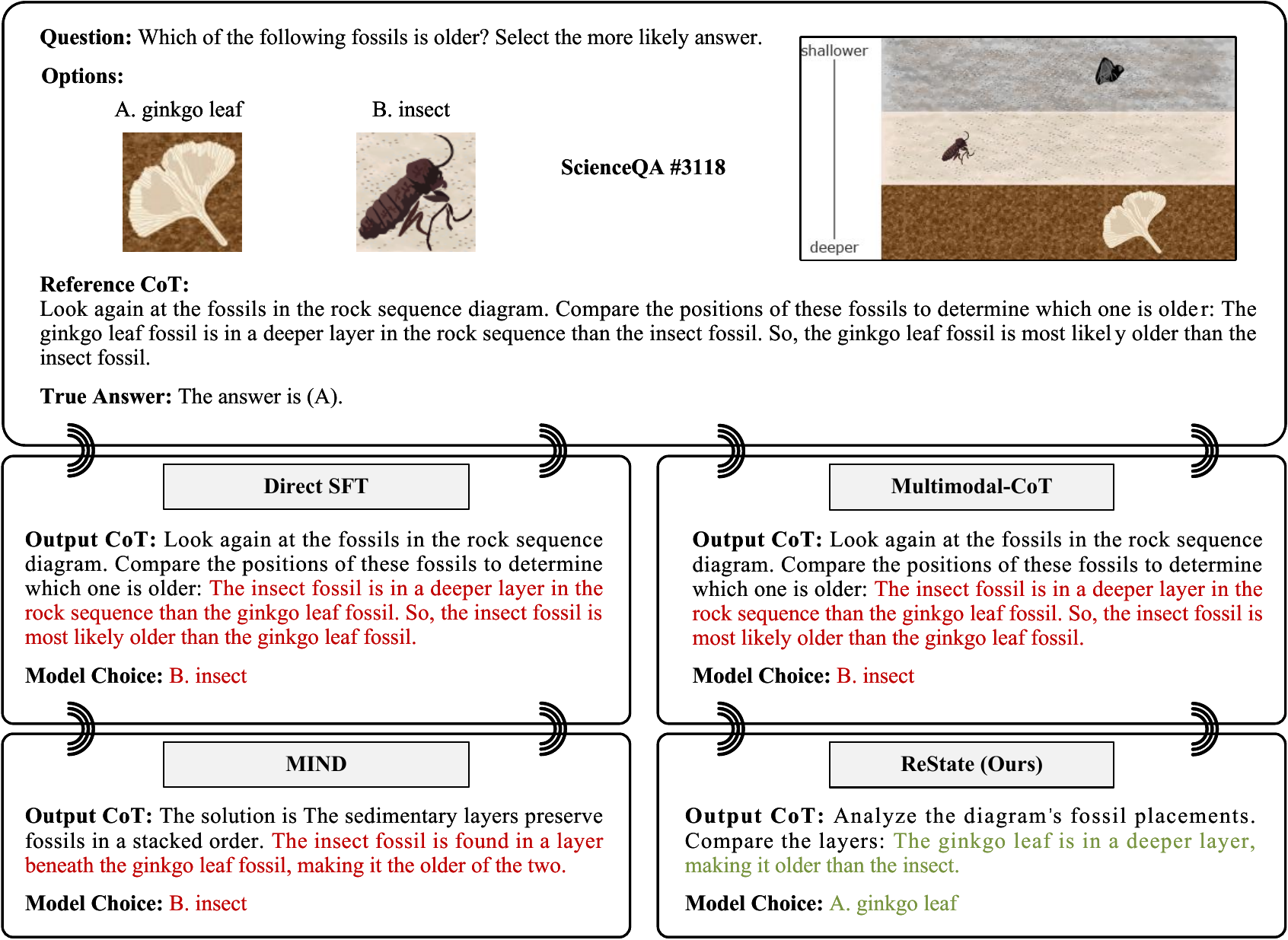}
    \vspace{-0.5cm}
    \caption{Qualitative comparison on an example of fossil age reasoning in ScienceQA.}
    \label{figs:scienceqa-case1}
    \vspace{-0.2cm}
\end{figure*}

\begin{figure*}[t]
    \centering
    \includegraphics[width=\linewidth]{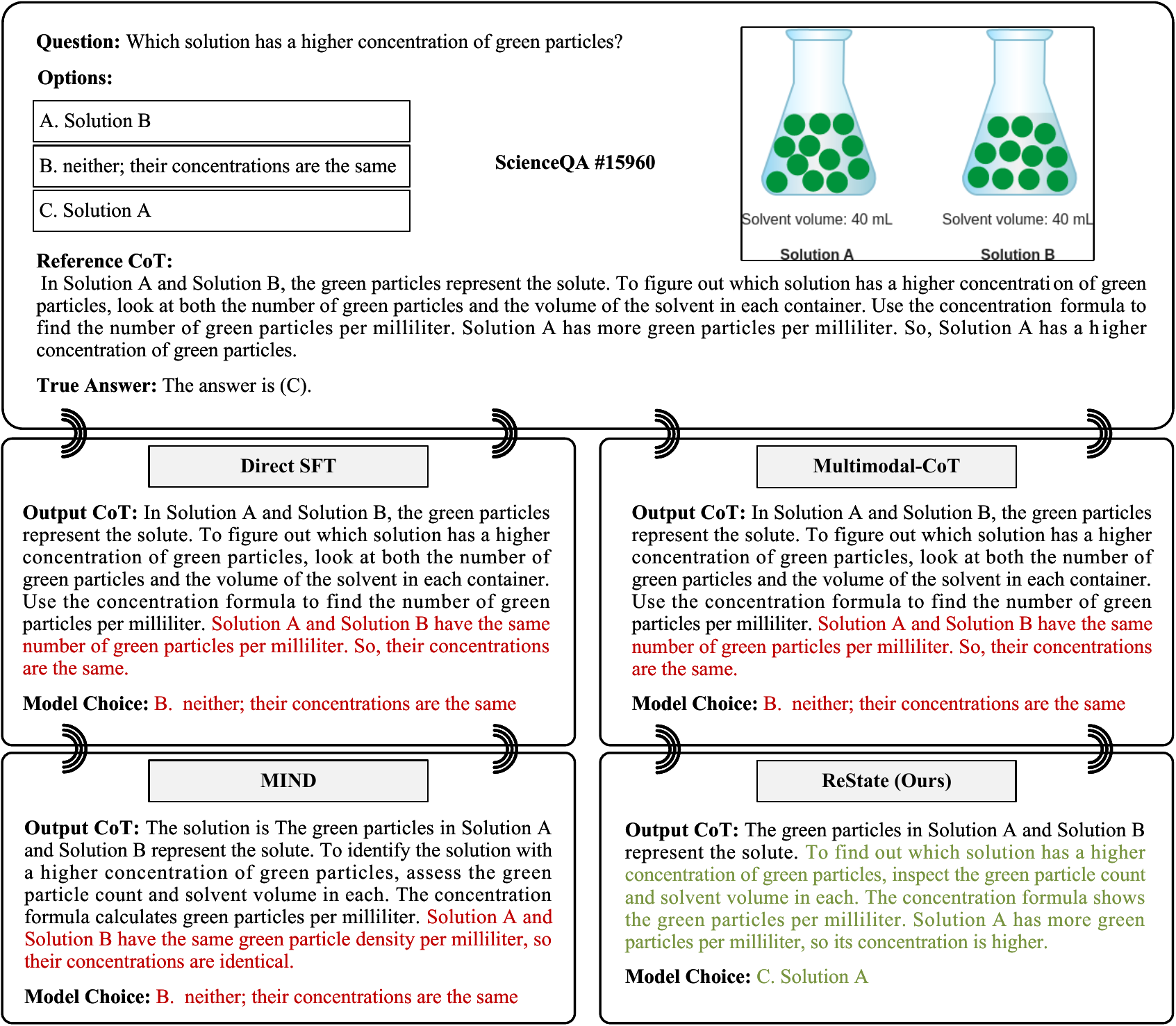}
    \vspace{-0.5cm}
    \caption{Qualitative comparison on an example of solution concentration reasoning in ScienceQA.}
    \label{figs:scienceqa-case2}
    \vspace{-0.2cm}
\end{figure*}

\begin{figure*}[t]
    \centering
    \includegraphics[width=\linewidth]{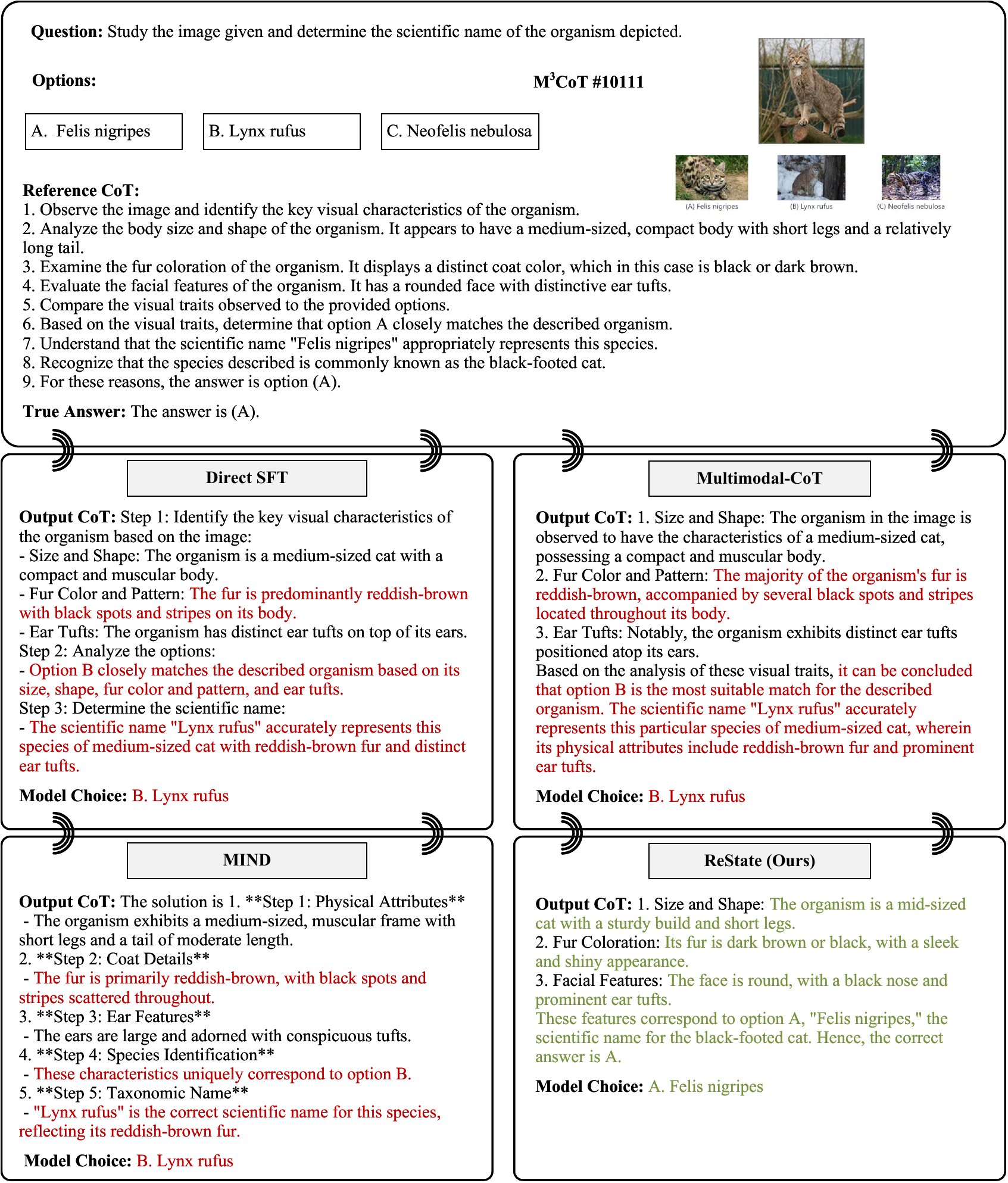}
    \vspace{-0.5cm}
    \caption{Qualitative comparison on an example of species recognition in M$^3$CoT.}
    \label{figs:m3cot-case1}
    \vspace{-0.2cm}
\end{figure*}

\begin{figure*}[t]
    \centering
    \includegraphics[width=\linewidth]{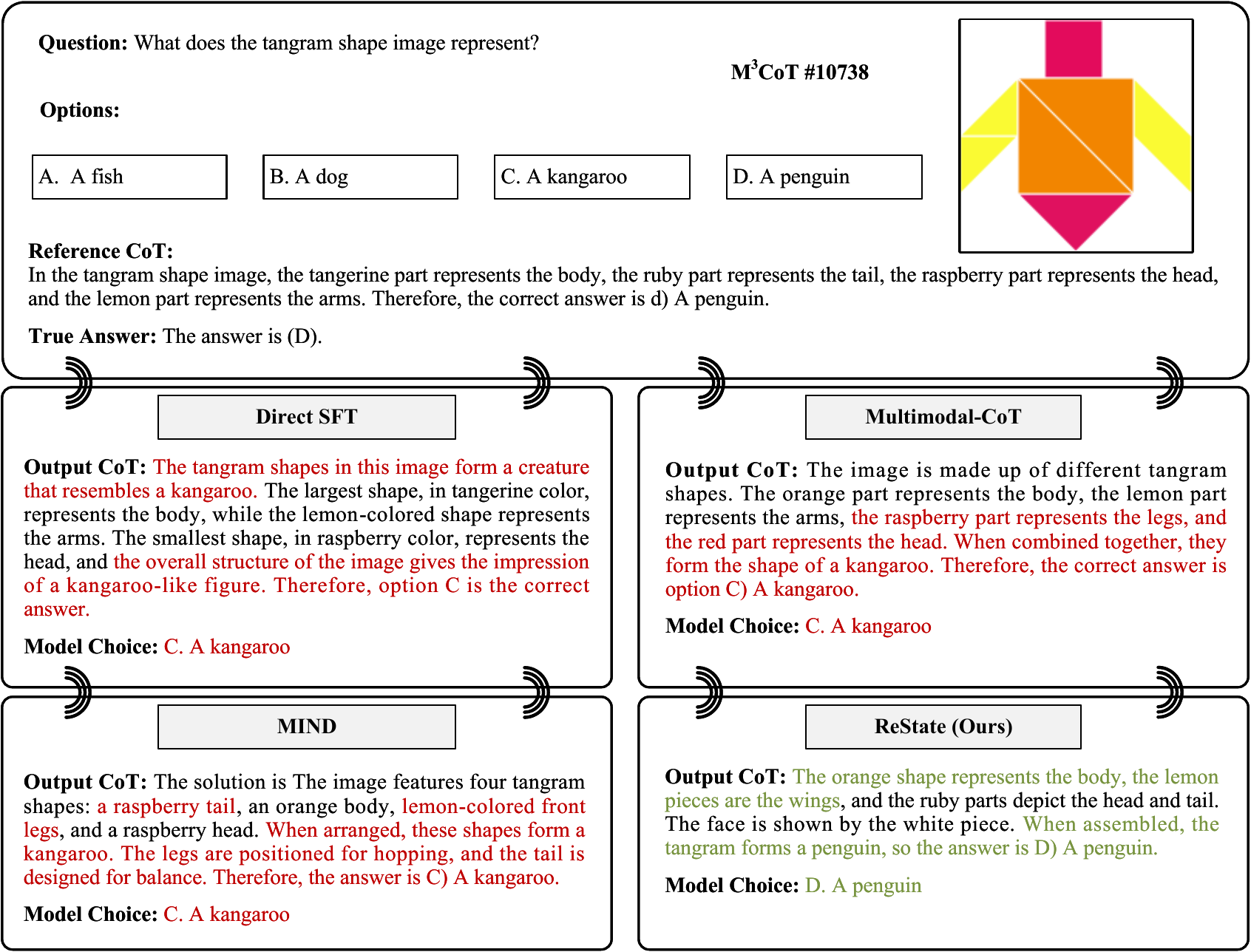}
    \vspace{-0.5cm}
    \caption{Qualitative comparison on an example of tangram recognition in M$^3$CoT.}
    \label{figs:m3cot-case2}
    \vspace{-0.2cm}
\end{figure*}

\subsection{Evaluation on the M$^3$CoT Dataset}
\label{sec:s-evaluation-on-the-m3cot-dataset}

To provide a more comprehensive evaluation of complex multimodal reasoning, we extend the M$^3$CoT comparisons reported in \cref{tab:tab03} of the main text. The complete results in \cref{tab:s-tab10} cover tool-usage methods~\citep{shen2023hugginggpt,wu2023visual,you2023idealgpt,lu2023chameleon}, several families of zero-shot MLLMs~\citep{peng2023kosmos,dai2023instructblip,liu2024improved,wang2024cogvlm,team2023gemini,openai2023gpt4v,bai2025qwen2,bai2025qwen3vltechnicalreport,qwen3.5,qwen36plus,qwen37,lai2026minimaxsparseattention,vteam2026glm45vglm41vthinkingversatilemultimodal,xai2026grok46,qwen3.8,openai2024gpt4ocard,openai2026gpt56,google2025gemini3flash,google2026gemini35flash,anthropic2025haiku45,anthropic2026sonnet5,anthropic2026opus5,bytedance2026seed21,kimiteam2026kimik3openfrontier}, and additional fine-tuning methods~\citep{zhang2023llama,liu2024improved,wang2024cogvlm,zhang2024multimodal,tan2024boosting,yu2026mind,bai2025qwen2,bai2025qwen3vltechnicalreport,qwen3.5}. M$^3$CoT spans science, commonsense, and mathematics, with subtasks characterized by substantially different knowledge structures and reasoning patterns, providing a challenging testbed for heterogeneous multimodal reasoning.

Existing methods exhibit substantial performance variation on M$^3$CoT. Early tool-usage methods achieve average accuracies below 35\%, while GPT-4V~\citep{openai2023gpt4v} reaches 56.95\%. With stronger MLLM backbones, zero-shot accuracy increases to 59.79\%, 61.43\%, and 62.81\% for Qwen2.5-VL-7B~\citep{bai2025qwen2}, Qwen3-VL-8B~\citep{bai2025qwen3vltechnicalreport}, and Qwen3.5-9B~\citep{qwen3.5}, respectively. More recent MLLMs achieve substantially stronger results: Gemini-3.5-Flash~\citep{google2026gemini35flash}, Kimi-K3~\citep{kimiteam2026kimik3openfrontier}, Grok-4.6~\citep{xai2026grok46}, and Gemini-3-Flash~\citep{google2025gemini3flash} reach 82.96\%, 81.32\%, 81.02\%, and 80.33\%, respectively. Despite these advances, substantial performance differences remain across individual subtasks, reflecting the heterogeneous reasoning requirements of M$^3$CoT.

More importantly, under matched backbone and training settings, ReState consistently achieves the highest average performance across all three backbones. With Qwen2.5-VL-7B, ReState reaches 84.99\%, outperforming Direct SFT, Multimodal-CoT~\citep{zhang2024multimodal}, and MIND~\citep{yu2026mind} by 5.57\% (from 79.42\% to 84.99\%), 8.67\% (from 76.32\% to 84.99\%), and 0.18\% (from 84.81\% to 84.99\%), respectively. With Qwen3-VL-8B, ReState achieves 87.96\%, corresponding to gains of 6.34\% (from 81.62\% to 87.96\%), 9.01\% (from 78.95\% to 87.96\%), and 0.51\% (from 87.45\% to 87.96\%). With Qwen3.5-9B, ReState further reaches 89.00\%, improving by 4.27\% (from 84.73\% to 89.00\%), 13.07\% (from 75.93\% to 89.00\%), and 0.91\% (from 88.09\% to 89.00\%), respectively.

In particular, the gains over Multimodal-CoT range from 8.67\% to 13.07\%, substantially larger than those observed on ScienceQA and VersaCoT-10K. This suggests that explicitly modeling and recursively refining reasoning states is especially beneficial on M$^3$CoT, where diverse knowledge structures and reasoning patterns place greater demands on intermediate reasoning. ReState with Qwen3.5-9B also achieves an average accuracy of 89.00\%, exceeding all evaluated zero-shot MLLMs in \cref{tab:s-tab10}. We emphasize that fine-tuned and zero-shot models are evaluated under different settings and therefore do not constitute strictly matched comparisons. Nevertheless, this result shows that ReState can substantially enhance the reasoning capability of a medium-scale MLLM and enable highly competitive performance on complex multimodal reasoning. It also narrows the gap to the reported human performance of 91.61\% to 2.61\%.

The fine-grained results further show that the gains of ReState extend across science, commonsense, and mathematics. With Qwen3.5-9B, ReState achieves the best results within the matched-backbone comparison on multiple subtasks, including 80.00\% on \textit{Algebra}, 83.75\% on \textit{Geometry}, and 90.48\% on \textit{Theory}. In particular, \textit{Theory} accuracy increases from 61.90\% with Direct SFT to 90.48\% with ReState. Similar improvements across multiple subtasks indicate that the overall performance gain is not driven by a single reasoning category, but instead reflects broader benefits across heterogeneous reasoning patterns.

Overall, the extended M$^3$CoT results demonstrate particularly clear gains from ReState on a benchmark with diverse and challenging reasoning requirements. ReState consistently improves average performance across all three backbones and remains effective across scientific, commonsense, and mathematical reasoning. Together with the results on VersaCoT-10K and ScienceQA, these findings further support the effectiveness and cross-dataset applicability of reformulating multimodal reasoning from stage-centric pipelines into state-centric iterative refinement.

\section{Qualitative Analysis}
\label{sec:qualitative-analysis}

\subsection{Comparison with Different Methods}
To further examine the reasoning behaviors of different methods, we present representative examples from VersaCoT-10K, ScienceQA, and M$^3$CoT, comparing Direct SFT, Multimodal-CoT, MIND, and ReState in~\cref{figs:versacot10k-case1,figs:versacot10k-case2,figs:scienceqa-case1,figs:scienceqa-case2,figs:m3cot-case1,figs:m3cot-case2}. Beyond final-answer correctness, these examples reveal clear differences in visual evidence interpretation, relation modeling, and reasoning consistency.

On VersaCoT-10K, Direct SFT, Multimodal-CoT, and MIND are more susceptible to misleading local visual or textual cues. In the ``small loan dilemma'' example, all three methods overemphasize the ``reciprocal relationship'' and high repayment rate, and consequently misidentify loan trust as the central issue. In contrast, ReState reasons over the broader semantic structure of the diagram and correctly identifies repayment ability as the key factor. Similarly, in the ``rock cycle'' example, the compared methods incorrectly identify E as the direct product of melting. ReState instead distinguishes the magma region from the solidified intrusive rock and correctly identifies A as the material produced by melting. These examples suggest that ReState more reliably captures global visual structures and process relations rather than relying primarily on local surface associations.

A similar pattern is observed on ScienceQA and M$^3$CoT. In the ``fossil age'' and ``solution concentration'' examples, the compared methods make incorrect predictions due to errors in spatial interpretation, quantitative comparison, or proportional reasoning, whereas ReState more accurately integrates the positions, quantities, and relations. In the ``fine-grained species recognition'' and ``tangram recognition'' examples, ReState also combines multiple visual attributes and global structural cues more effectively, reducing errors caused by over-reliance on isolated appearance features.

Overall, the compared methods can produce fluent and superficially plausible CoTs even when their intermediate reasoning is grounded in incorrect visual observations or misleading local associations. Such errors may then propagate through the reasoning chain and affect the final prediction. In contrast, ReState organizes and updates explicit reasoning states around task-relevant multimodal evidence, yielding stronger consistency among visual observations, intermediate reasoning, and final answers. These qualitative examples provide intuitive evidence that state-centric reasoning can better support robust multimodal reasoning than conventional stage-centric pipelines.

\begin{figure*}[t]
    \centering
    \includegraphics[width=\linewidth]{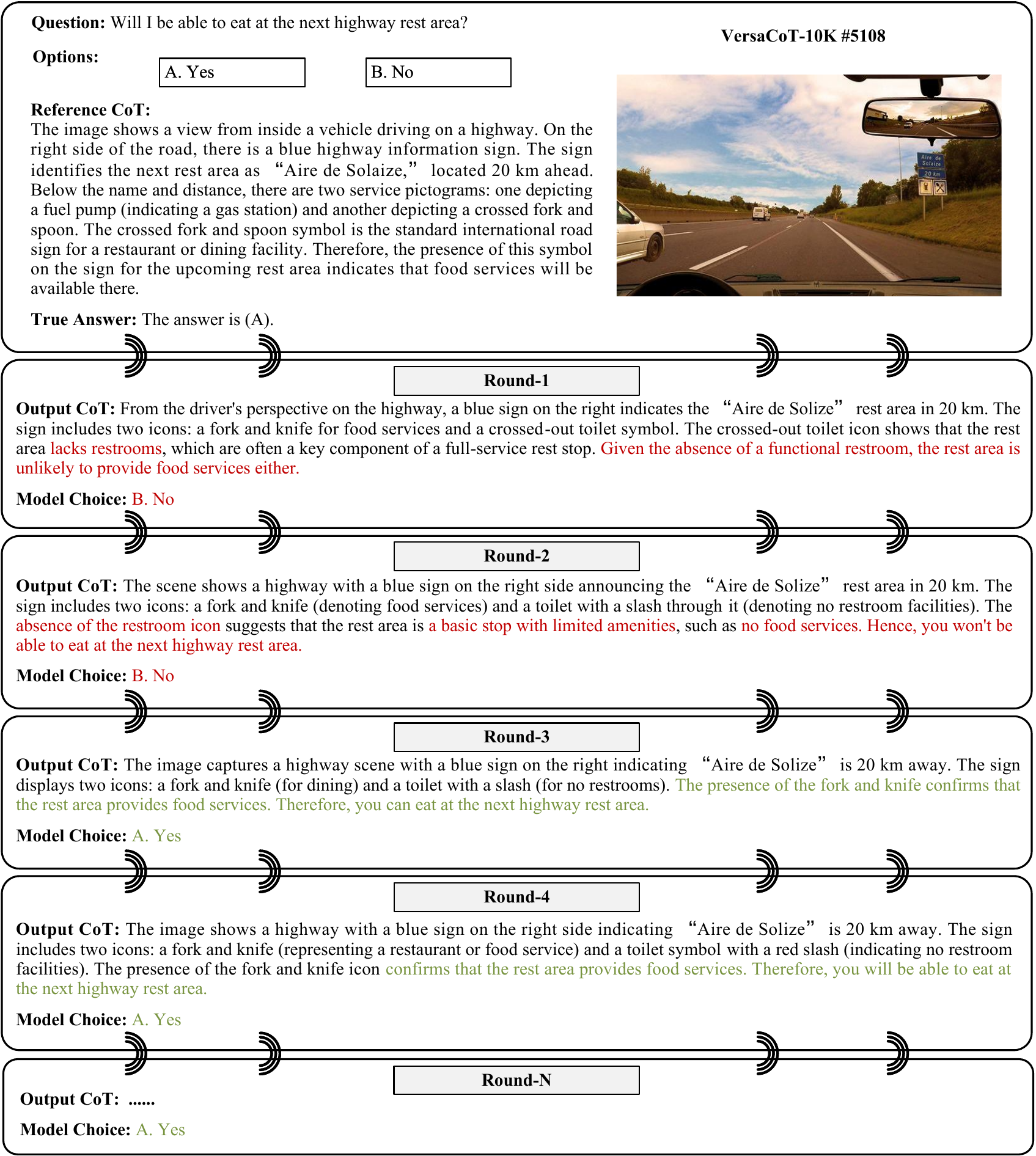}
    \vspace{-0.5cm}
    \caption{Iterative reasoning-state refinement of ReState on VersaCoT-10K.}
    \label{figs:versa10k-restate}
    \vspace{-0.2cm}
\end{figure*}

\begin{figure*}[t]
    \centering
    \includegraphics[width=\linewidth]{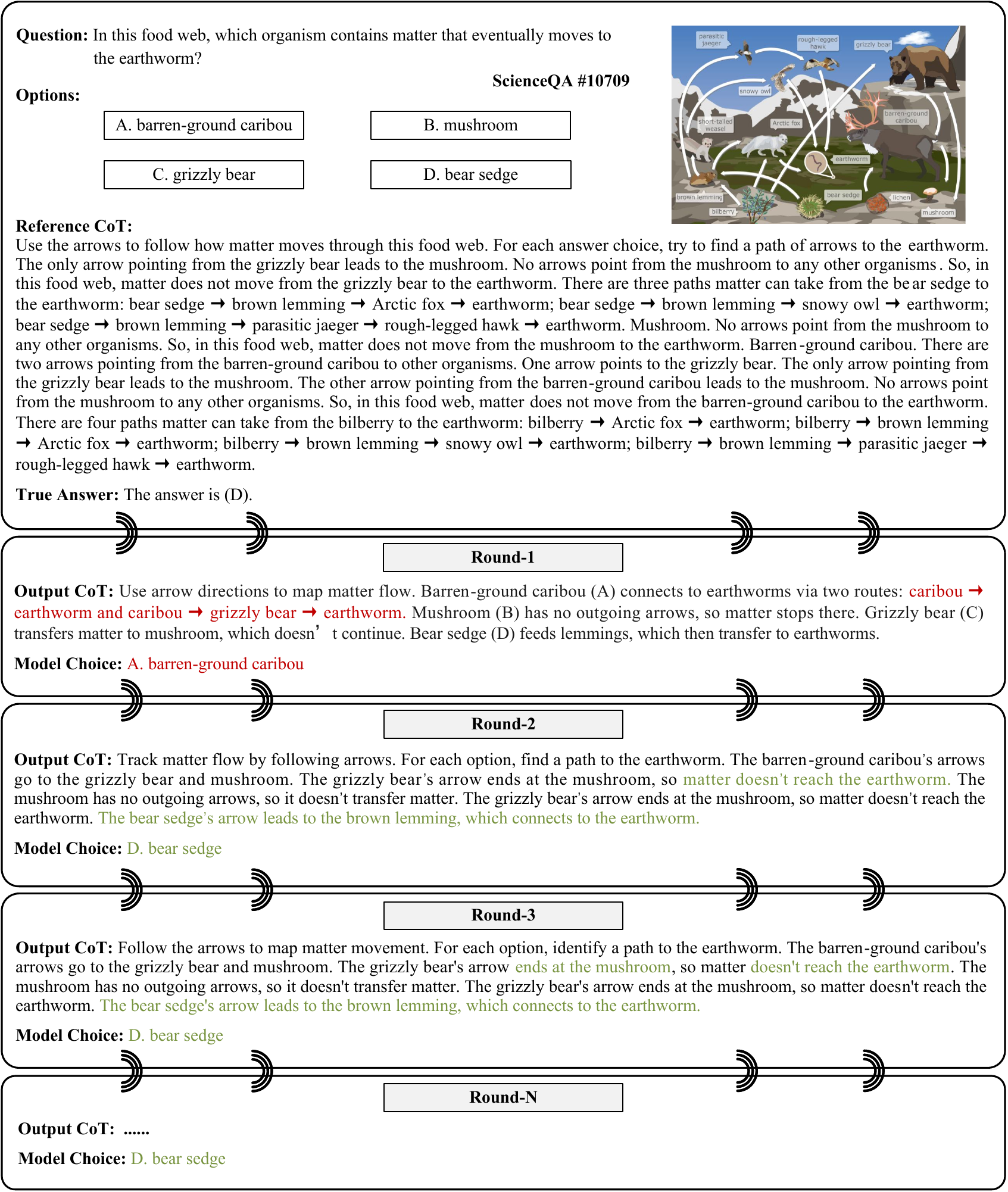}
    \vspace{-0.5cm}
    \caption{Iterative reasoning-state refinement of ReState on ScienceQA.}
    \label{figs:scienceqa-restate}
    \vspace{-0.2cm}
\end{figure*}

\begin{figure*}[t]
    \centering
    \includegraphics[width=\linewidth]{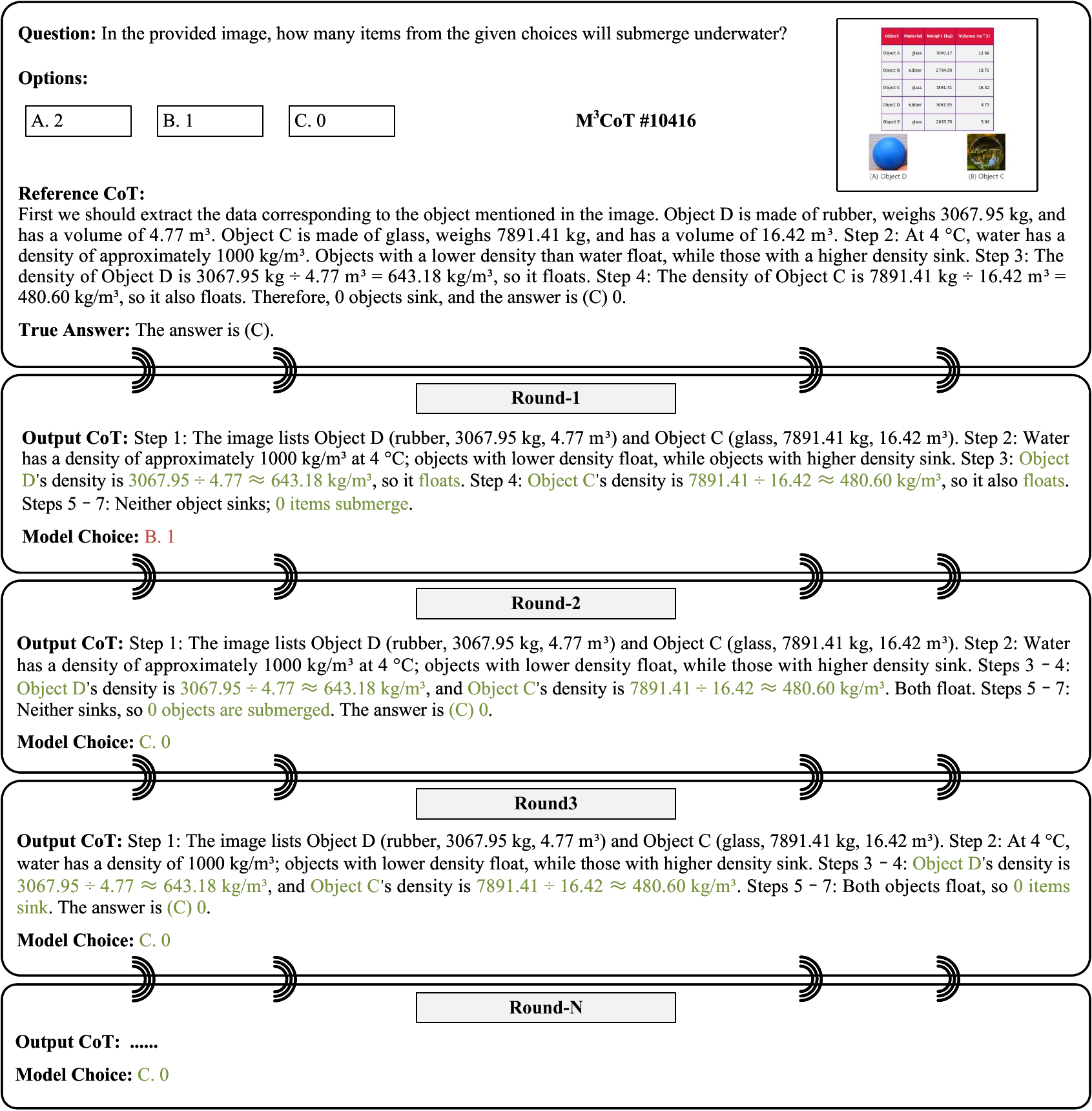}
    \vspace{-0.5cm}
    \caption{Iterative reasoning-state refinement of ReState on M$^3$CoT.}
    \label{figs:m3cot-restate}
    \vspace{-0.2cm}
\end{figure*}

\subsection{Iterative Self-Refinement Capability of ReState}
We further visualize the evolution of reasoning states during recursive inference to examine the iterative self-refinement capability of ReState. As shown in~\cref{figs:versa10k-restate,figs:scienceqa-restate,figs:m3cot-restate}, when an early reasoning state contains an incorrect judgment, ReState can feed the current state back into the same state-update operator, re-examine task-relevant visual evidence and reasoning relations, and progressively revise the erroneous state.

For example, in the highway scene in~\cref{figs:versa10k-restate}, the initial reasoning state is distracted by irrelevant facility information and incorrectly predicts that food is unavailable. After recursive state feedback, the model shifts its attention to the fork-and-knife icon that is directly relevant to the question, revises the answer to ``Yes,'' and preserves the corrected reasoning in subsequent iterations. In the ``food web'' example in~\cref{figs:scienceqa-restate}, the initial state constructs an incorrect matter-transfer relation. After receiving its own reasoning state as input again, the model re-examines the arrow directions and inter-organism connections, corrects the intermediate relation, and reaches the correct answer. In~\cref{figs:m3cot-restate}, the initial state correctly computes the densities of the two objects and concludes that neither should sink, but nevertheless produces an inconsistent final choice of ``B. 1.'' Through recursive state feedback, ReState preserves the valid density calculations and flotation judgments while correcting the final answer to ``C. 0,'' which is then maintained in subsequent iterations.

These examples illustrate a typical reasoning-state evolution pattern: \textit{erroneous state $\rightarrow$ evidence re-evaluation $\rightarrow$ state correction $\rightarrow$ correct-state preservation}. Importantly, this process is not merely repeated answer generation. Instead, the model explicitly conditions each update on its current reasoning state, allowing previously generated reasoning to be preserved, reconsidered, or corrected. This qualitative behavior is consistent with the quantitative results in the main text, where most gains are obtained within the first few refinement rounds before performance gradually stabilizes.

The entire refinement process reuses the same MLLM and the same reasoning-state update operator, without additional reflection module, critic, or independent correction model. By incorporating self-generated historical states during training, CRSL exposes the model to states closer to those encountered during recursive inference. At inference, ReState follows a closed loop of ``state feedback $\rightarrow$ re-evaluation $\rightarrow$ state update,'' progressively correcting erroneous states while preserving valid reasoning. These qualitative examples provide intuitive evidence for the iterative self-refinement capability enabled by closed-loop reasoning-state learning.

\section{More Discussion}
\label{sec:more-discussion}

\subsection{Generalization Evaluation}
To further evaluate the cross-dataset transferability of ReState, we directly evaluate models trained on VersaCoT-10K on ScienceQA and M$^3$CoT without additional fine-tuning on the target datasets. We compare ReState with the original pretrained models without SFT and models trained with Direct SFT on the same VersaCoT-10K data. This setting examines whether the reasoning-state transition capability learned by ReState can transfer to unseen benchmarks beyond the dataset used for training.

As shown in \cref{tab:s-tab11}, under the VersaCoT-10K $\rightarrow$ ScienceQA setting, ReState achieves 85.05\% with Qwen2.5-VL-7B and 91.94\% with Qwen3.5-9B. These results improve over the corresponding original models by 0.66\% and 0.69\%, and over Direct SFT by 0.35\% and 0.40\%, respectively. With Qwen3-VL-8B, ReState reaches 90.10\%, which is 0.52\% lower than the original model but still higher than the 89.55\% obtained by Direct SFT. Overall, on ScienceQA, where the pretrained backbones already exhibit strong performance, ReState largely preserves their original capabilities while providing more consistent cross-dataset transfer than Direct SFT.

The advantage is more evident under the VersaCoT-10K $\rightarrow$ M$^3$CoT setting. ReState achieves 61.17\%, 62.12\%, and 65.10\% with Qwen2.5-VL-7B, Qwen3-VL-8B, and Qwen3.5-9B, respectively, improving over the corresponding original models by 1.38\%, 0.69\%, and 2.29\%, and over Direct SFT by 2.24\%, 4.14\%, and 1.99\%, respectively. Notably, Direct SFT decreases performance relative to the original models by 0.86\% on Qwen2.5-VL-7B and 3.45\% on Qwen3-VL-8B, whereas ReState improves all three backbones. These results indicate that the benefits of ReState are not confined to the VersaCoT-10K training distribution and can transfer to a different multimodal reasoning benchmark.

Overall, the cross-dataset results demonstrate promising transferability of ReState. Unlike Direct SFT, which primarily learns a direct mapping from the training data to target outputs, ReState jointly learns reasoning construction, correct reasoning preservation, and erroneous reasoning correction through a shared reasoning-state transition mechanism, while CRSL further exposes the model to self-generated historical states during training. This learning paradigm may encourage reasoning behaviors that are less tied to a specific dataset format and more reusable across tasks. Although ReState does not improve over the original model in every transfer setting, it more consistently preserves or enhances performance across datasets and backbones, with particularly clear gains on M$^3$CoT. These results further support the cross-dataset applicability of state-centric reasoning and its potential to improve reasoning transfer across heterogeneous multimodal tasks.

\subsection{Training--Inference Reasoning-State Gap}
As shown in \cref{figs:fig-01} and \cref{tab:tab01,tab:tab02,tab:tab03} of the main text, the two-stage Multimodal-CoT baseline does not consistently outperform single-stage Direct SFT under matched backbone and training settings. Instead, it performs worse than Direct SFT in eight of the nine dataset-backbone combinations. This degradation is particularly evident on M$^3$CoT. With Qwen2.5-VL-7B and Qwen3-VL-8B, Multimodal-CoT decreases accuracy by 3.10\% (from 79.42\% to 76.32\%) and 2.67\% (from 81.62\% to 78.95\%), respectively. With Qwen3.5-9B, the decrease further reaches 8.80\% (from 84.73\% to 75.93\%). Similar performance drops are observed across all three backbones on ScienceQA and on two of the three backbones on VersaCoT-10K.

This empirical pattern is consistent with the training-inference reasoning-state gap discussed for stage-centric MCoT methods. A typical stage-centric pipeline decomposes reasoning into CoT generation and answer inference. During training, the answer-inference stage is commonly optimized with human-annotated or pre-constructed reasoning chains, whereas at inference time it receives reasoning generated by the preceding model. Such model-generated states may contain omissions, local errors, or reasoning patterns that differ from those observed during training, resulting in a mismatch between the reasoning-state distributions encountered during training and inference. Moreover, existing stage-centric methods~\citep{zhang2024multimodal,tan2024boosting,he2024multi,zheng2023ddcot,wang2024t} generally treat intermediate CoTs as stage-specific outputs rather than reusable states that can be continuously re-evaluated and refined. Consequently, errors introduced by an upstream reasoning stage may be directly propagated to downstream answer prediction.

In contrast, Direct SFT does not condition a separately optimized downstream stage on an intermediate reasoning state, thus avoiding this source of cross-stage mismatch. The comparison suggests that explicit CoT supervision alone does not necessarily guarantee improved reasoning performance. Effective reasoning also depends on how intermediate reasoning states are modeled and how well their training-time and inference-time distributions are aligned. Although these comparisons do not isolate the reasoning-state gap as the sole cause of the observed degradation, they provide empirical motivation for explicitly modeling model-induced reasoning states during learning.

ReState addresses this issue by reformulating intermediate reasoning from stage-specific outputs into continuously updatable states. URSM unifies reasoning construction, correct reasoning preservation, and erroneous reasoning correction as transitions within a shared reasoning-state space. Building on this unified state-update mechanism, CRSL further incorporates self-generated historical states during training, exposing the model to model-induced states that more closely resemble those encountered during recursive inference. At inference, the same state-update operator recursively feeds back and refines the current reasoning state. Consistent with this design, ReState outperforms Multimodal-CoT across all three datasets and all three backbones. On M$^3$CoT in particular, the gains range from 8.67\% (from 76.32\% to 84.99\%) to 13.07\% (from 75.93\% to 89.00\%).

Overall, these results suggest that effective learning-based MCoT requires both explicit reasoning supervision and consistent reasoning-state modeling across training and inference. By treating intermediate CoTs as iteratively updatable states and incorporating model-induced historical states into closed-loop learning, ReState provides a unified state-centric mechanism for mitigating the training-inference reasoning-state gap and enabling iterative refinement.

\begin{table*}[!t]
\centering
\caption{Cross-dataset generalization of ReState from VersaCoT-10K to ScienceQA and M$^3$CoT.}
\vspace{-6pt}
\label{tab:s-tab11}
\setlength{\tabcolsep}{1.5mm}
\renewcommand{\arraystretch}{1.2}
\resizebox{1\linewidth}{!}{

\begin{tabular}{c|c|c|c|c|c|c}
\hline
\multirow{2}{*}{Setting}
& \multicolumn{3}{c|}{VersaCoT-10K $\rightarrow$ ScienceQA}
& \multicolumn{3}{c}{VersaCoT-10K $\rightarrow$ M$^3$CoT}
\\
\cline{2-7}

& Qwen2.5-VL-7B
& Qwen3-VL-8B
& Qwen3.5-9B
& Qwen2.5-VL-7B
& Qwen3-VL-8B
& Qwen3.5-9B
\\
\hline

Origin (No SFT)
& 84.39 ({\color[HTML]{7030A0}+0.66})
& \textbf{90.62} ({\color[HTML]{7030A0}-0.52})
& 91.25 ({\color[HTML]{7030A0}+0.69})
& 59.79 ({\color[HTML]{7030A0}+1.38})
& 61.43 ({\color[HTML]{7030A0}+0.69})
& 62.81 ({\color[HTML]{7030A0}+2.29})
\\
\hline

Origin (Direct SFT)
& 84.70 ({\color[HTML]{7030A0}+0.35})
& 89.55 ({\color[HTML]{7030A0}+0.55})
& 91.54 ({\color[HTML]{7030A0}+0.40})
& 58.93 ({\color[HTML]{7030A0}+2.24})
& 57.98 ({\color[HTML]{7030A0}+4.14})
& 63.11 ({\color[HTML]{7030A0}+1.99})
\\
\hline

\rowcolor[HTML]{F1F1F1}
\textbf{ReState (Ours)}
& \textbf{85.05}
& 90.10
& \textbf{91.94}
& \textbf{61.17}
& \textbf{62.12}
& \textbf{65.10}
\\
\hline

\end{tabular}

}

\end{table*}

\subsection{Quantitative Analysis of Generated CoTs}
To further evaluate the quality of the generated reasoning processes beyond final-answer accuracy, we employ Qwen3.8-27B~\citep{qwen3.8} as a unified judge model to assess CoTs produced by different methods on VersaCoT-10K. We evaluate five dimensions: \textit{Overall}, \textit{Correctness}, \textit{Relevance}, \textit{Coherence}, and \textit{Solving Process}. As shown in \cref{tab:s-tab12,tab:s-tab13,tab:s-tab14}, ReState achieves the highest \textit{Overall} score across Qwen2.5-VL-7B, Qwen3-VL-8B, and Qwen3.5-9B, while also obtaining the best result on every fine-grained metric. These results indicate that the benefits of ReState extend beyond final-answer accuracy to the quality and consistency of the generated reasoning process.

With Qwen2.5-VL-7B, ReState achieves an \textit{Overall} score of 8.12, improving by 0.12 (from 8.00 to 8.12) over MIND, the strongest competing method. Its \textit{Correctness}, \textit{Relevance}, \textit{Coherence}, and \textit{Solving Process} scores reach 6.82, 9.05, 8.96, and 7.65, respectively. The improvements are more pronounced with Qwen3-VL-8B, where ReState obtains an \textit{Overall} score of 8.85, exceeding the best competing result by 0.36 (from 8.49 to 8.85). Its four fine-grained scores further reach 7.73, 9.47, 9.43, and 8.78, respectively. In particular, the \textit{Solving Process} score improves by 0.49 (from 8.29 to 8.78) over the strongest competing result, suggesting a clearer and more reliable problem-solving trajectory. With Qwen3.5-9B, ReState likewise achieves the highest \textit{Overall} score of 8.51 and consistently outperforms the compared methods across all four fine-grained dimensions. Overall, ReState achieves the strongest result for every evaluated reasoning-quality metric across all three backbones, demonstrating consistent improvements in generated CoT quality.

These improvements are consistent with the design objectives of ReState. URSM jointly models reasoning construction from empty states, correct reasoning preservation and refinement, and erroneous reasoning correction within a shared reasoning-state space. CRSL further exposes the model to self-generated historical states during training, enabling the state-update operator to better handle and refine model-induced reasoning states. Consequently, the gains of ReState are reflected not only in final prediction accuracy, but also in stronger reasoning correctness, relevance, coherence, and problem-solving quality. Together with the qualitative examples, these judge-based evaluations provide complementary evidence that state-centric reasoning improves both reasoning outcomes and the intermediate reasoning processes that lead to them.

\subsection{Limitations and Future Work}
Although ReState consistently improves performance across multiple backbones and multimodal reasoning benchmarks, several limitations remain and warrant further investigation. First, the current experiments primarily focus on multiple-choice reasoning. Although the cross-dataset results demonstrate some transfer capability, applicability to open-ended VQA, video reasoning, and longer-horizon multimodal reasoning remains to be established. Future work could extend ReState to more diverse task formats and model families to evaluate the generality and scalability of state-centric reasoning. Second, ReState achieves iterative self-refinement by repeatedly invoking the same MLLM, maintaining a unified reasoning architecture without additional reflection modules, critics, or independent correction models. However, multiple recursive updates still incur additional inference cost. The current results show that most gains occur in the first few rounds, after which performance gradually stabilizes, indicating that much of the benefit can be obtained with a relatively small number of refinement rounds. Moreover, combining the results in \cref{tab:tab01} and \cref{tab:tab04}, even without iterative self-refinement, i.e., under an inference cost comparable to MIND, ReState still improves accuracy by 0.15\%-0.90\% across the three backbones. This indicates that the gains of ReState do not solely depend on iterative self-refinement. Future work could explore more efficient recurrent reasoning mechanisms, such as Looped Transformers~\citep{giannou2023looped}, to reduce computational cost while retaining iterative refinement capability. Overall, ReState provides a unified modeling perspective that shifts multimodal reasoning from stage-centric pipelines to state-centric iterative reasoning. Future work could further improve generality, efficiency, and long-horizon refinement stability, extending state-centric reasoning to more open-ended and complex multimodal tasks.

\begin{table*}[t]
\centering
\caption{CoT quality evaluation on VersaCoT-10K with Qwen2.5-VL-7B.}
\vspace{-6pt}
\label{tab:s-tab12}
\setlength{\tabcolsep}{4mm}
\renewcommand{\arraystretch}{1.1}
\resizebox{0.95\linewidth}{!}{

\begin{tabular}{c|c|c|c|c|c}
\hline
Model
& Overall
& Correctness
& Relevance
& Coherence
& Solving Process
\\
\hline

Origin (No SFT)
& 7.30 ({\color[HTML]{7030A0}+0.82})
& 5.85 ({\color[HTML]{7030A0}+0.97})
& 8.44 ({\color[HTML]{7030A0}+0.61})
& 8.62 ({\color[HTML]{7030A0}+0.34})
& 6.29 ({\color[HTML]{7030A0}+1.36})
\\

Origin (Direct SFT)
& 7.82 ({\color[HTML]{7030A0}+0.30})
& 6.44 ({\color[HTML]{7030A0}+0.38})
& 8.84 ({\color[HTML]{7030A0}+0.21})
& 8.61 ({\color[HTML]{7030A0}+0.35})
& 7.38 ({\color[HTML]{7030A0}+0.27})
\\

Multimodal-CoT
& 7.56 ({\color[HTML]{7030A0}+0.56})
& 6.29 ({\color[HTML]{7030A0}+0.53})
& 8.37 ({\color[HTML]{7030A0}+0.68})
& 8.40 ({\color[HTML]{7030A0}+0.56})
& 7.20 ({\color[HTML]{7030A0}+0.45})
\\

MIND
& 8.00 ({\color[HTML]{7030A0}+0.12})
& 6.63 ({\color[HTML]{7030A0}+0.19})
& 9.01 ({\color[HTML]{7030A0}+0.04})
& 8.88 ({\color[HTML]{7030A0}+0.08})
& 7.49 ({\color[HTML]{7030A0}+0.16})
\\
\hline

\rowcolor[HTML]{F1F1F1}
\textbf{ReState (Ours)}
& \textbf{8.12}
& \textbf{6.82}
& \textbf{9.05}
& \textbf{8.96}
& \textbf{7.65}
\\
\hline

\end{tabular}

}

\end{table*}

\begin{table*}[t]
\centering
\caption{CoT quality evaluation on VersaCoT-10K with Qwen3-VL-8B.}
\vspace{-6pt}
\label{tab:s-tab13}
\setlength{\tabcolsep}{4mm}
\renewcommand{\arraystretch}{1.1}
\resizebox{0.95\linewidth}{!}{

\begin{tabular}{c|c|c|c|c|c}
\hline
Model
& Overall
& Correctness
& Relevance
& Coherence
& Solving Process
\\
\hline

Origin (No SFT)
& 8.21 ({\color[HTML]{7030A0}+0.64})
& 6.93 ({\color[HTML]{7030A0}+0.80})
& 9.19 ({\color[HTML]{7030A0}+0.28})
& 9.15 ({\color[HTML]{7030A0}+0.28})
& 7.55 ({\color[HTML]{7030A0}+1.23})
\\

Origin (Direct SFT)
& 8.49 ({\color[HTML]{7030A0}+0.36})
& 7.25 ({\color[HTML]{7030A0}+0.48})
& 9.28 ({\color[HTML]{7030A0}+0.19})
& 9.13 ({\color[HTML]{7030A0}+0.30})
& 8.29 ({\color[HTML]{7030A0}+0.49})
\\

Multimodal-CoT
& 8.27 ({\color[HTML]{7030A0}+0.58})
& 7.28 ({\color[HTML]{7030A0}+0.45})
& 8.87 ({\color[HTML]{7030A0}+0.60})
& 8.85 ({\color[HTML]{7030A0}+0.58})
& 8.06 ({\color[HTML]{7030A0}+0.72})
\\

MIND
& 8.35 ({\color[HTML]{7030A0}+0.50})
& 7.57 ({\color[HTML]{7030A0}+0.16})
& 9.13 ({\color[HTML]{7030A0}+0.34})
& 8.65 ({\color[HTML]{7030A0}+0.78})
& 8.04 ({\color[HTML]{7030A0}+0.74})
\\
\hline

\rowcolor[HTML]{F1F1F1}
\textbf{ReState (Ours)}
& \textbf{8.85}
& \textbf{7.73}
& \textbf{9.47}
& \textbf{9.43}
& \textbf{8.78}
\\
\hline

\end{tabular}

}

\end{table*}

\begin{table*}[!t]
\centering
\caption{CoT quality evaluation on VersaCoT-10K with Qwen3.5-9B.}
\vspace{-6pt}
\label{tab:s-tab14}
\setlength{\tabcolsep}{4mm}
\renewcommand{\arraystretch}{1.1}
\resizebox{0.95\linewidth}{!}{

\begin{tabular}{c|c|c|c|c|c}
\hline
Model
& Overall
& Correctness
& Relevance
& Coherence
& Solving Process
\\
\hline

Origin (No SFT)
& 7.97 ({\color[HTML]{7030A0}+0.54})
& 6.68 ({\color[HTML]{7030A0}+0.64})
& 9.01 ({\color[HTML]{7030A0}+0.31})
& 8.78 ({\color[HTML]{7030A0}+0.51})
& 7.42 ({\color[HTML]{7030A0}+0.71})
\\

Origin (Direct SFT)
& 8.25 ({\color[HTML]{7030A0}+0.26})
& 6.81 ({\color[HTML]{7030A0}+0.51})
& 9.12 ({\color[HTML]{7030A0}+0.20})
& 9.06 ({\color[HTML]{7030A0}+0.23})
& 8.00 ({\color[HTML]{7030A0}+0.13})
\\

Multimodal-CoT
& 8.17 ({\color[HTML]{7030A0}+0.34})
& 7.04 ({\color[HTML]{7030A0}+0.28})
& 8.81 ({\color[HTML]{7030A0}+0.51})
& 8.91 ({\color[HTML]{7030A0}+0.38})
& 7.93 ({\color[HTML]{7030A0}+0.20})
\\

MIND
& 8.42 ({\color[HTML]{7030A0}+0.09})
& 7.22 ({\color[HTML]{7030A0}+0.10})
& 9.25 ({\color[HTML]{7030A0}+0.07})
& 9.25 ({\color[HTML]{7030A0}+0.04})
& 7.98 ({\color[HTML]{7030A0}+0.15})
\\
\hline

\rowcolor[HTML]{F1F1F1}
\textbf{ReState (Ours)}
& \textbf{8.51}
& \textbf{7.32}
& \textbf{9.32}
& \textbf{9.29}
& \textbf{8.13}
\\
\hline

\end{tabular}

}

\end{table*}

\end{document}